%% file: neurips_2026.tex
\documentclass{article}
\PassOptionsToPackage{hyphens}{url} 

\PassOptionsToPackage{numbers,square,comma,sort&compress}{natbib}

 \usepackage[main, final]{neurips_2026}

\usepackage[utf8]{inputenc} 
\usepackage[T1]{fontenc}    
\usepackage{hyperref}       
\usepackage{url}            
\usepackage{booktabs}       
\usepackage{amsfonts}       
\usepackage{nicefrac}       
\usepackage{microtype}      
\usepackage{xcolor}         
\newcommand{\rev}{}

\title{Learning to Drive in New Cities Without\\ Human Demonstrations}

\usepackage{algorithm}
\usepackage{algpseudocode}

\usepackage{amsmath}
\usepackage{amssymb}
\usepackage{amsfonts}
\usepackage{bbm}
\usepackage{mathtools}
\usepackage{amsthm}
\usepackage{mathrsfs}
\usepackage{tabularx}
\usepackage{titletoc}
\usepackage{appendix}
\usepackage{multirow}
\usepackage{booktabs}
\usepackage{subcaption}
\usepackage{graphicx}
\usepackage{wrapfig}

\definecolor{linkcolor}{HTML}{671372}
\definecolor{citecolor}{HTML}{002147}
\definecolor{urlcolor}{HTML}{980000}
\hypersetup{
    colorlinks=true,
    linkcolor=linkcolor,
    citecolor=citecolor,
    urlcolor=urlcolor,
    filecolor=linkcolor,
}

\usepackage[capitalize,noabbrev,nameinlink]{cleveref}

\AtBeginDocument{\addtocontents{toc}{\protect\setcounter{tocdepth}{-1}}}

\author{%
  Zilin Wang$^{1,\dagger,*}$ \quad
  Saeed Rahmani$^{1,2,\dagger}$ \quad
  Daphne Cornelisse$^{3}$ \quad
  Bidipta Sarkar$^{1}$ \\
  \textbf{Alexander David Goldie}$^{1}$ \quad
  \textbf{Jakob Nicolaus Foerster}$^{1,\ddagger}$ \quad
  \textbf{Shimon Whiteson}$^{1,\ddagger}$ \\[0.4em]
  {\small $^{1}$BOLD, University of Oxford \quad
  $^{2}$Delft University of Technology \quad
  $^{3}$NYU Tandon School of Engineering}
}

\renewcommand{\acksection}{\section*{Acknowledgments}}

\begin{document}

\maketitle

{\renewcommand{\thefootnote}{\fnsymbol{footnote}}
\footnotetext[1]{Corresponding author: \texttt{zilin.wang@lmh.ox.ac.uk}.}
\footnotetext[2]{Core contributors.}
\footnotetext[3]{Equal supervision.}}

\input{src/abstract}
\input{src/introduction}
\input{src/relatedwork}

\input{src/preliminary}
\input{src/interactionmodel}
\input{src/methodology}
\input{src/experimentssetup}

\input{src/results}
\input{src/conclusion}
\input{src/impact}

\begin{ack}
Compute for this project is graciously provided by the Isambard-AI National AI Research Resource, under the project ``Robustness via Self-Play RL.''
Some experiments were also made possible by a generous equipment grant from NVIDIA.
This work is supported by the British Open-ended Learning and Discovery (BOLD) Lab.
Zilin Wang is funded by a generous grant from Waymo.
Saeed Rahmani is partially funded by the Transport \& Mobility Institute at Delft University of Technology.
Daphne Cornelisse is partially supported by the Cooperative AI Foundation and a Chishiki-AI SCIPE Fellowship.
Bidipta Sarkar is supported by the Clarendon Fund Scholarship in partnership with a Department of Engineering Science Studentship for his Oxford DPhil.
Alex David Goldie is funded by the EPSRC Centre for Doctoral Training in Autonomous Intelligent Machines and Systems EP/S024050/1.
Jakob Nicolaus Foerster is partially funded by the UKRI grant EP/Y028481/1 (originally selected for funding by the ERC).
The authors thank
Lukas Seier, Theo Wolf, Zengyuan Guo, Nathan Monette, Yulin Wang, Shashank Reddy, Juan Duque, Tingchen Fu, Ravi Hammond, Lu Li, Darius Muglich, and Clarisse Wibault
for helpful discussions.
We also thank the anonymous reviewers for their constructive feedback, which substantially improved this work.
\end{ack}

\bibliographystyle{unsrtnat}
\bibliography{references}


\newpage
\input{src/appendices}

\input{checklist.tex}

\end{document}

%% file: src/abstract.tex
\begin{abstract} 
While autonomous vehicles have achieved reliable performance within specific operating regions, their deployment to new cities remains costly and slow.
A key bottleneck is the need to collect many human demonstration trajectories when adapting driving policies to new cities that differ from those seen in training in terms of road geometry, traffic rules, and interaction patterns.
In this paper, we show that self-play multi-agent reinforcement learning can adapt a driving policy to a substantially different target city using only the map and meta-information,
\emph{without requiring any human demonstrations from that city}.
We introduce \textbf{NO} data \textbf{M}ap-based self-play for \textbf{A}utonomous \textbf{D}riving (NOMAD), which enables policy adaptation in a simulator constructed based on the target-city map.
Using a simple reward function, NOMAD substantially improves both task success rate and trajectory realism in target cities, demonstrating an effective and scalable alternative to data-intensive city-transfer methods.
\par\smallskip\noindent{\rev Project page: \url{https://zerlinwang.github.io/NOMAD/}}
\end{abstract}

%% file: src/introduction.tex
\section{Introduction}
\vspace{-0.5\baselineskip}

\begin{wrapfigure}{r}{0.5\linewidth}
    \centering
    \vspace{-1.0\baselineskip}
    \includegraphics[width=\linewidth]{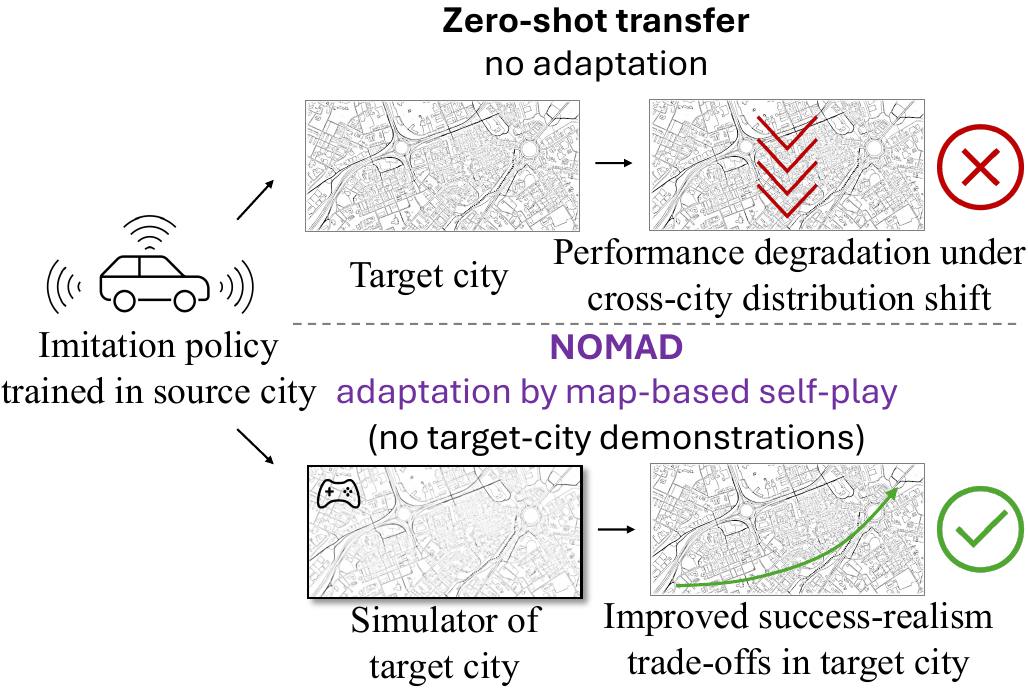}
    \vspace{-1.0\baselineskip}
    \caption{
    \textbf{Zero-shot transfer versus NOMAD.}
    }
    \vspace{-1.0\baselineskip}
    \label{fig:teaser}
\end{wrapfigure}
Autonomous vehicles have improved dramatically over the past few years and now outperform humans in certain environments \citep{kusano2025comparison, di2024comparative}. 
However, they still operate reliably in only a small fraction of the global road network \citep{waymo2025ridesmap}. 
For many current deployments, scaling and expansion remain a gradual, city-by-city effort \citep{waymo2025expansion}.
This is mainly because different cities, especially those in different countries or continents, vary in road geometry, traffic rules, and interaction patterns, i.e., the spatiotemporal behaviors induced by road topology, intersection design, and traffic density \citep{sun2023cross, li2024driving, vasudevan2025planning}. 
As a result, imitation policies trained in one city can fail to generalize reliably under cross-city deployment~\citep{feng2024unitraj}, leading to uncomfortable maneuvers, increased collision risk, and other safety-critical failures~\citep{yasarla2025roca}.


Current deployment pipelines address this problem by collecting new human demonstrations in target cities and finetuning policies on the new data \citep{reuters_waymo_tokyo_2025}.
However, this approach is slow and expensive, and hinders rapid expansion.
If human demonstrations were not needed for geographic expansion, we could substantially reduce both the cost and the time needed to deploy autonomous vehicles at scale. This motivates a fundamental question:

\begin{center}
    \textbf{\emph{Can we adapt driving policies to new cities without target-city demonstrations?}}
\end{center}

Fortunately, although collecting driving trajectories in new cities is resource intensive, other forms of city information are more readily accessible. In particular, lane-level maps and traffic meta-information, such as speed limits and traffic density, are prevalent and inexpensive to collect.


Motivated by this observation, we introduce \textbf{NO} data \textbf{M}ap-based self-play for \textbf{A}utonomous \textbf{D}riving (NOMAD).
As illustrated in \cref{fig:teaser}, 
NOMAD first constructs a simulator of the target city using its lane-level map and readily available traffic meta-information, such as speed limits and traffic density. 
This design is motivated by the hypothesis that much of the disparity in optimal driving behavior across cities is determined at the map level.
A policy trained via imitation learning in the source city is then adapted through self-play multi-agent reinforcement learning (MARL) within this simulator.
During self-play, the policy interacts with other agents and explores feasible maneuvers and interaction dynamics induced by the target-city map, thereby acquiring experience that effectively substitutes for target-city demonstrations.
Importantly, the adaptation relies on a \emph{simple reward function}, avoiding city-specific reward engineering or manual tuning. 
We achieve this by anchoring the adaptation to the source-city policy via KL regularization, which preserves the behavioral prior learned from real human driving data in the source city while allowing deviations only where required by the target-city map and interaction structure.
This selective adaptation substantially improves both success rate and behavioral realism in the target city.

Extensive closed-loop experiments demonstrate that NOMAD substantially improves zero-shot policy performance in unseen cities without relying on target-city human trajectories or complex reward engineering.
In the Boston-to-Singapore transfer setting, NOMAD increases the success rate from 42\% under zero-shot transfer to over 90\%, while simultaneously improving trajectory realism from 0.679 to approximately 0.765, as measured by the WOSAC~\citep{montali2023waymo} realism metric in closed-loop evaluation.
Furthermore, selected checkpoints produced by NOMAD form a Pareto frontier over success rate and realism, with each frontier checkpoint representing the best achievable policy for a particular success--realism trade-off.
This enables practitioners to select deployment policies based on problem-specific requirements. 

We also conduct systematic analysis, including the role of behavioral priors, the necessity of the target-city map, comparison against the policy with target-city demonstration access, generalization across diverse cities, and a sensitivity study on KL regularization strength.
Overall, these results indicate that NOMAD substantially narrows cross-city generalization gaps, supporting scalable deployment of autonomous driving systems across diverse environments and highlighting the promise of self-play MARL for improving safety and robustness.





%% file: src/relatedwork.tex
\vspace{-0.8\baselineskip}
\section{Related Work}

\vspace{-0.5\baselineskip}
\subsection{City Transfer of Autonomous Driving}
\vspace{-0.5\baselineskip}
Transferring policies across cities presents a challenge for large-scale deployment of autonomous driving services.
\citet{vasudevan2025planning} demonstrate significant behavioral differences between different cities in the nuPlan dataset.
UniTraj~\citep{feng2024unitraj} shows that even state-of-the-art trajectory prediction models trained with large datasets struggle to generalize to new cities. 
Additionally, \citet{yasarla2025roca} show that L2 error and collision rate both increase substantially when planners are zero-shot transferred to a new city.
These findings motivate the need for adaptation mechanisms between cities. 

TeraSim-World~\citep{wang2025terasim} proposes a pipeline to generate driving scenarios worldwide using the Cosmos-Drive world model~\citep{ren2025cosmos}. 
However, this approach still depends on world models that may not generate accurate scenarios for unseen cities.
As an alternative, \citet{vasudevan2025planning} propose AdaptiveDriver, a rule-based planner that adapts its behavior in new cities via interactions with a learned reactive world model. 
While their approach shows modest generalization to held-out cities within nuPlan, it still requires logged trajectories to calibrate the reactive world model in the new cities. 
Furthermore, the world model relies on IDM's~\citep{kesting2010enhanced} single-agent car-following formulation, which limits the planner's ability to reason about complex multi-agent interactions.
Moreover,
RoCA \citep{yasarla2025roca} improves cross-domain robustness of end-to-end planners via a GP-based probabilistic token model and uncertainty-guided adaptation, achieved by prioritizing the most informative target data for finetuning.

As another avenue to consider, LLaDA~\citep{li2024driving} uses Large Language Models to interpret text-based traffic codes and correct plan violations. 
While effective for explicit constraints like traffic rules, this approach struggles to capture geometry-dependent dynamics that are crucial for smooth trajectory planning but not formally codified in text.
Although \citet{wayve2025ai500} 
shows that large foundation models can enable zero-shot transfer to target cities close enough to the training distribution, demonstration collection is necessary for cities with distinctly different traffic patterns.
In contrast, NOMAD enables city transfer without reliance on learned world models, target-city demonstrations, or industrial-scale training pipelines.

Domain randomization offers a complementary approach to distribution shift.
Dynamics randomization varies physical parameters to improve robustness to differences in vehicle dynamics~\citep{peng2018simtoreal}, and visual randomization varies rendering conditions to support transfer across appearance changes~\citep{tobin2017domain}. While these approaches build robustness across variations in dynamics or appearance, NOMAD uses the available target-city map to adapt directly to geographic differences in road geometry, lane topology, traffic handedness, and the interactions they induce.

\vspace{-0.5\baselineskip}
\subsection{Self-Play for Autonomous Driving}
\vspace{-0.5\baselineskip}
Self-play is a powerful paradigm for training agents by allowing policies to improve through interactions with \textit{copies of themselves}.
CoPO~\citep{peng2021learning} uses coordinated policy optimization to simulate self-driven particle systems in traffic by enabling agents to dynamically shift between cooperative and competitive behaviors.
More recently, \citet{cusumano2025robust} demonstrate that robust and naturalistic driving can emerge from self-play.

A key challenge in self-play for driving, however, is maintaining human-like behavior while optimizing for task completion.
\citet{cornelisse2024human} address this by regularizing self-play against a human reference policy.
They also show that scaling self-play across thousands of Waymo scenarios yields reliable agents with over 99\% goal completion and minimal collisions~\citep{cornelisse2025building}.
Similarly, SPACER~\citep{chang2025spacer} introduces a self-play framework that stabilizes RL training by anchoring decentralized agents to a centralized reference policy, mitigating non-stationarity while enabling human-like behavior learning.

Beyond trajectory planning, self-play has been applied to asymmetric scenario generation~\citep{zhang2024learning}, where a teacher policy learns to create challenging yet solvable scenarios for a student, and to natural language communication for cooperative driving~\citep{cui2025talking}.
While prior self-play methods target robustness and realism within a single domain, NOMAD repurposes self-play for cross-city transfer. By conducting self-play in map-based simulators of the target city, NOMAD enables adaptation without any target-city demonstrations.

%% file: src/preliminary.tex
\vspace{-0.5\baselineskip}
\section{Preliminaries and Problem Formulation} \label{sec:preliminary}
\vspace{-0.5\baselineskip}

We model a target city $C$ as a distribution over its map and traffic scenarios, $C=(\mathcal{M}_C,\Xi_C)$.
Specifically, we sample a map segment $m\sim\mathcal{M}_C$, namely road layout and traffic direction, from a given region of the target city. 
Conditioned on $m$, we then sample a scenario $\xi=(s_0,\{g^i\}_{i=0}^{N-1}, m)\sim\Xi_C(\cdot|m)$, which specifies the number of agents $N$, the initial state $s_0$, goal positions for all agents $\{g^i\}_{i=0}^{N-1}$, and the map segment $m$.
Given a traffic scenario $\xi$, rolling out a policy $\pi$ produces a trajectory $\tau \sim q_\pi(\cdot|\xi)$, where $q_\pi(\cdot|\xi)$ denotes the trajectory distribution induced by $\pi$ in closed loop. 
We evaluate the performance of policy $\pi$ with two metrics, success rate and realism, that are broadly accepted by the community \citep{li2022metadrive,montali2023waymo,pufferdrive2025github, cusumano2025robust}.

\textbf{Success rate} is the probability of reaching the goal-centered neighborhood within the horizon $H$:
\begin{equation}
\mathcal{S}(\pi,C) := \mathbb{E}_{\xi \sim \Xi_C}\left[\mathbb{E}_{\tau \sim q_\pi(\cdot|\xi)}\left[\mathbbm{1}(\exists t\leq H: d(p_t,g)\leq \epsilon)\right]\right],
\end{equation}
where $p_t$ and $g$ denote the ego position at time $t$ and the goal position, respectively,
$\mathbbm{1}(\cdot)$ is the indicator function, and $\epsilon$ is the error tolerance in reaching the goal.

\textbf{Realism} measures how well generated trajectories match the actual distribution of real-world driving behavior.
Because the true distribution of real-world driving is unknown, we estimate realism using the likelihood of logged human trajectories under the trajectory distribution induced by $\pi$:
\begin{equation}\label{eq:realism}
\mathcal{R}(\pi, C)
    := \mathbb{E}_{\xi \sim \Xi_C}\left[\mathbb{E}_{\tau\sim p(\cdot|\xi)}\left[\log q_\pi(\tau|\xi)\right]\right],
\end{equation}
where $p(\tau|\xi)$ denotes the distribution of logged human trajectories in the scenario $\xi$.
Importantly, this realism metric requires access to target-city human trajectories \emph{only for evaluation}.
These trajectories are never used during training or adaptation, and serve solely as an offline benchmark for assessing behavior realism.
Overall, we evaluate the policy $\pi$ in city $C$ by a multi-objective score
\begin{equation}
    \mathbf{J}(\pi, C)=(\mathcal{S}(\pi, C), \mathcal{R}(\pi, C)).
\end{equation}
\textbf{City adaptation without demonstrations.}
Let $\pi^0$ be a planner trained using source-city data (e.g., via behavior cloning) and deployed zero-shot in target city $C$.
Given target-city priors (the map $\mathcal{M}_C$ and meta-information $\mathcal{I}_C$ such as speed limits and traffic density), but no target-city demonstrations, our goal is to learn a \emph{set} of adapted policies
\begin{equation}
    \Pi^+(C)=\text{Adapt}(\pi^0;\mathcal{M}_C,\mathcal{I}_C)=\{\pi^{+(1)},\ldots,\pi^{+(\mathsf{N})}\}
\end{equation}
that offer improved success--realism trade-offs in the target city.
Namely, $\Pi^+(C)$ should contain policies that Pareto-dominate the zero-shot baseline:
\begin{equation}
    \exists\,\pi^+\in \Pi^+(C)\ \text{s.t.}\ \mathbf{J}(\pi^+,C)\succ \mathbf{J}(\pi^0,C),
\end{equation}
where $\succ$ indicates Pareto dominance.
In practice, $\Pi^+(C)$ contains multiple candidate policies (e.g., checkpoints), from which we extract an empirical Pareto frontier.
Ideally, we aim to expand the achievable Pareto frontier in the target city, yielding improved trade-offs across a wide range of success–realism preferences.

\textbf{Deployment.}
At deployment time, a practitioner selects a single policy $\pi^{\text{deploy}}\in\Pi^+(C)$ according to application-specific preferences (e.g., prioritizing success rate or trajectory realism), corresponding to choosing an operating point along the induced Pareto frontier.

%% file: src/interactionmodel.tex
\vspace{-0.5\baselineskip}
\section{Multi-Agent Interaction Model}
\label{sec:interactionmodel}
\vspace{-0.5\baselineskip}

We formulate the city adaptation problem as a partially observable stochastic game~(POSG) \citep{hansen2004dynamic} defined by the tuple $(N,S,A,O,P,Z,r,H,\gamma,b_0)$, where each agent in the game is indexed by $i\in\{0,1,\cdots,N-1\}$, and $S$ is the set of possible states.
P is the state transition function, which determines the distribution of the next state $s_{t+1}$ given the current state $s_t$ and joint action $a_t$.
For agent $i$, $A^i$ and $O^i$ are the set of possible actions and observations, respectively.
$Z^i(o^i_t|s_t,a_t,s_{t+1})$ is the observation function.
$r^i(s_t,a_t,s_{t+1})$ is the scalar reward function.
$\gamma$ is the discount factor and $b_0$ is the probability of initial state $s_0$.
We focus on homogeneous agents: all vehicles share the same observation space, action space, and reward function. 
This symmetry naturally supports a shared-parameter policy, which we exploit for scalable self-play training.

\textbf{Observation space.} 
At each timestep, each dynamic vehicle $i$ receives a partial observation $o_t^i$ consisting of its local ego state (e.g., velocity and goal) and a limited-range perception of nearby vehicles and road information (e.g., edges
and lanes in a neighborhood).
All observations are expressed in an ego-centric coordinate frame (\cref{fig:obs_space}).

\textbf{Action space.}
We adopt decentralized policies $\pi(a^i_t|o_t^i)$, where temporal information such as velocity is encoded directly in the observation.
At each step, the policy predicts discrete action increments in position and heading, $a_t^i = (\delta_x^i, \delta_y^i, \delta_h^i)$.
These increments are defined in the ego vehicle’s coordinate frame and are deterministically applied using kinematic pose updates.
Each action dimension is bounded and uniformly discretized into multiple bins.
A discretized action space improves training stability during reinforcement learning and supports multi-modal behavior naturally during imitation learning.
It is commonly used in trajectory prediction and planning tasks \citep{philion2024trajeglish, cornelisse2024human, wu2024smart, zhang2025closed}.

\textbf{Reward function.}
Reward design in autonomous driving is non-trivial due to the difficulty of balancing safety, progress, and other factors \citep{knox2023reward}.
Moreover, precise reward engineering often requires substantial iterative tuning \citep{wurman2022outracing} and can be brittle under transfer, since optimized policies may overfit to environment-specific reward proxies or simulator details, leading to degraded performance under distribution shift \citep{zhang2018study, pan2022the}.
To demonstrate that NOMAD's effectiveness stems from the adaptation framework itself rather than reward tuning, we deliberately adopt a minimal reward formulation:
\begin{equation}
    r^i(s_t,a_t,s_{t+1})=w_g\mathbbm{1}_{\mathcal{G}^i(s_{t+1})}
    +w_c\mathbbm{1}_{\mathcal{C}^i(s_{t+1})}
    +w_o\mathbbm{1}_{\mathcal{O}^i(s_{t+1})}
\end{equation}
where $\mathbbm{1}(\cdot)$ is the indicator function.
$\mathcal{G}^i(s_t)$, $\mathcal{C}^i(s_t)$, and $\mathcal{O}^i(s_t)$ indicate whether agent $i$ has reached its goal, collided, or gone off-road at timestep $t$, respectively.
$w_g$, $w_c$, and $w_o$ are corresponding weights, set as $1.0$, $-0.75$, and $-0.75$, respectively, in all experiments.

%% file: src/methodology.tex
\begin{figure*}[htbp]
    \centering
    \vspace{-0.0\baselineskip}
    \includegraphics[width=0.95\linewidth]{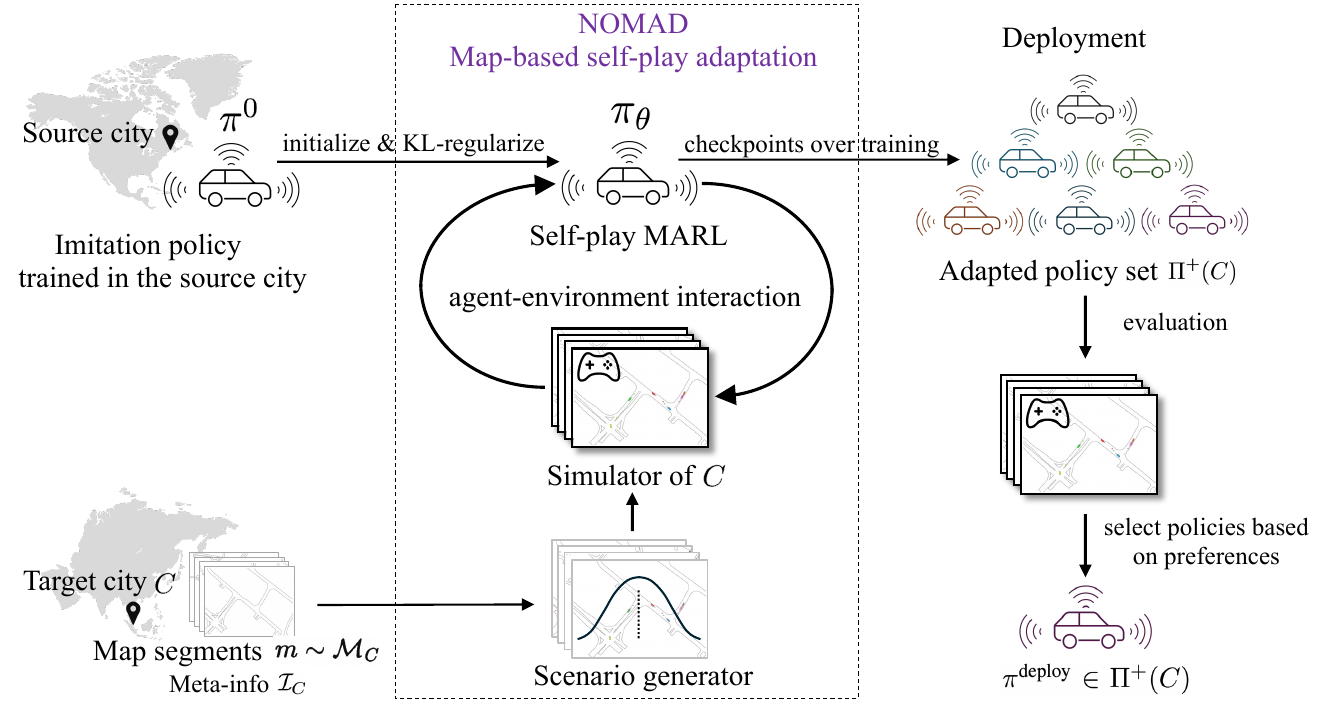}
    \vspace{-0.5\baselineskip}
    \caption{
    \textbf{NOMAD overview}.
    Starting from a source-city imitation policy $\pi^0$, NOMAD adapts it to a target city $C$ using map segments $m\sim \mathcal{M}_C$ and meta-information $\mathcal{I}_C$.
    A scenario generator samples initial states and goals.
    The generated scenarios are loaded in a data-driven multi-agent driving simulator, yielding a simulator of $C$.
    The policy $\pi_\theta$ is initialized from $\pi^0$ and optimized via KL-regularized self-play MARL.
    Checkpoints along training yield an adapted policy set $\Pi^+(C)$, from which practitioners select a deployment policy $\pi^{\text{deploy}}$ to suit their preferences.
    }
    \vspace{-1.0\baselineskip}
    \label{fig:nomad}
\end{figure*}

\vspace{-0.5\baselineskip}
\section{NOMAD}
\vspace{-0.8\baselineskip}

\subsection{Overview}
\vspace{-0.5\baselineskip}

\cref{fig:nomad} provides an overview of NOMAD.
Let $C$ be the city we want to transfer to, assuming the target-city map $\mathcal{M}_C$ and the city-specific meta-information $\mathcal{I}_C$ are accessible. $\mathcal{I}_C$ consists only of coarse information that is typically known a priori (e.g., speed limits and traffic density).
Importantly, $\mathcal{I}_C$ is low-dimensional and can be easily acquired from traffic data providers, without any trajectory data or city-specific learning (see \cref{sec:scenegen_local}).

Based on this, we construct a scenario generator $\Xi_{\varphi}(\xi|m,\mathcal{I}_C)$  
with a learned or heuristic generic parameter $\varphi$ to approximate the true distribution of scenarios $\Xi_C(\xi|m)$ in $C$.
The generator is used only to sample initial agent states and goal locations; it does
\emph{not} generate expert trajectories, driving behaviors, or policies. Crucially, NOMAD does not require a high-fidelity or carefully engineered scenario generator; any procedure that produces collision-free initial configurations on the lane network suffices, and we show in \cref{sec:sensitivitytoerrorsintrafficdensity} that performance is robust to imprecise traffic density estimates.
Details about the scenario generator used in our experiments are described in \cref{sec:scenegen}.
The generated scenarios $\xi=(s_0,\{g^i\}_{i=0}^{N-1}, m)$ are next loaded into a multi-agent data-driven autonomous driving simulator like GPUDrive \citep{kazemkhani2025gpudrive}, Waymax \citep{gulino2023waymax}, or Nocturne \citep{vinitsky2022nocturne}, yielding a simulator of the target city $C$.

Starting from an imitation learning policy $\pi^0$ from the source city, we initialize a policy $\pi_\theta$ and adapt it to the target city through KL-regularized self-play within the simulator.
Training produces a set of adapted policies $\Pi^+(C)$ that achieve Pareto improvements over the zero-shot policy $\pi^0$. 
At deployment time, practitioners select a policy $\pi^{\text{deploy}}\in\Pi^+(C)$ along the induced Pareto frontier based on their preferences over success rate and realism.

\vspace{-0.5\baselineskip}
\subsection{Policy Adaptation via Regularized Self-Play}
\vspace{-0.5\baselineskip}

NOMAD learns a shared policy for the POSG described in \cref{sec:interactionmodel} through regularized self-play.
By requiring all agents to share a single policy $\pi_\theta (a_t|o_t)$ parameterized by $\theta$, we can scale RL training to dense urban environments more effectively since the number of parameters is independent of the number of vehicles controlled.
In addition, it also increases training efficiency by allowing a single policy to learn from the diverse experiences collected by all agents simultaneously.
We train $\pi_\theta$ with independent PPO (IPPO) due to its simplicity and efficacy \citep{de2020independent, yu2022surprising}. 
IPPO treats each agent as an independent learner that optimizes a standard PPO objective using its own local observations and rewards, modeling the other agents as part of a non-stationary environment:
\begin{equation}
\resizebox{\linewidth}{!}{$\displaystyle
    \max_{\theta}\; J_{\text{PPO}}(\theta) = \mathbb{E}_{(o,a)\sim\pi_{\theta_\text{old}}} \bigg[ \min \bigg( \frac{\pi_\theta(a|o)}{\pi_{\theta_\text{old}}(a|o)} A_{\pi_{\theta_\text{old}}}(o,a),\;
    \text{clip} \left( \frac{\pi_\theta(a|o)}{\pi_{\theta_\text{old}}(a|o)}, 1-\epsilon, 1+\epsilon \right) A_{\pi_{\theta_\text{old}}}(o,a) \bigg) \bigg].
$}
\end{equation}

Given our minimal reward function, optimization without regularization could yield policies that complete goals but deviate from human-like driving patterns.
To retain useful knowledge from the source city and mitigate reward hacking, we regularize the learned policy toward the prior $\pi^0$ using forward Kullback--Leibler divergence.
This choice penalizes dropping probability mass from actions the prior deems likely, thereby retaining human-like driving while allowing selective adaptation when required by the target city.
The regularized objective is
\begin{gather}
\max_{\theta}\; J(\theta)
=J_{\text{PPO}}(\theta)
-\lambda_\mathrm{KL}\cdot \mathbb{E}_{\pi_\theta}\!\Bigg[\frac{1}{H}\sum_{t=0}^{H-1}
D_{\mathrm{KL}}\!\big(\pi^0(\cdot|o_t)\,\|\,\pi_\theta(\cdot|o_t)\big)
\Bigg],
\label{eq:kl_reg_obj}
\end{gather}
where $D_{\mathrm{KL}}$ is the KL divergence and $\lambda_{\mathrm{KL}}$ is a regularization weight.

%% file: src/experimentssetup.tex

\vspace{-0.6\baselineskip}
\section{Experimental Setup}
\label{sec:experiment_setup}
\vspace{-0.7\baselineskip}

\textbf{Datasets and Simulator.}
We use nuPlan \citep{caesar2021nuplan}, a large-scale dataset collected from the U.S. and Singapore.
NuPlan stores the scenarios separately by cities and provides cross-continental coverage, which enables studying city transfer under large distribution shifts. 
For instance, Singapore and Boston differ both in road topology and traffic handedness.
In our experiments, we randomly select 4,000 traffic scenarios for each city, partitioned into 3,200 for training and 800 for test{\rev; split and scenario statistics are given in \cref{sec:scenario_stats}}.  
For closed-loop traffic simulation, we use GPUDrive~\citep{kazemkhani2025gpudrive}, a GPU-accelerated data-driven simulation tool suitable for fast and efficient RL training.
We converted the nuPlan dataset into the GPUDrive-supported format, with 9-second scenarios discretized at 10~Hz.

\textbf{Scenario Generation.} \label{sec:scenegeneration}
We use a simple heuristic scenario generator $\Xi_\varphi(\cdot|m,\mathcal{I}_C)$ (described in \cref{sec:scenegen}) conditioned on the map, speed limits, and traffic density to produce feasible spawn and goal points for self-play simulation in the target city.
For each map segment, we generate $K=8$ distinct scenarios, yielding a total of $3{,}200 \times 8 = 25{,}600$ unique traffic scenarios for training. 


\textbf{Metrics.}
Evaluation is based on two core metrics, \emph{success rate} and \emph{realism}, defined in \cref{sec:preliminary}.
For realism, we adopt the Waymo Open Sim Agents Challenge (WOSAC) evaluation metric~\citep{montali2023waymo}.
WOSAC aggregates weighted components over kinematics (e.g., speed and acceleration), interaction features (e.g., time-to-collision and collisions), and map compliance (e.g., road departures). Using target-city human trajectories as the reference allows these features to reflect local motion and interaction patterns that can be a result of cultural differences in driving.
However, informal communication, such as signaling, horn use, and courtesy gestures, is not directly assessed by the metric, although its effects may be reflected in the resulting trajectories.
The detailed definitions and weights of each component, as well as motivation for choosing WOSAC evaluation metric over other metrics, are provided in \cref{sec:extendedmetrics}. We also report the results on each sub-metric, collision rate, offroad rate, and average displacement error (ADE) in \cref{sec:extendedmetricanalysis}.

\textbf{Training details and baselines.} 
Our main experiments focus on policy transfer from Boston to Singapore, as this pair involves two different continents with differences in driving side (investigated in \cref{sec:drivingside} and \cref{sec:samesidetransfer}), road geometry, and traffic rules. 
We report the results for other city pairs in \cref{sec:othercities}, however, unless stated otherwise, the results refer to a Boston-to-Singapore transfer by default. 
For the zero-shot baseline $\pi^0$, we train a trajectory planning model using behavior cloning (BC) using cross-entropy loss with trajectories from the source city.
The network architecture and training hyperparameters are detailed in \cref{sec:network} and \cref{sec:hyperparameters}, respectively.
We also train its counterpart in the target city for comparison. Additional baselines include random and constant velocity policies, along with an oracle from logged demonstrations.
We also report several diagnostic variants, including a variant with additional supervision beyond NOMAD.
During adaptation, we initialize the network backbone and the actor head of $\pi_\theta$ with $\pi^0$ but learn the critic head from scratch, and use $\lambda_{\mathrm{KL}}=0.08$ for all city pairs (a sensitivity study is provided in \cref{sec:klstudy}).
We set the interaction budget as 1 billion (72 NVIDIA A100 GPU hours) in all experiments to ensure a plateau of both the realism meta score and success rate.
During evaluation, all dynamic vehicles are controlled by our policy; results involving other sim agents are reported separately in \cref{sec:nonselfplay}.

%% file: src/results.tex
\begin{figure*}[htpb]
  \centering
  \vspace{-1.0\baselineskip}
  \includegraphics[width=\linewidth]{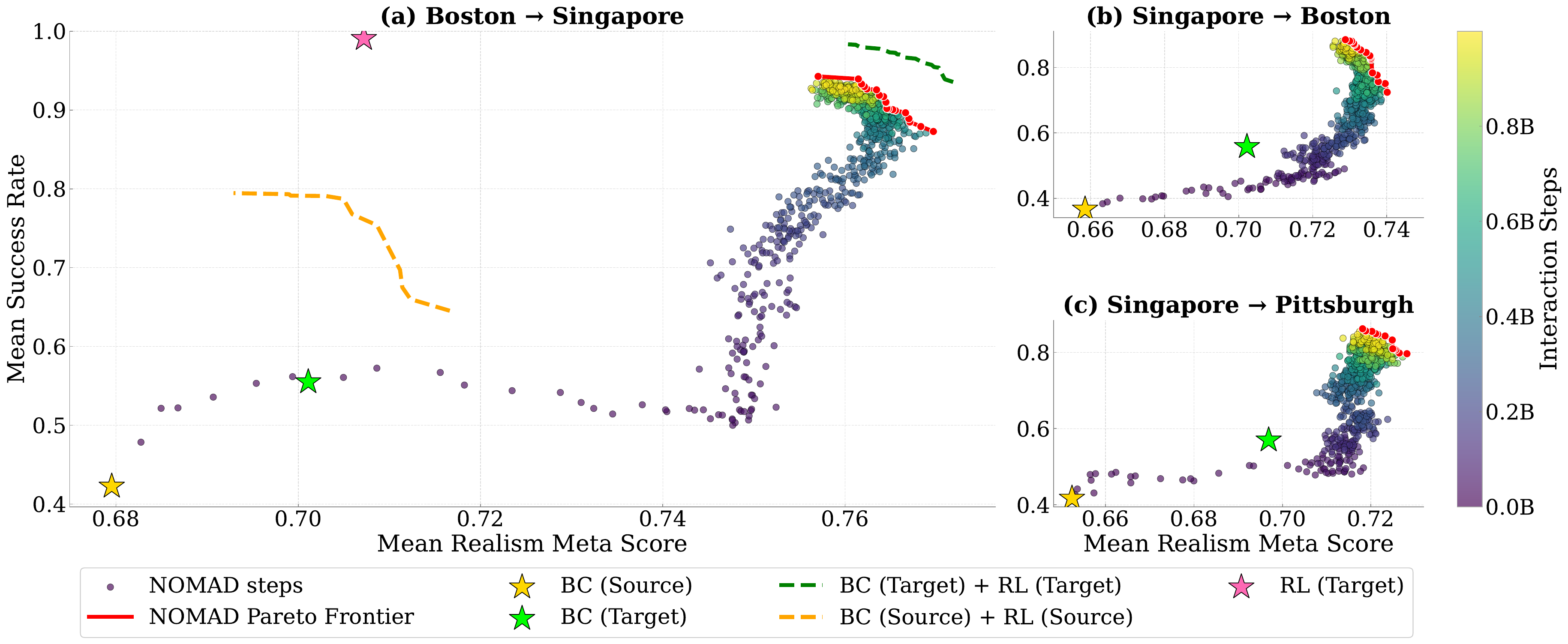}
  \vspace{-1.5\baselineskip}
  \caption{\textbf{NOMAD consistently improves the success--realism Pareto frontier in unseen target cities.}
    We plot mean success rate versus mean realism meta score over 5 runs for three transfer settings:
    (a) Boston-to-Singapore (primary), (b) Singapore-to-Boston, and (c) Singapore-to-Pittsburgh.
    Each dot is a NOMAD training checkpoint, colored by the cumulative number of interaction steps; the red dashed curve denotes the empirical Pareto frontier over NOMAD checkpoints.
    Stars denote BC reference policies, while dashed curves denote RL-based ablations initialized or trained under different data/settings.
    }
    
  \vspace{-1.0\baselineskip}
  \label{fig:frontier}
\end{figure*}

\vspace{-0.8\baselineskip}
\section{Results}\label{sec:results}
\vspace{-0.6\baselineskip}

\subsection{Main Results}
\vspace{-0.5\baselineskip}

\cref{fig:frontier} (a) visualizes the empirical Pareto frontier of success rate versus realism meta score across different training checkpoints throughout self-play training in Boston-to-Singapore transfer (training curves are provided in \cref{fig:realismvssuccess}). 
Each point represents an adapted policy (checkpoint) $\pi^+\in\Pi^+(C)$ with the color shade indicating the number of interaction steps since the beginning of training. 
The yellow star denotes $\pi^0$, while the lime star shows behavior cloning using Singapore data.
The modest performance of behavior cloning with 3,200 Singapore scenarios reflects the inherent data demands of imitation learning: empirical scaling analysis in autonomous driving shows that imitation-based policies are quite data hungry, with robustness and generalization improving only with substantially larger and more diverse datasets, while compounding errors dominate at smaller scales \citep{baniodeh2025scaling}.
By contrast, NOMAD achieves better success–realism trade-offs even without demonstrations in the target city, indicating that map-based self-play can be a more effective adaptation strategy than imitation under this target-data budget.



\begin{wraptable}{r}{0.5\linewidth}
\vspace{-0.0\baselineskip}
\caption{
\textbf{NOMAD substantially improves both success and realism over zero-shot transfer and imitation baselines.}
We report success rate and realism meta score for representative policies in closed-loop Singapore evaluation.
$\pi^{\text{expert}}_d$ denotes a policy obtained by inferring actions from logged expert trajectories and discretizing them to match the action space; its realism meta score serves as an approximate upper bound under the current POSG formulation.}
\vspace{-0.3\baselineskip}
\centering
\footnotesize
\begin{tabular}{@{}lcc@{}}
\toprule
\textbf{Policy} & \textbf{Realism} & \textbf{Success} \\ \midrule
$\pi^{\text{expert}}_d$ & \textbf{0.8056} & \textbf{88.26\%} \\ \midrule
Random               & 0.4074 & 4.30\%  \\
Constant Velocity    & 0.6147 & 19.45\% \\
$\pi^0$              & 0.6795 & 42.25\% \\
BC (Singapore)       & 0.7011 & 55.49\% \\ \midrule
NOMAD Frontier       & \textbf{0.7570$\sim$0.7697} & \textbf{87.28$\sim$94.27\%} \\ \bottomrule
\end{tabular}
\label{tab:baselines}
\vspace{-1.5\baselineskip}
\end{wraptable}
Interestingly, the resulting Pareto frontier forms a clear upper envelope and dominates both the zero-shot transfer policy $\pi^0$ and behavior cloning trained on 3,200 Singapore scenarios. 
This dominance spans the entire spectrum of success--realism trade-offs, which means that regardless of the desired balance between these two objectives, NOMAD offers a policy that outperforms both baselines in both metrics simultaneously. 
This means that practitioners may select a deployment policy $\pi^{\text{deploy}}$ anywhere along this frontier, from settings that prioritize realism to settings that maximize task completion, while still benefiting from substantial improvements over zero-shot transfer.

Notably, the observed trade-off between success rate and realism is mild: substantial gains in success rate are achieved with only marginal reductions in realism (a $7$-percentage-point gain in success rate corresponds to only a $0.012$ decrease in realism). 
Therefore, in practice, one can simply select the checkpoint with the highest success rate, \emph{without requiring target-city trajectories to estimate realism meta scores}.
Sub-metric results can be found in \cref{fig:detailed_metrics_boston_to_sin}.
In addition, we conduct a hold-out evaluation in \cref{sec:holdouteval} to assess potential selection bias, i.e., overestimation when constructing the empirical Pareto frontier.

To ground the frontier visualization with concrete values, \cref{tab:baselines} reports the success rate and realism meta score of representative baselines, the oracle policy, and the range spanned by the NOMAD Pareto frontier.
NOMAD consistently achieves superior success–realism trade-offs compared to zero-shot transfer and other baselines, approaching the expert-derived realism upper bound while exceeding it in task success without access to any target-city demonstrations.


\vspace{-0.5\baselineskip}
\subsection{The Role of Behavioral Priors} \label{sec:pureselfplay}
\vspace{-0.5\baselineskip}
NOMAD demonstrates that map-based self-play can effectively adapt a driving policy to a new city using only a simple reward function, which raises a natural question: \emph{is an imitation policy necessary to initialize and regularize policy learning?}
To answer this, we conduct an ablation study in which we train the policy purely via map-based self-play in the target city without regularization from scratch, using the same reward function and training budget.
``RL (Target)'' in \cref{fig:frontier} shows the performance of the best checkpoint (training curves are provided in \cref{fig:pureselfplay}).
Without an imitation policy, self-play can indeed achieve near-perfect success rates and improved realism meta score, compared to zero-shot transfer and behavior cloning policies in the target city.
However, its realism plateaus at a notably lower value. 
By contrast, NOMAD improves the success rate while maintaining a high realism meta score, ultimately achieving a substantially better success--realism trade-off.

\begin{wrapfigure}{r}{0.5\linewidth}
  \centering
  \vspace{-1.0\baselineskip}
  \includegraphics[width=\linewidth]{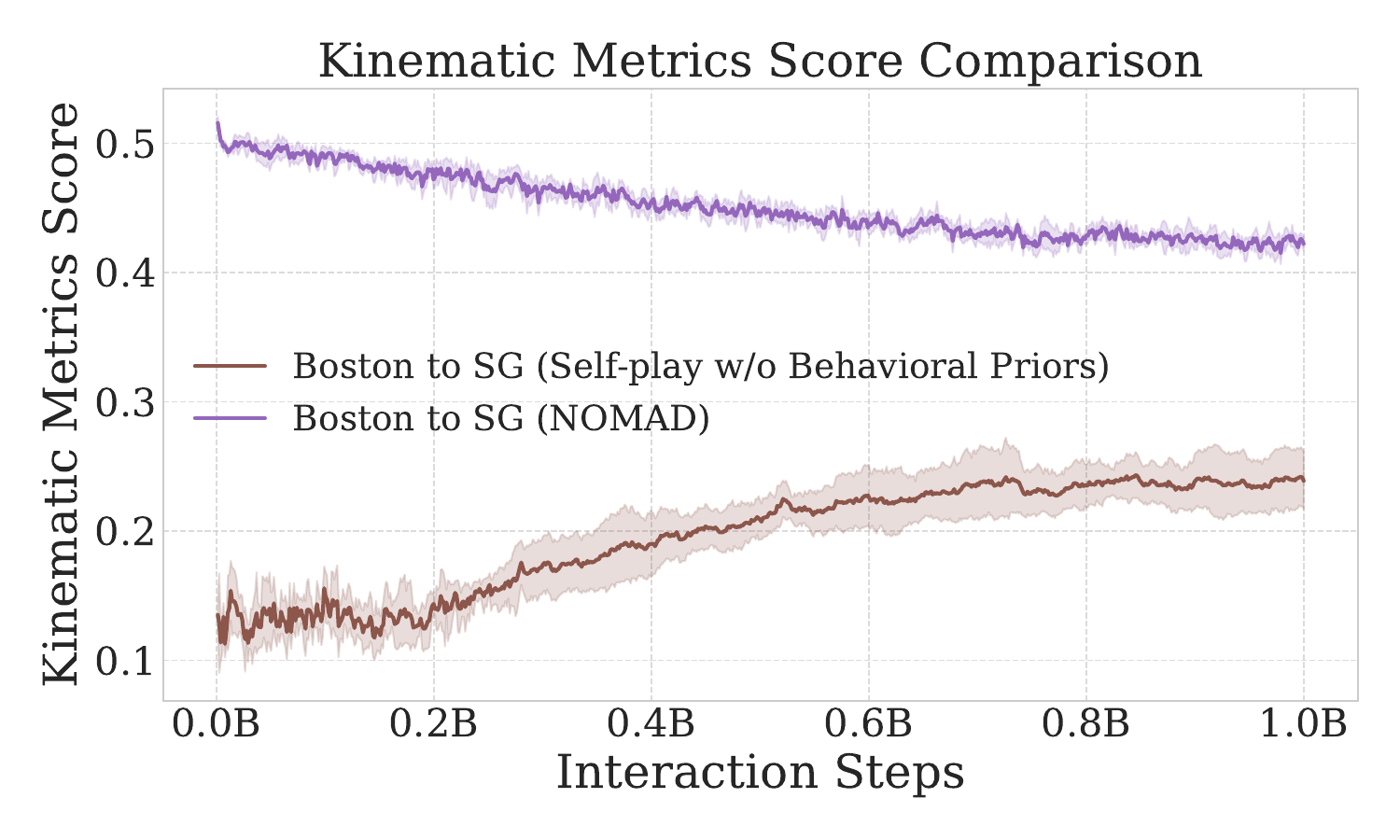}
  \vspace{-1.0\baselineskip}
  \caption{\textbf{Behavioral priors are critical for preserving kinematically realistic driving during self-play.}
  We compare the kinematic realism score of NOMAD against self-play from scratch in the Boston-to-Singapore transfer setting.
  Self-play without behavioral priors struggles to learn kinematically realistic behavior, while NOMAD preserves substantially more realistic motion patterns despite lacking explicit kinematic rewards.
  }
  \vspace{-0.5\baselineskip}
  \label{fig:pureselfplay_kinematics}
\end{wrapfigure}

To understand the source of this gap, we compare the kinematic metrics between these two methods in \cref{fig:pureselfplay_kinematics} (other metrics are compared in \cref{fig:detailed_metrics_comparison}).
Self-play without pretraining and regularization fails to achieve high kinematic scores.
While NOMAD exhibits a mild decline in kinematic realism due to the absence of explicit realism rewards, its initialization from behavioral priors provides a superior starting point compared to training from scratch. 

Furthermore, by anchoring the self-play process to human-like prior, NOMAD maintains significantly higher kinematic scores throughout the adaptation process.
These results highlight that self-play alone is insufficient for learning realistic driving behavior under minimal reward supervision.
The pretrained BC policy both serves as an optimization warm-start and provides critical behavioral priors that constrain exploration to human driving patterns and mitigate reward hacking.
This enables NOMAD to achieve superior realism without sacrificing task success.


\vspace{-0.5\baselineskip}
\subsection{The Necessity of the Target-City Map}
\vspace{-0.5\baselineskip}
To investigate the role of the target-city map in the adaptation process, we conduct behavior cloning followed by self-play with logged scenarios entirely in the \textit{source} city (Boston) and evaluate the resulting policy zero-shot in the target city (Singapore).
This baseline uses the same BC initialization and self-play training protocol as NOMAD, but never accesses the target-city map or scenarios during training.
The ``BC (Source) + RL (Source)'' curve in \cref{fig:frontier} shows the Pareto frontier of source-city self-play (detailed results are provided in \cref{fig:bos_to_bos_frontier}).
While this baseline improves upon the zero-shot transfer policy, its Pareto frontier remains significantly inferior to that of NOMAD. Specifically, self-play conducted in Boston saturates at substantially lower success rates and yields only marginal improvements in realism when evaluated in Singapore.


These findings provide strong justification for map-based self-play in the \emph{target} city. Self-play alone is not a silver bullet: improvements learned through interaction are largely city-specific and do not reliably transfer across distinct urban environments.
Effective adaptation requires interaction dynamics to be grounded in the geometry, topology, and traffic structure of the deployment city, rather than relying solely on additional optimization in the source city.


\vspace{-0.5\baselineskip}
\subsection{NOMAD vs. Demonstration-Based Training}
\vspace{-0.5\baselineskip}
We further examine how closely NOMAD can approach the performance of a policy trained with access to target-city demonstrations.
Specifically, we compare NOMAD, which uses neither target-city demonstrations nor logged scenarios, against a data-driven policy trained with access to 3,200 scenarios with human driving trajectories from Singapore.
This policy is first pretrained via BC and subsequently trained using self-play on both generated and logged scenarios.
The reward function, training budget, and overall training protocol are identical to those of NOMAD, with the only difference being that all training is conducted directly using target-city demonstrations.

This policy, denoted as ``BC (Target) + RL (Target)'' in \cref{fig:frontier} (full results in \cref{fig:sin_to_sin_frontier}), achieves only a modest improvement over NOMAD along the success–realism Pareto frontier.
Notably, the performance gap between this policy and NOMAD is substantially smaller than the gap between NOMAD and zero-shot transfer.
This indicates that map-based self-play provides the majority of the gains typically provided by target-city demonstrations.

\vspace{-0.5\baselineskip}
\subsection{Generalization to Other Cities}\label{sec:othercities}
\vspace{-0.5\baselineskip}
To verify that the effectiveness of NOMAD is not specific to a single target city, we evaluate cross-city transfer from Singapore to both Boston and Pittsburgh. 
\cref{fig:frontier} (b) and (c) report the success–realism Pareto frontiers for these two transfer settings, respectively.
These results demonstrate that NOMAD consistently delivers significant performance gains across different target cities. 
In all cases, the adapted policies substantially outperform the zero-shot baseline, despite the absence of any demonstrations from Boston or Pittsburgh. 
In particular, this generalization is achieved using a single reward function and a shared set of hyperparameters (reported in \cref{sec:hyperparameters}) across all target cities, highlighting the robustness and scalability of NOMAD. 
It is worth noting that realism scores are city-dependent (discussed in \cref{sec:wosaccitydependent}), and should not be compared directly across cities.
Some qualitative results are provided in \cref{sec:qualitativeresults}.


\vspace{-0.0\baselineskip}
\subsection{Performance Gain Analysis} \label{sec:failurebreakdown}
\vspace{-0.5\baselineskip}

To understand the source of NOMAD's success rate improvement, we break down the failed episodes by their cause.
An episode fails if the vehicle does not reach its goal within the horizon $H$, and we categorize each failure by whether the vehicle drove off-road, collided with another vehicle, both, or neither (i.e., it simply timed out).
\cref{tab:failure_modes} reports the results for the zero-shot policy $\pi^0$ and NOMAD in the Boston-to-Singapore and Singapore-to-Boston transfer settings.
In Boston-to-Singapore transfer, NOMAD reduces the overall failure rate from 57.7\% to 7.7\%, off-road-involved failures (off-road only and both) from 36.0\% to 0.9\%, collision-involved failures (collision only and both) from 3.9\% to 0.7\%, and timeout-only failures from 20.5\% to 6.5\%.
The same trend holds in reverse transfer.
Thus, the largest gain is in map compliance, which is consistent with our motivation that road geometry and topology differ substantially across cities, while collision-involved failures also decrease substantially.

\begin{table}[htbp]
\vspace{-0.5\baselineskip}
\caption{\textbf{NOMAD eliminates most failures in the target city.}
We report the failure rate (\%) and its breakdown by failure mode in closed-loop evaluation for Boston-to-Singapore and Singapore-to-Boston transfer.
NOMAD results are reported as mean $\pm$ standard deviation over 5 runs.}
\centering
\footnotesize
\setlength{\tabcolsep}{4pt}
\begin{tabular}{@{}lcccccc@{}}
\toprule
\textbf{Transfer} & \textbf{Policy} & \textbf{Failure} & \textbf{Off-road only} & \textbf{Collision only} & \textbf{Both} & \textbf{Neither (timeout)} \\ \midrule
\multirow{2}{*}{Boston $\to$ Singapore} & $\pi^0$ & 57.7\% & 33.3\% & 1.2\% & 2.7\% & 20.5\% \\
 & NOMAD & \textbf{7.7\% $\pm$ 1.8} & \textbf{0.6\% $\pm$ 0.2} & \textbf{0.4\% $\pm$ 0.2} & \textbf{0.3\% $\pm$ 0.3} & \textbf{6.5\% $\pm$ 1.7} \\ \midrule
\multirow{2}{*}{Singapore $\to$ Boston} & $\pi^0$ & 63.3\% & 18.8\% & 3.1\% & 8.1\% & 33.3\% \\
 & NOMAD & \textbf{13.6\% $\pm$ 5.6} & \textbf{1.3\% $\pm$ 0.3} & \textbf{1.5\% $\pm$ 0.5} & \textbf{1.1\% $\pm$ 0.3} & \textbf{9.7\% $\pm$ 4.3} \\ \bottomrule
\end{tabular}
\label{tab:failure_modes}
\vspace{-1.0\baselineskip}
\end{table}

\subsection{\texorpdfstring{{\rev Qualitative Analysis}}{Qualitative Analysis}} \label{sec:qualitativemain}

\begin{figure}[htbp]
  \centering
  \includegraphics[width=\linewidth]{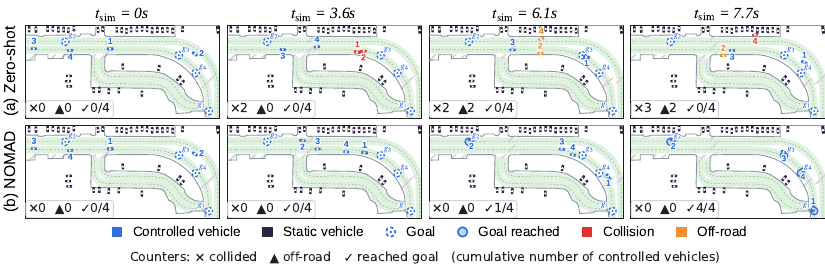}
  \caption{\rev\textbf{Qualitative comparison of the zero-shot policy and NOMAD under Singapore-to-Boston transfer.}
  Both rows start from the same scenario and initial state.
  Numbers identify the controlled vehicles and $g_i$ marks the goal of vehicle $i$; a controlled vehicle is drawn in \textcolor[HTML]{E03131}{red} while it is in collision and in \textcolor[HTML]{F28E2B}{orange} while it is off-road.
  More examples are in \cref{sec:qualitativeresults}.}
  \label{fig:qual_main}
  \vspace{-0.5\baselineskip}
\end{figure}

{\rev We complement the quantitative results with a qualitative analysis of representative transfer cases. \cref{fig:qual_main} shows the visualizations of two rollouts from $\pi^0$ and NOMAD respectively for a Singapore-to-Boston case,  in which the failures of $\pi^0$ arise from driving-side bias.
Trained under left-hand traffic, the zero-shot policy keeps driving on the left in Boston: vehicle 2 drifts into the lower lane, which belongs to oncoming traffic, and collides head-on with vehicle 1 at $t_{\mathrm{sim}}=3.6$\,s. Vehicle 4 drifts to its left, and by $t_{\mathrm{sim}}=6.1$\,s, vehicles 2 and 4 have both left the road.
By the end of the episode, three of the four controlled vehicles have collided, two have gone off-road, and none has reached its goal.
Under NOMAD, vehicle 2 moves to the upper lane, on its right-hand side, and all vehicles keep to the right through the curve, so all four reach their goals without collision or off-road events.
Additional examples for both transfer directions, covering narrow lanes, cluttered static obstacles, and multi-vehicle interactions, are given in \cref{sec:qualitativeresults}.}



%% file: src/conclusion.tex
\vspace{-0.5\baselineskip}
\section{Conclusion}
\vspace{-0.5\baselineskip}
This paper challenges a central assumption in city transfer in autonomous driving: that effective adaptation to a new city requires collecting human demonstrations in that city.
We introduce NOMAD, a framework that adapts autonomous-driving policies to new cities using only the target-city map and meta-information, without requiring target-city human demonstrations for adaptation.
Through extensive closed-loop evaluations, we show that NOMAD consistently expands the success--realism Pareto frontier, transforming brittle zero-shot transfer into a diverse set of policies spanning different trade-offs between task success and behavioral realism.
These results suggest that much of the disparity in optimal driving behavior across cities 
can be effectively addressed through map-based self-play.
As traffic simulation continues to improve, we believe that scalable multi-agent self-play offers a viable and principled path toward robust, large-scale deployment of autonomous driving.

\vspace{-0.0\baselineskip}
\paragraph{Limitations.}\label{para:limitations}
NOMAD uses lane-level maps to generate training scenarios, which makes map coverage and scenario representativeness important for adaptation.
Incomplete map annotations can leave some configurations under-represented during training, including curbside parking and unmapped parking or access areas as shown in \cref{sec:failurecasestudy}. 
Secondly, differences in local driving culture and social conventions remain challenging to capture through map structure alone, suggesting the need for richer representations of local interaction norms beyond geometric and traffic-rule information. Thirdly, our simulation abstracts perception and low-level vehicle dynamics through a delta-pose action space, as is standard in closed-loop planning benchmarks; validating NOMAD in higher-fidelity simulators is an important complementary direction. Therefore, policies adapted in simulation may still fail under real-world long-tail events, imperfect maps, or local driving conventions, and our results should be interpreted as a cost-effective tool for policy adaptation rather than as a claim of deployment readiness. Any real-world use would require further validation, safety monitoring, and compliance with local traffic regulations.

%% file: src/impact.tex
\vspace{-0.0\baselineskip}
\paragraph{Impact Statement.} This work aims to reduce the cost of adapting autonomous-driving policies to new cities by reducing the dependence on target-city demonstrations.
Its potential positive impact is to make adaptation more scalable and less costly across geographically diverse environments.
We do not identify direct negative societal impacts specific to this work.

%% file: src/appendices.tex
\crefalias{section}{appendix}
\addtocontents{toc}{\protect\setcounter{tocdepth}{2}}

\begingroup
  \makeatletter
  \let\old@starttoc\@starttoc
  \def\@starttoc#1{\begingroup\parskip=2pt\old@starttoc{#1}\endgroup}
  \makeatother
  \tableofcontents
\endgroup

\appendix

\clearpage
\input{appendices/NetworkArchitecture}
\clearpage
\input{appendices/DetailedExperimentalSetup}
\clearpage
\input{appendices/SceneGeneration}

\clearpage
\input{appendices/TrafficDensitySensitivity}
\clearpage
\input{appendices/VisualizationofScenarios}
\clearpage
\input{appendices/DetailedMetrics}

\clearpage
\input{appendices/ExtendedMetricAnalysis}
\clearpage
\input{appendices/AdditionalResults}
\clearpage
\input{appendices/EvalNonSelfplay}
\clearpage
\input{appendices/SameSideTransfer}
\clearpage
\input{appendices/DrivingSide}
\clearpage
\input{appendices/HoldoutEval}
\clearpage
\input{appendices/FailureCases}
\clearpage
\input{appendices/Qualitative}

%% file: appendices/NetworkArchitecture.tex
\section{Network Architecture and Observation Visualization} \label{sec:network}

\subsection{Policy Network Architecture}

\begin{figure}[htpb]
  \centering
  \includegraphics[width=0.75\linewidth]{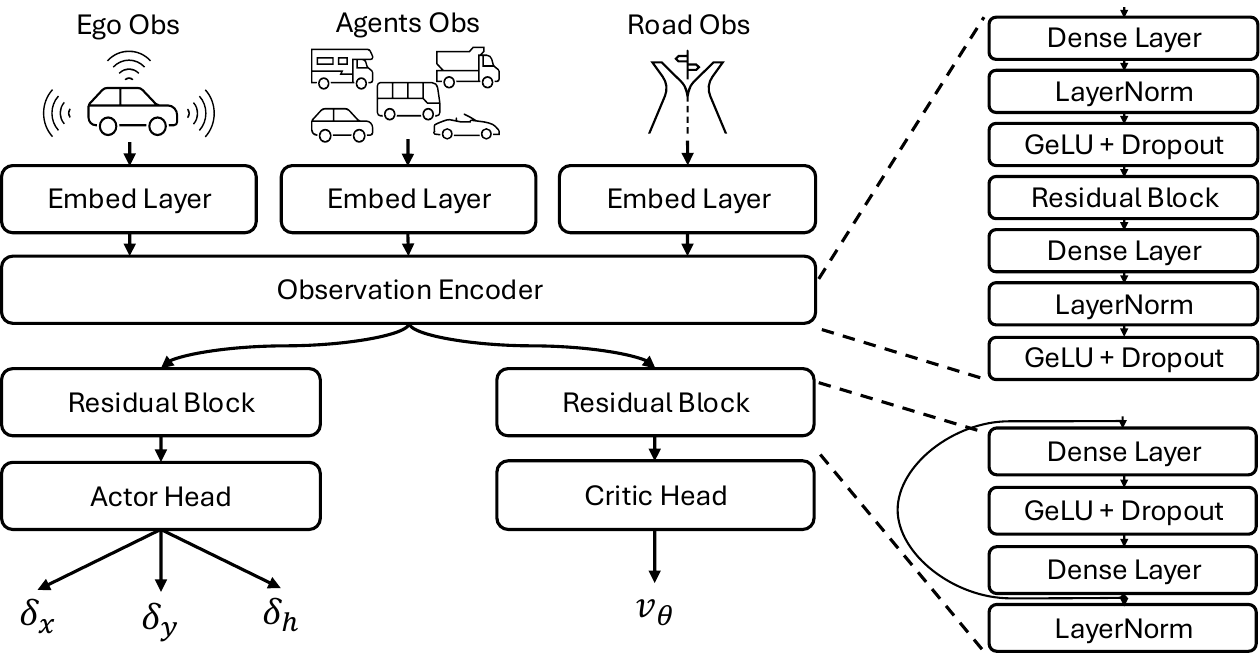}
  \caption{\textbf{Policy Network architecture.} The policy consists of modality-specific embedding layers for ego state, surrounding agents, and road observations, followed by a shared observation encoder. The encoded features are passed to separate actor and critic branches with residual blocks. The actor outputs trajectory increments $(\delta_x, \delta_y, \delta_h)$, while the critic predicts the scalar state value $v_\theta$.
  }
  \label{fig:networkarchitecture}
\end{figure}

We adopt a shared actor–critic architecture with a modular observation encoder, as shown in \cref{fig:networkarchitecture}. 
At each timestep, the policy receives three structured inputs: ego-centric state features, observations of surrounding agents (illustrated in \cref{fig:obs_space}), and road-level information derived from the map. 
Each modality is first processed by a dedicated embedding layer, and the resulting embeddings are fused by a shared observation encoder composed of stacked fully connected layers with residual connections. 
The encoded representation is then passed to two separate branches. 
The actor branch outputs incremental trajectory commands $(\delta_x, \delta_y, \delta_h)$, representing relative position and heading updates, while the critic branch predicts a scalar value estimate $v_\theta$. 
Both branches employ residual blocks with normalization and dropout to stabilize training and improve generalization across cities.

\subsection{Visualization of Observation}
\begin{figure}[htpb]
    \centering
    \includegraphics[width=0.5\textwidth]{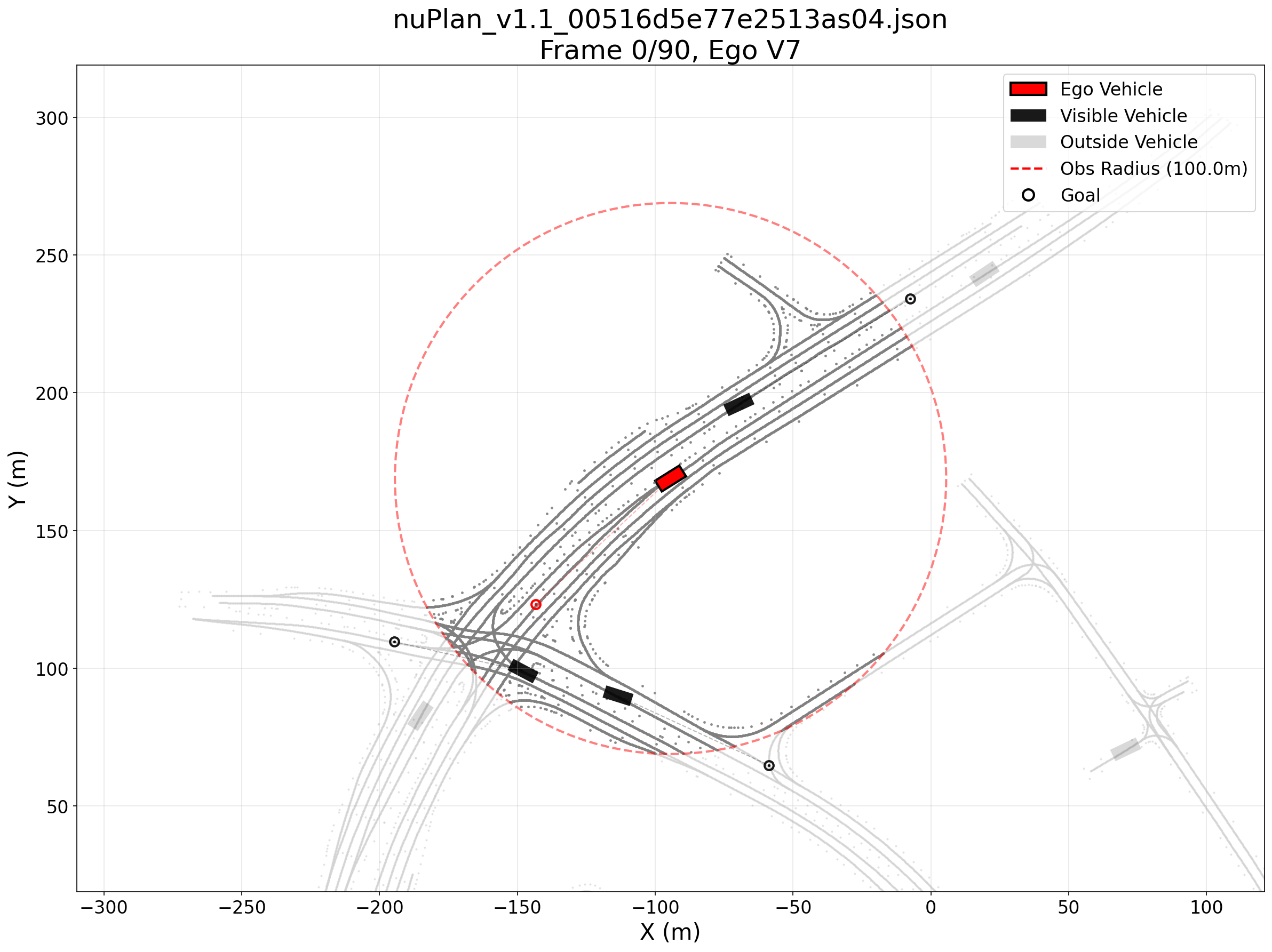}
    \caption{Visualization of the ego-centric observation space.}
    \label{fig:obs_space}
\end{figure}

We provide an example of observation of our agents in \cref{fig:obs_space}.

%% file: appendices/DetailedExperimentalSetup.tex
\section{Detailed Experimental Setup}\label{sec:hyperparameters}
The hyperparameters of the network used in the experiments are shown in \cref{tab:networkhyperparameter}. 

\begin{table}[htbp]
\centering
\caption{Network hyperparameters.}
\label{tab:networkhyperparameter}
\begin{tabularx}{0.5\linewidth}{@{}>{\raggedright\arraybackslash}X r@{}}
\toprule
\textbf{Hyperparameter} & \textbf{Value} \\
\midrule
Observation Dim & 2984 \\
Observation Embedding Dim & 64 $\times$ 3 \\
Encoder Width & 512 \\
Dropout & 0.01 \\
Residual Block Num & 1 \\
Projection Width & 256 \\
Pre-head Width & 256 \\
\bottomrule
\end{tabularx}
\end{table}

The detailed setting of our environment is shown in \cref{tab:environmenthyperparameter}.
We deployed a discretized action space.
This design is motivated by these considerations:
First, continuous actions introduce additional optimization challenges under hard constraints, e.g., prediction--execution mismatch caused by clipping or squashing in continuous RL algorithms.
In addition, discretization offers multi-modal exploration and stable policy gradient estimation.
And for imitation learning, a common practice in autonomous driving, it mitigates averaging in continuous regression and improves imitation of multi-modal expert behavior.
Furthermore, empirically, compared to inverse bicycle models, delta-pose actions enable better replay of expert trajectories, leading to a higher success rate of policy with inferred expert actions.
However, since this abstraction bypasses vehicle dynamics and actuation constraints, it may induce lower kinematic realism.

The granularity of the discretization induces a trade-off between imitation fidelity and RL exploration.
With coarse bins, the RL policy can still reach a high success rate in the target city, but the large quantization error makes it difficult to accurately reproduce demonstrated actions, substantially degrading the source-city behavior cloning policy and its trajectory realism.
Conversely, very fine bins reduce the quantization error but make exploration considerably more difficult, as the effective action space grows rapidly.
We use 256 bins for each of the three action dimensions $(\delta_x, \delta_y, \delta_h)$, i.e., a multi-discrete action space of size $256^3$, as a compromise between behavior cloning quality and exploration difficulty.

\begin{table}[htbp]
\centering
\caption{Environment configurations.}
\label{tab:environmenthyperparameter}
\begin{tabularx}{0.5\linewidth}{@{}>{\raggedright\arraybackslash}X r@{}}
\toprule
\textbf{Hyperparameter}     & \textbf{Value}               \\ \midrule
Observation Radius          & 100.0                         \\
Max Controllable Agents Num & 64                           \\
Max Road Observation Num    & 200                          \\
Ego observation             & 1$\times$6      \\
Agents Observation          & (64-1)$\times$6 \\
Road Observation            & 200$\times$13   \\
Action Space Type           & Multi-Discrete               \\
$\delta_x$            & {[}-2.0,2.0,256{]}               \\
$\delta_y$             & {[}-2.0,2.0,256{]}               \\
$\delta_h$             & {[}-3.14/4, 3.14/4, 256{]}        \\
Action Space Size           & 256$^\text{3}$       \\
Goal Reaching Reward        & 1.0                          \\
Collision Reward            & -0.75                        \\
Offroad Reward              & -0.75                        \\
Goal Reaching Behavior          & Disappear                       \\
Collision Behavior          & Ignore                       \\
Offroad Behavior          & Ignore                       \\
Goal Distance Threshold     & 2.0                          \\
Initial Step                & 0                            \\
Horizon                     & 90                           \\ \bottomrule
\end{tabularx}
\end{table}

Hyperparameters of behavior cloning and PPO are shown in \cref{tab:bchyperparameter} and \cref{tab:ppohyperparameter}, respectively.

\begin{table}[htbp]
\centering
\caption{Behavior Cloning hyperparameters.}
\label{tab:bchyperparameter}
\begin{tabularx}{0.5\linewidth}{@{}>{\raggedright\arraybackslash}X r@{}}
\toprule
\textbf{Hyperparameters} & \textbf{Value}   \\ \midrule
Dataset Scenario Num     & 3200             \\
Batch Size               & 64               \\
Loss Function            & Cross Entropy    \\
Learning Rate            & 1e-3             \\
Weight Decay             & 1e-2             \\
Scheduler                & Cosine Annealing \\ \bottomrule
\end{tabularx}
\end{table}

\begin{table}[htbp]
\centering
\caption{PPO hyperparameters.}
\label{tab:ppohyperparameter}
\begin{tabularx}{0.5\linewidth}{@{}>{\raggedright\arraybackslash}X r@{}}
\toprule
\textbf{Hyperparameters} & \textbf{Value} \\ \midrule
World Num                & 1,000          \\
Batch Size               & 8,192          \\
Buffer Size              & 131,072        \\
Update Epoch             & 4              \\
Discount Factor $\gamma$                 & 0.99           \\
GAE Discount Factor $\lambda$            & 0.95           \\
Advantage Normalization           & True           \\
Clip Coefficient $\epsilon$         & 0.2            \\
Value Loss Weight        & 0.3            \\
Entropy Loss Weight      & 1e-4           \\
KL Weight                & 0.08           \\
Max Gradient Norm        & 0.5            \\
Total Timestep           & 1 billion      \\ \bottomrule
\end{tabularx}
\end{table}

%% file: appendices/SceneGeneration.tex
\section{Scenario Generation} \label{sec:scenegen}

\subsection{Scenario Generation Algorithm}

Given a lane-level map segment $m$ from the target city, our simple
heuristic scenario generator $\Xi_\varphi(\cdot|m, I_C)$ produces a
traffic scenario $\xi = (s_0, \{g^i\}_{i=0}^{N-1}, m)$. We emphasize that the utilized generator in this study is intentionally minimal, and any alternative generator satisfying basic feasibility constraints would be expected to work similarly (as evidenced by the density robustness results in Appendix\ref{sec:sensitivitytoerrorsintrafficdensity}). A generated
scenario only specifies the initial conditions: the initial state
$s_0=\{(p^i, \psi^i, v^i)\}_{i=0}^{N-1}$, where each vehicle $i$ is
characterized by its initial position $p^i \in \mathbb{R}^2$, heading
$\psi^i$, and speed $v^i$, along with its goal position
$g^i \in \mathbb{R}^2$. The trajectory connecting $s_0$ to the goals is
produced by the learned policy during simulation. The generation process
consists of four phases.

\textbf{First}, we extract lane centerlines $\mathcal{L}$ from the map
and construct a directed lane connectivity graph $G$. \textbf{Second},
we determine the number of vehicles to be generated by sampling a traffic density $\rho$ (vehicles per kilometer) from a
city-specific Gamma distribution, $\rho \sim \mathrm{Gamma}(\alpha,
\theta) + \rho_0$, which can be obtained from urban traffic sensors or
aerial imagery. We then set $N_{\mathrm{dyn}} = \max(1,\lfloor \rho
\cdot \sum_{\ell \in \mathcal{L}} \mathrm{len}(\ell) / 1000 \rceil)$ and
optionally add static (parked) vehicles with probability
$p_{\mathrm{static}}$. \textbf{Third}, we place vehicles on the lane
network using rejection sampling with collision avoidance: each
candidate samples a lane segment with probability proportional to its
length, computes the centerline position and heading, and applies small
Gaussian perturbations for lateral offset and heading. Dynamic vehicles
receive speeds drawn independently from a LogNormal distribution scaled by the
speed limit of the map segment, while static vehicles are assigned $v^i = 0$;
in both cases we enforce a minimum safe distance
$d_{\mathrm{safe}}=\max(v \cdot \Delta t_{\mathrm{gap}}, d_{\min})$.
\textbf{Fourth}, we generate goals by traversing $G$. For static
vehicles, $g^i=p^i$. For dynamic vehicles, we sample a travel distance
$d_{\mathrm{goal}} \sim \mathcal{N}(\kappa_\mu v^i, \kappa_\sigma v^i)$,
clip it to $[d_{\min}^g, d_{\max}^g]$, and advance along successor lanes
until the distance is exhausted or a dead-end is reached. Small Gaussian
noise is added to the final goal for robustness. 

\begin{algorithm}[b!]
\caption{Heuristic Scenario Generation $\Xi_\varphi(\cdot|m, I_C)$}
\label{alg:scene_generation}
\begin{algorithmic}[1]
\Require Map segment $m$, city meta-information $I_C$, parameters $\varphi$
\Ensure Traffic scenario $\xi = (s_0, \{g^i\}_{i=0}^{N-1}, m)$
\State $\mathcal{L}, G \gets \textsc{ExtractLaneNetwork}(m)$
\State $\rho \sim \mathrm{Gamma}(\alpha, \theta) + \rho_0$
   
\State $N_{\mathrm{dyn}} \gets \max(1, \lfloor \rho \cdot
   \mathrm{TotalLaneLength}(\mathcal{L}) / 1000 \rceil)$
\State $N_{\mathrm{static}} \gets
   \textsc{SampleStaticCount}(N_{\mathrm{dyn}}, p_{\mathrm{static}})$
\State $\mathcal{V} \gets \emptyset$
\For{$i = 1$ \textbf{to} $N_{\mathrm{dyn}} + N_{\mathrm{static}}$}
    \State $(p, \psi) \gets \textsc{SampleLanePosition}(\mathcal{L})$
    \State $v \gets 0$ \textbf{if} $i > N_{\mathrm{dyn}}$
       \textbf{else} $\textsc{SampleSpeed}()$
    \If{$\textsc{CollisionFree}(p, v, \mathcal{V},
       \Delta t_{\mathrm{gap}}, d_{\min})$}
        \State $\mathcal{V} \gets \mathcal{V} \cup \{(p, \psi, v)\}$
    \EndIf
\EndFor
\For{\textbf{each} $(p^i, \psi^i, v^i) \in \mathcal{V}$}
    \State $d^i_{\mathrm{goal}} \sim
       \mathcal{N}(\kappa_\mu v^i, \kappa_\sigma v^i)$
    \State $g^i \gets p^i$ \textbf{if} $v^i = 0$ \textbf{else}
       $\textsc{AdvanceOnGraph}(G, p^i, d^i_{\mathrm{goal}})$
\EndFor
\State \Return $\xi = (s_0, \{g^i\}_{i=0}^{|\mathcal{V}|-1}, m)$
\end{algorithmic}
\end{algorithm}

\subsection{Parameter Specification}
\label{sec:scenegen_params}

For reproducibility, we report all parameters $\varphi$ used in the scenario generation algorithm. The city-specific traffic density parameters (\cref{tab:density_params}) were estimated by fitting a Gamma distribution to vehicle density statistics computed from {\rev the training split of} the nuPlan dataset for each target city using maximum likelihood estimation. While this uses aggregate density distributions from logged data, it does not use individual trajectory information. These statistics could alternatively be obtained from traffic sensors, aerial imagery, or simple visual observation, and no dataset access is required in principle.

\begin{table}[h!]
\centering
\caption{City-specific traffic density parameters for $\rho \sim \mathrm{Gamma}(\alpha, \theta) + \rho_0$ (vehicles/km), where $\alpha$ is the shape and $\theta$ is the scale parameter.}
\label{tab:density_params}
\begin{tabular}{lccc}
\toprule
City & $\alpha$ & $\rho_0$ & $\theta$ \\
\midrule
Boston & 1.4694 & 0.1897 & 0.4252 \\
Pittsburgh & 2.1327 & 0.0000 & 0.3982 \\
Singapore & 1.2630 & 0.1363 & 0.2500 \\
\bottomrule
\end{tabular}
\end{table}

The shared parameters for vehicle placement and goal generation are reported in \cref{tab:shared_params}. Dynamic vehicle speeds are sampled from a LogNormal distribution with mean $\bar{v} = 0.85 \cdot v_{\mathrm{max}}$ and standard deviation $\sigma_v = 0.22 \cdot v_{\mathrm{max}}$, where $v_{\mathrm{max}}$ is the speed limit of the map segment. Since the map data available to us does not include posted speed-limit annotations, we use the maximum observed initial speed in each map segment, among all logged agents, as a coarse proxy for its local speed limit; in practical deployment, posted per-segment speed limits would be a drop-in replacement for this proxy. This reflects typical free-flowing traffic behavior where vehicles operate below maximum speed~\citep{treiber2013traffic}. The collision avoidance parameters---minimum time gap $\Delta t_{\mathrm{gap}}$ and minimum distance $d_{\min}$---are consistent with calibrated values from the Intelligent Driver Model~\citep{treiber2000congested}. For goal generation, the distance coefficients $\kappa_\mu$ and $\kappa_\sigma$ determine how far ahead vehicles plan to travel based on their current speed: a vehicle traveling at speed $v$ receives a goal at distance $d_{\mathrm{goal}} \sim \mathcal{N}(\kappa_\mu v, \kappa_\sigma v)$, encouraging faster vehicles to have farther goals. Small Gaussian perturbations are added to lateral position, heading, and goal location for realism.

\begin{table}[h!]
\centering
\caption{Shared scenario generation parameters.}
\label{tab:shared_params}
\begin{tabular}{llc}
\toprule
Category & Parameter & Value \\
\midrule
\multirow{3}{*}{\textit{Vehicle Placement}} 
 & Minimum time gap $\Delta t_{\mathrm{gap}}$ & $1.2\,\mathrm{s}$ \\
 & Minimum distance $d_{\min}$ & $5.0\,\mathrm{m}$ \\
 & Static vehicle probability $p_{\mathrm{static}}$ & $0.5$ \\
\midrule
\multirow{2}{*}{\textit{Speed Distribution}} 
 & Mean speed fraction $\bar{v}/v_{\max}$ & $0.85$ \\
 & Speed std fraction $\sigma_v/v_{\max}$ & $0.22$ \\
\midrule
\multirow{3}{*}{\textit{Goal Generation}} 
 & Distance mean coefficient $\kappa_\mu$ & $9.0$ \\
 & Distance std coefficient $\kappa_\sigma$ & $3.0$ \\
 & Goal distance range $[d_{\min}^g, d_{\max}^g]$ & $[5, 200]\,\mathrm{m}$ \\
\bottomrule
\end{tabular}
\end{table}

\subsection{Local vs.\ City-Level Meta-Information}
\label{sec:scenegen_local}

The meta-information $\mathcal{I}_C$ combines local and city-level components.
\textbf{Speed} is handled locally: the speed limit is specified per map segment, so highways and inner-city streets receive different speed scales.
Moreover, speeds are not fixed even within a scenario, since the initial speed of each vehicle is drawn independently from the LogNormal distribution above, yielding heterogeneous per-agent speeds around the local limit.

\textbf{Traffic density} is given only as a coarse city-level prior, which we deliberately adopt as the weakest and cheapest form of meta-information, as target cities without fine-grained data are exactly where adaptation is most needed.
This treatment is nevertheless not uniform.
First, a density is sampled per scenario from the city-level Gamma distribution, so training spans sparse to dense conditions, reflecting that each map segment may experience different densities at different times of a day or year; the number of vehicles further scales with the total lane length of each segment.
Second, unlike speed, the feasible density per lane-kilometer is physically bounded by vehicle lengths and safe headways~\citep{treiber2013traffic} (enforced by $d_{\mathrm{safe}}$ in our generator), so its plausible range varies far less across a city; a coarse city-level distribution is therefore a reasonable prior in a way it would not be for speed.
Third, \cref{sec:sensitivitytoerrorsintrafficdensity} shows that the success--realism frontier is nearly unchanged under moderate density misspecification, as the density prior shapes only initial states and goals, and never supervision or reward.
If per-region density information (e.g., from traffic sensors) is available, the generator accepts it as a drop-in replacement, and our results can be viewed as a lower bound on what such richer meta-information would yield.

\subsection{\texorpdfstring{{\rev Training and Evaluation Scenario Statistics}}{Training and Evaluation Scenario Statistics}}
\label{sec:scenario_stats}
{\rev
NOMAD is trained only on generated scenes and evaluated on held-out logged scenarios, so a natural concern is whether the training scenes were constructed to resemble the evaluation traffic.
To address it, this section describes how the training and evaluation data are partitioned and compares their road geometry and traffic statistics.
We also test whether the training and evaluation traffic can be told apart: we train a binary classifier (gradient-boosted trees, scikit-learn \texttt{HistGradientBoostingClassifier}) on the standardized traffic features defined below to separate 800 randomly drawn scenes of a set from the 800 evaluation scenes, and report ROC-AUC under 5-fold stratified cross-validation, averaged over 5 random draws and fold splits; an AUC of 0.5 means the two distributions cannot be told apart.

\paragraph{Data partition.}
We report statistics for Boston$\rightarrow$Singapore, our main transfer setting.
We randomly partition the nuPlan Singapore scenarios into a training split (3{,}200) and a held-out evaluation split (800) before any training.
Because all nuPlan Singapore scenarios come from the one-north district, the two splits share the city's road network; this matches our setting, in which the target city's maps are available while its traffic is held out.
NOMAD never trains on target-city logged scenarios.
Adaptation uses only scenes generated on training-split maps: for each map, we sample 8 scenes by stochastically drawing vehicle placements, densities, speeds, routes, and goals (25{,}600 scenes in total).
Speeds are scaled by the speed range observed in the corresponding training scenario (\cref{sec:scenegen_params}); no logged agent state, trajectory, or behavior is reused.
The city-level density prior (\cref{tab:density_params}) is fit on the training split only, so the evaluation scenarios are never accessed by the generator or the policy, and no behavior-based selection or similarity filter is applied.

\paragraph{Statistics.}
We compute all statistics on the initial scene ($t=0$) of each scenario.
\cref{tab:scenario_stats} describes each scene by its road geometry and by 9 traffic features of the vehicles present at $t=0$:
\emph{\# vehicles} and \emph{density} (vehicles per kilometer of lane);
\emph{moving-vehicle ratio}, the fraction of vehicles with an initial speed above 0.5\,m/s;
the mean, standard deviation, and 90th percentile of the vehicles' \emph{initial speed};
\emph{nearest-vehicle spacing}, the distance from each vehicle to its nearest neighbor, averaged over vehicles, which measures how crowded the scene is;
\emph{goal distance}, the straight-line distance from each moving vehicle to its goal, averaged over vehicles;
and \emph{turning-vehicle ratio}, the fraction of moving vehicles whose goal lies more than $30^\circ$ away from their initial heading, i.e., vehicles that must turn rather than continue straight.
The logged scenarios of the training split are not used for training and are reported only as an i.i.d.\ reference.

\begin{table}[htbp]
\centering
\rev
\caption{Scenario statistics for Boston$\rightarrow$Singapore. All statistics are computed on the initial scene ($t=0$). Each entry is the median over scenes, followed in brackets by the interquartile range (IQR), i.e., the 25th and 75th percentiles, which span the middle 50\% of scenes. The row marked $^\dagger$ reports the mean instead, as its median is 0 in all sets. Generated scenes reuse the training-split maps. The classifier AUC is computed on the 9 traffic features against the held-out evaluation set and reported as mean $\pm$ standard deviation over 5 random draws.}
\label{tab:scenario_stats}
\small
\setlength{\tabcolsep}{4pt}
\begin{tabular}{lccc}
\toprule
 & \textbf{Generated} & \textbf{Logged} & \textbf{Held-out eval} \\
 & \textbf{(training)} & \textbf{(reference)} & \textbf{(evaluation)} \\
\midrule
\# scenes & 25{,}600 & 3{,}200 & 800 \\
\midrule
\multicolumn{4}{l}{\textit{Road geometry}} \\
Lane length (km) & 9.9 [7.0, 13.0] & 9.9 [7.0, 13.0] & 10.1 [7.5, 13.0] \\
\# lane segments & 401 [283, 490] & 401 [283, 490] & 410 [312, 495] \\
\# crosswalks & 19 [13, 26] & 19 [13, 26] & 20 [14, 26] \\
Map extent (ha) & 23.5 [16.8, 29.3] & 23.5 [16.8, 29.3] & 24.1 [16.8, 29.5] \\
\midrule
\multicolumn{4}{l}{\textit{Traffic at $t=0$ (9 features)}} \\
\# vehicles & 4 [2, 6] & 3 [2, 5] & 3 [2, 5] \\
Density (veh./lane-km) & 0.36 [0.24, 0.54] & 0.31 [0.18, 0.57] & 0.31 [0.17, 0.59] \\
Moving-vehicle ratio & 1.00 [0.75, 1.00] & 0.50 [0.33, 1.00] & 0.50 [0.33, 1.00] \\
Initial speed, mean (m/s) & 4.7 [3.3, 6.5] & 3.3 [1.5, 6.2] & 3.3 [1.5, 6.0] \\
Initial speed, std.\ (m/s) & 1.2 [0.4, 2.5] & 2.0 [0.5, 3.2] & 2.2 [0.7, 3.3] \\
Initial speed, 90th pct.\ (m/s) & 6.2 [4.4, 8.3] & 5.9 [3.6, 8.1] & 6.1 [3.7, 8.3] \\
Nearest-vehicle spacing (m) & 124 [100, 179] & 27 [15, 61] & 27 [15, 56] \\
Goal distance (m) & 42 [28, 58] & 46 [31, 60] & 45 [30, 60] \\
Turning-vehicle ratio$^\dagger$ & 0.22 & 0.18 & 0.16 \\
\midrule
Classifier AUC vs.\ eval & 0.993 $\pm$ 0.001 & 0.501 $\pm$ 0.019 & -- \\
\bottomrule
\end{tabular}
\end{table}

\paragraph{Findings.}
Generated scenes share the training split's road geometry, and their vehicle counts, densities, and goal distances are in a similar range to logged scenes.
They differ in motion and spacing: almost all generated vehicles are moving at $t=0$ (median moving-vehicle ratio 1.00 vs.\ 0.50 in the evaluation set), and they are more widely spaced (median nearest-vehicle spacing 124\,m vs.\ 27\,m).
The classifier separates generated training scenes from evaluation scenes almost perfectly (AUC 0.993).
By contrast, the logged training split is indistinguishable from the evaluation split (AUC 0.501), which is consistent with an unbiased random partition.
The generated training scenes are therefore not selected to resemble the evaluation traffic.
}

%% file: appendices/TrafficDensitySensitivity.tex
\section{Sensitivity of NOMAD to traffic density meta-information} \label{sec:sensitivitytoerrorsintrafficdensity}

NOMAD uses target-city meta-information to construct traffic scenarios that approximate the traffic distribution of the deployment city. However, such meta-information may be imperfect in practice, particularly for traffic density, which varies over time and may only be available as a coarse estimate. This section therefore evaluates the sensitivity of NOMAD to errors in the estimated traffic density used by the scenario generator.

\begin{figure*}[htpb]
  \centering
  \includegraphics[width=1.0\linewidth]{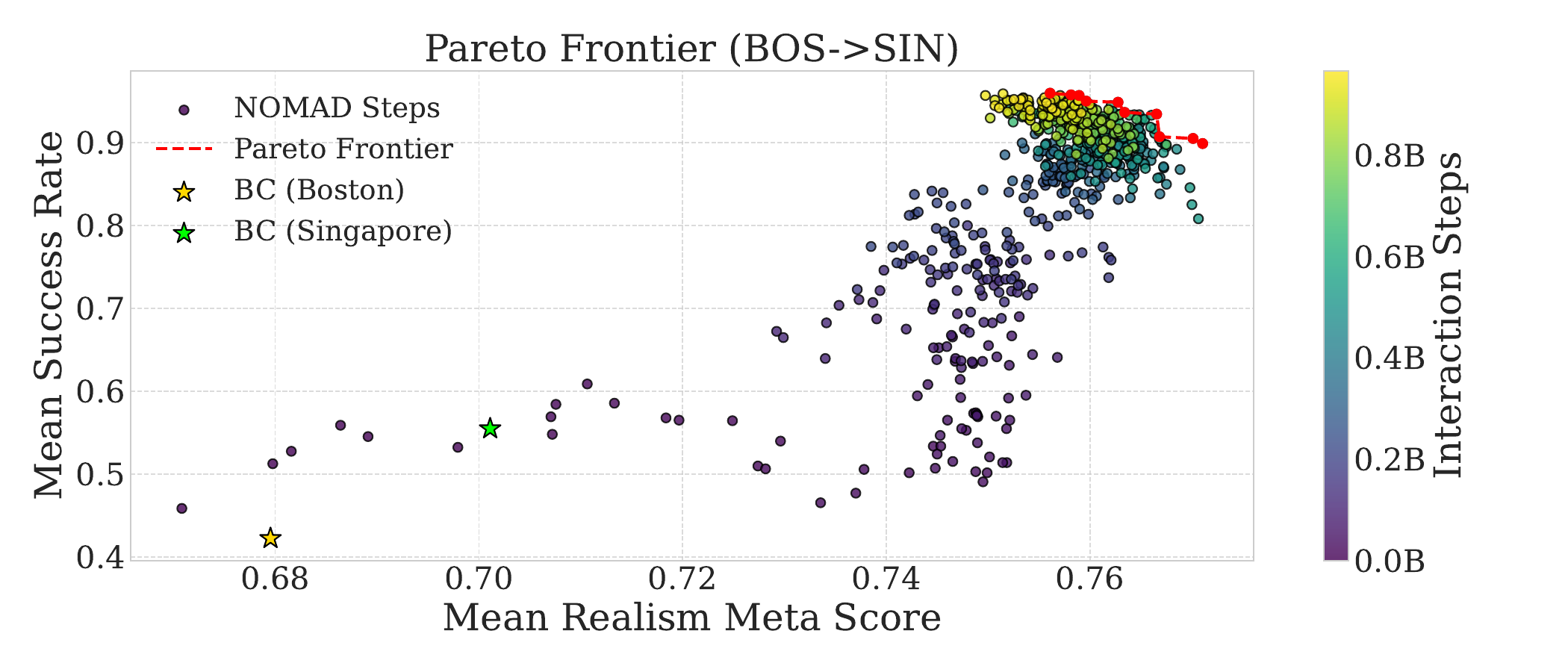}
  \caption{\textbf{Pareto frontier of success rate versus realism meta score across training checkpoints for Boston-to-Singapore transfer with small traffic density perturbation.}
  Each point corresponds to a policy checkpoint, colored by the number of interaction steps.
  The yellow star denotes the zero-shot transfer policy $\pi^0$ from Boston, while the lime star shows behavior cloning using Singapore trajectories.
  The red dashed line and points indicate the empirical Pareto frontier on this test set.
  }
  \label{fig:bos_to_sin_small_density_perturb}
\end{figure*}

\begin{figure*}[htpb]
  \centering
  \includegraphics[width=1.0\linewidth]{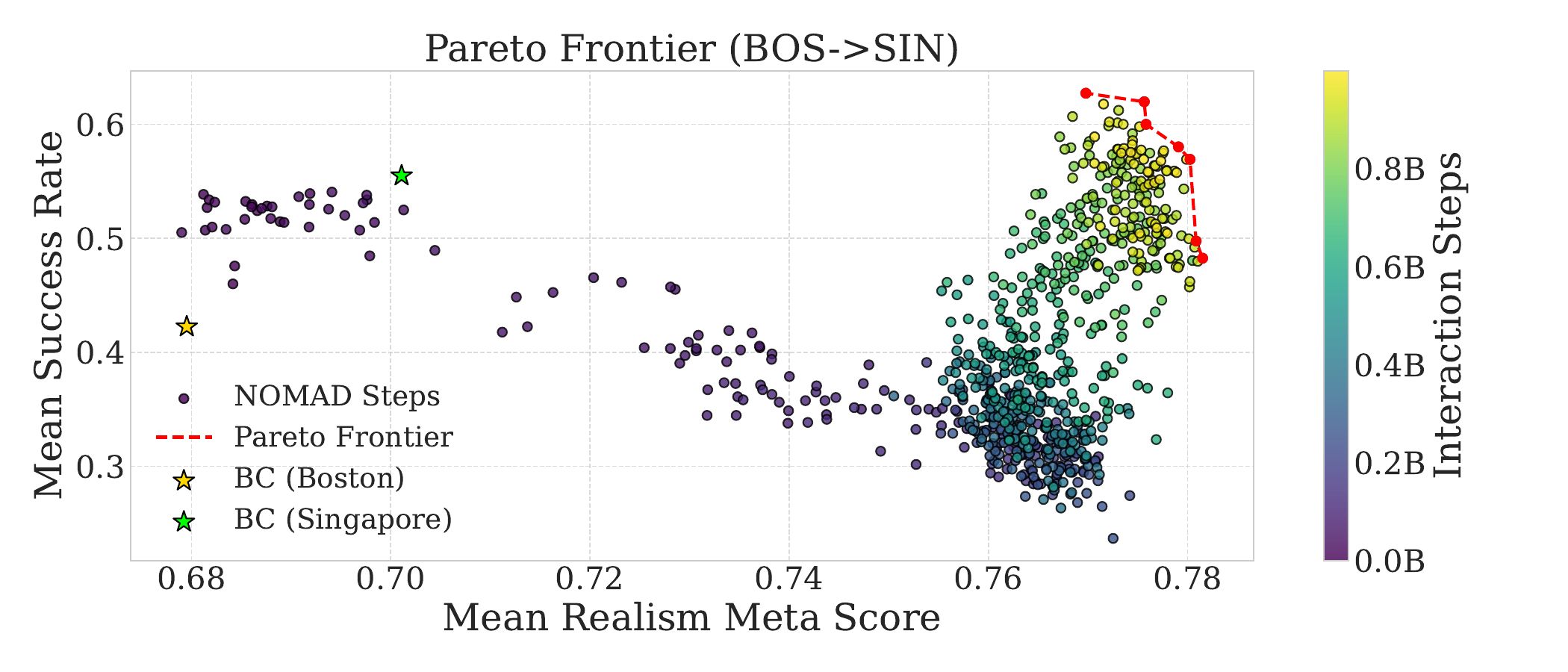}
  \caption{\textbf{Pareto frontier of success rate versus realism meta score across training checkpoints for Boston-to-Singapore transfer with large traffic density perturbation.}
  Each point corresponds to a policy checkpoint, colored by the number of interaction steps.
  The yellow star denotes the zero-shot transfer policy $\pi^0$ from Boston, while the lime star shows behavior cloning using Singapore trajectories.
  The red dashed line and points indicate the empirical Pareto frontier on this test set.
  }
  \label{fig:bos_to_sin_large_density_perturb}
\end{figure*}

We evaluate the robustness of NOMAD to inaccuracies in the target city's traffic-density meta-information using the Boston-to-Singapore transfer setting. Specifically, we perturb the density prior used by the scenario generator and compare the resulting policies with those trained using the nominal estimate. \cref{fig:bos_to_sin_small_density_perturb} shows that, under moderate perturbations, the learned policies achieve success–realism trade-offs that are nearly indistinguishable from the nominal setting. In contrast, large density perturbations, shown in \cref{fig:bos_to_sin_large_density_perturb}, lead to a clear performance drop, primarily in success rate.

These results align with the design of NOMAD. The traffic-density prior only shapes the distribution of the \textit{initial conditions and goals}; it does not provide supervision, demonstration labels, or reward information. As a result, moderate errors do not substantially alter the interaction distribution encountered during self-play. However, severe density misspecification can generate traffic scenarios that are systematically too sparse or too dense relative to the target city, reducing the relevance of self-play adaptation. Overall, NOMAD does not require an exact traffic-density estimate, but its performance benefits from a prior that remains within a plausible range for the target city.

%% file: appendices/VisualizationofScenarios.tex
\section{Examples of Scenarios}

\subsection{Logged Scenarios}
\begin{figure*}[h!]
  \centering
  \begin{subfigure}[b]{0.45\textwidth}
    \centering
    \includegraphics[width=\linewidth]{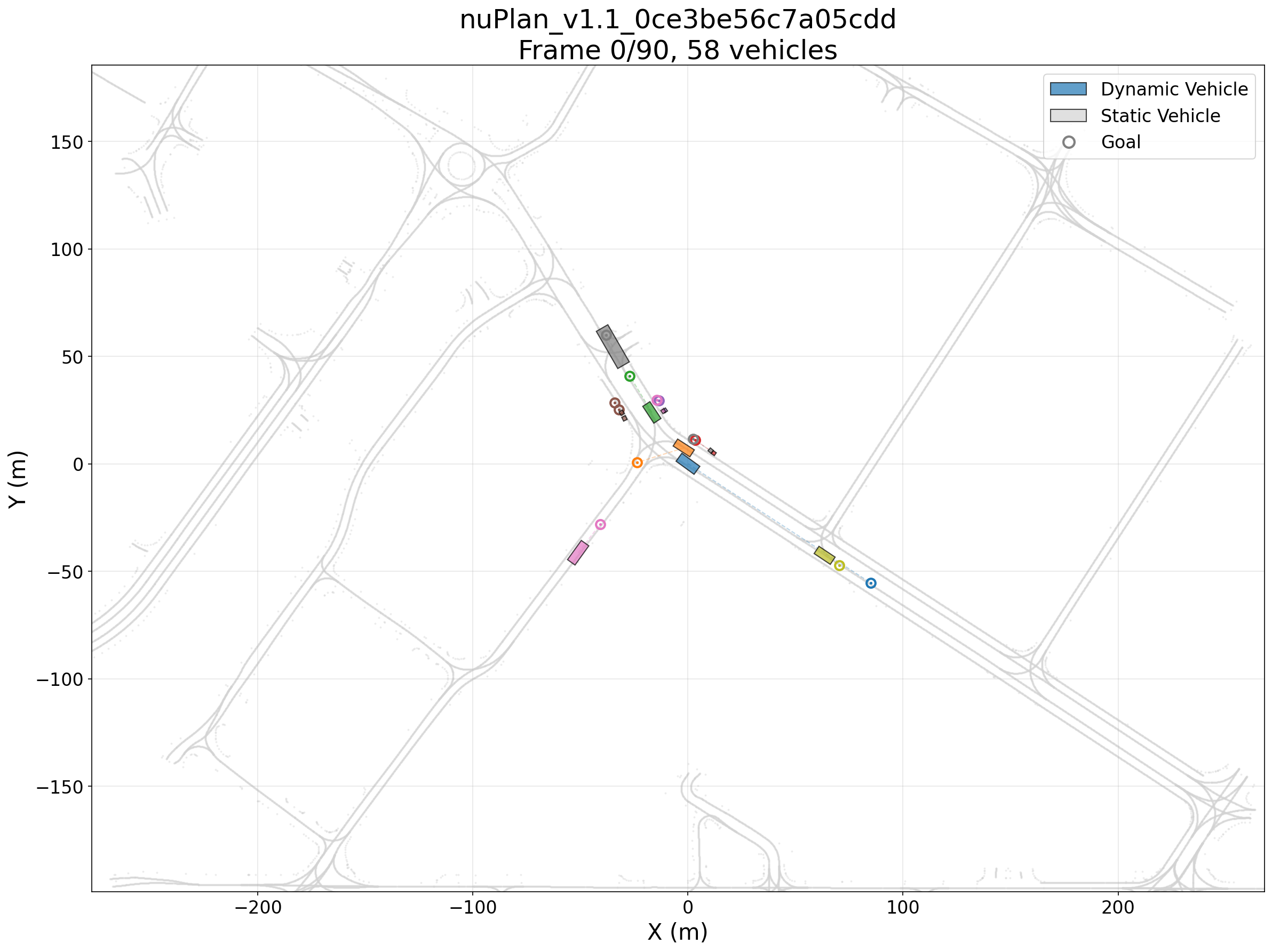}
    \caption{Boston}
    \label{fig:1a}
  \end{subfigure}
  \hfill
  \begin{subfigure}[b]{0.45\textwidth}
    \centering
    \includegraphics[width=\linewidth]{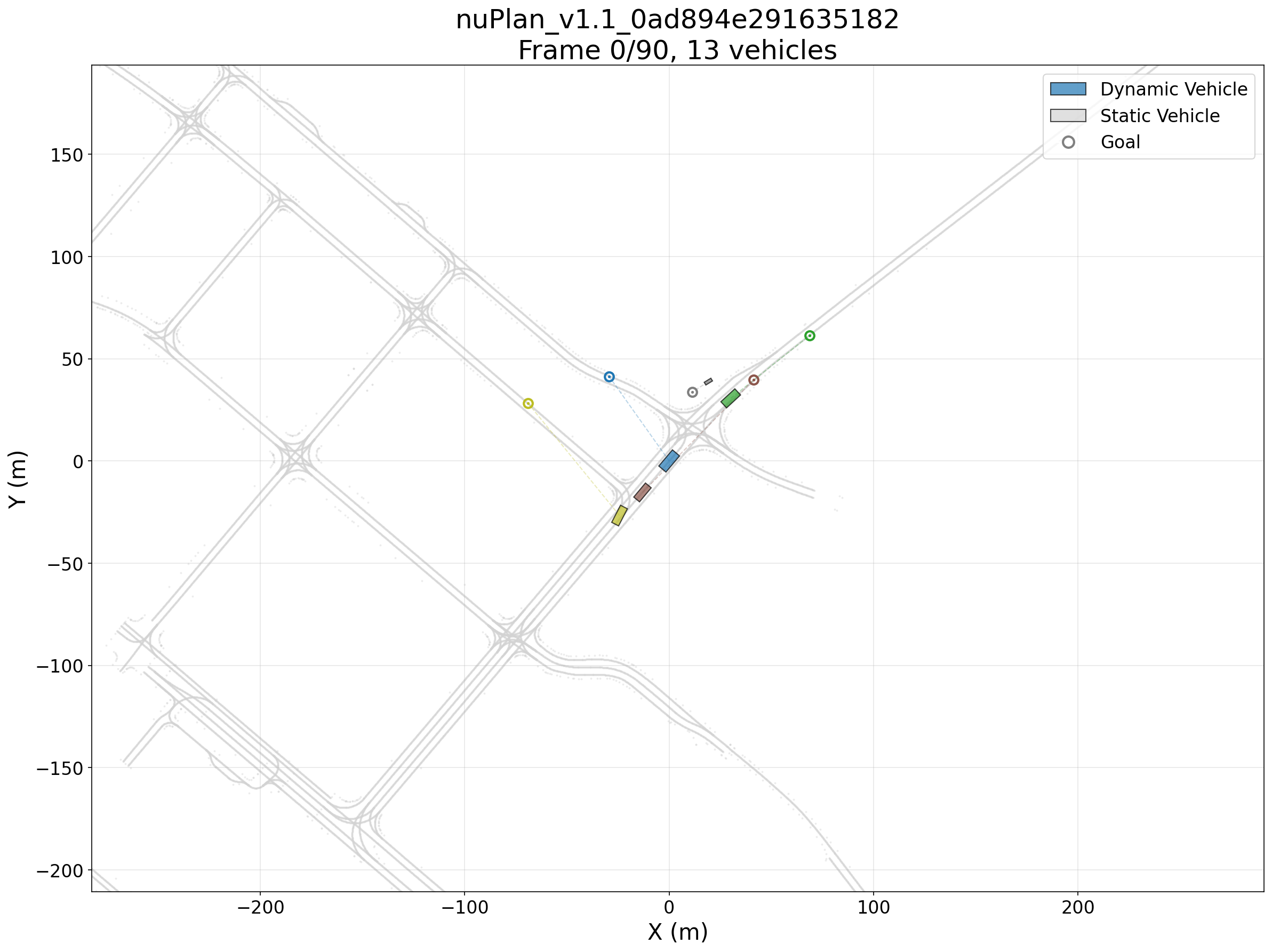}
    \caption{Pittsburgh}
    \label{fig:1b}
  \end{subfigure}

  \vspace{1ex}

  \begin{subfigure}[b]{0.45\textwidth}
    \centering
    \includegraphics[width=\linewidth]{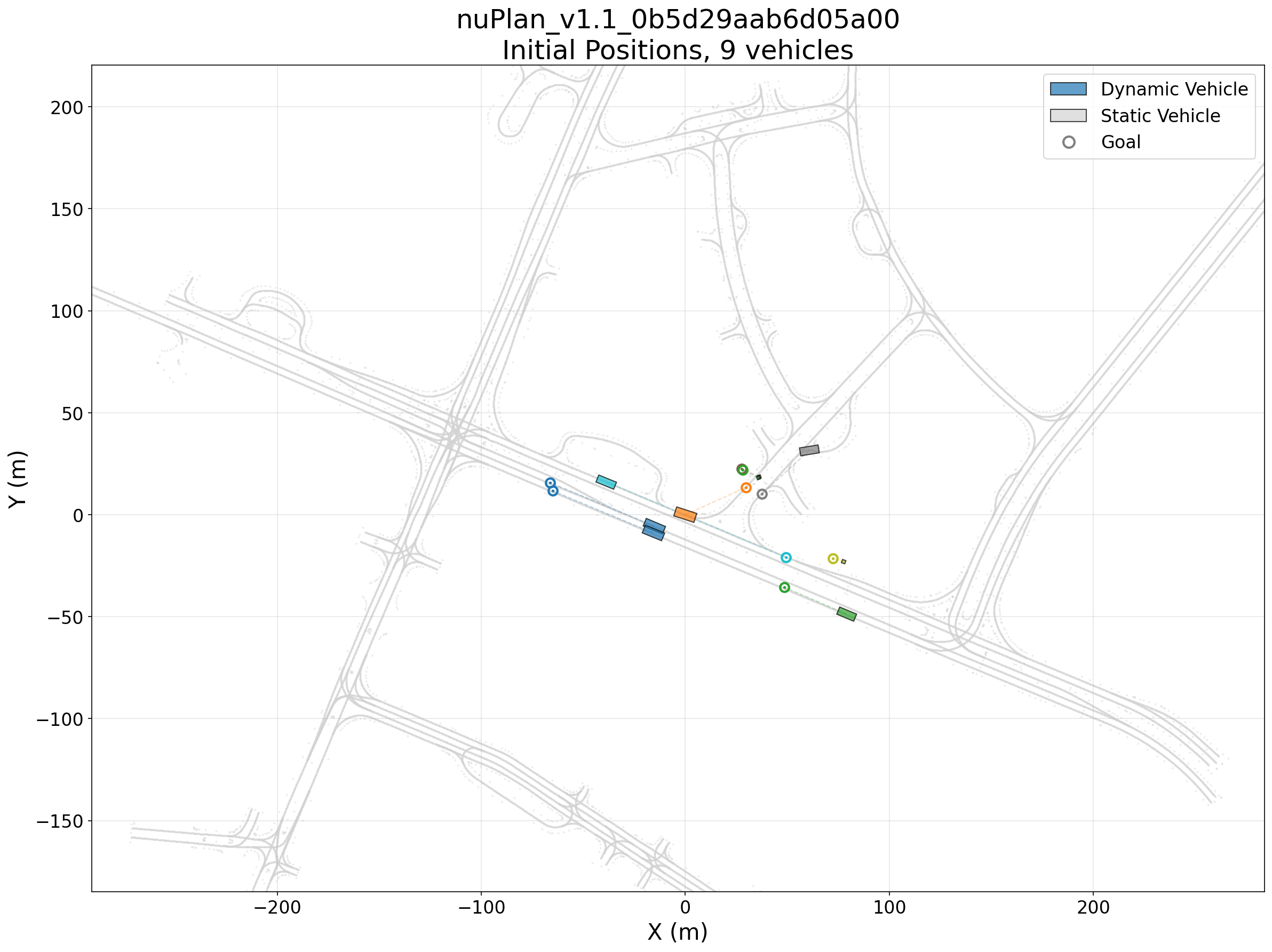}
    \caption{Singapore (original)}
    \label{fig:2a}
  \end{subfigure}
  \hfill
  \begin{subfigure}[b]{0.45\textwidth}
    \centering
    \includegraphics[width=\linewidth]{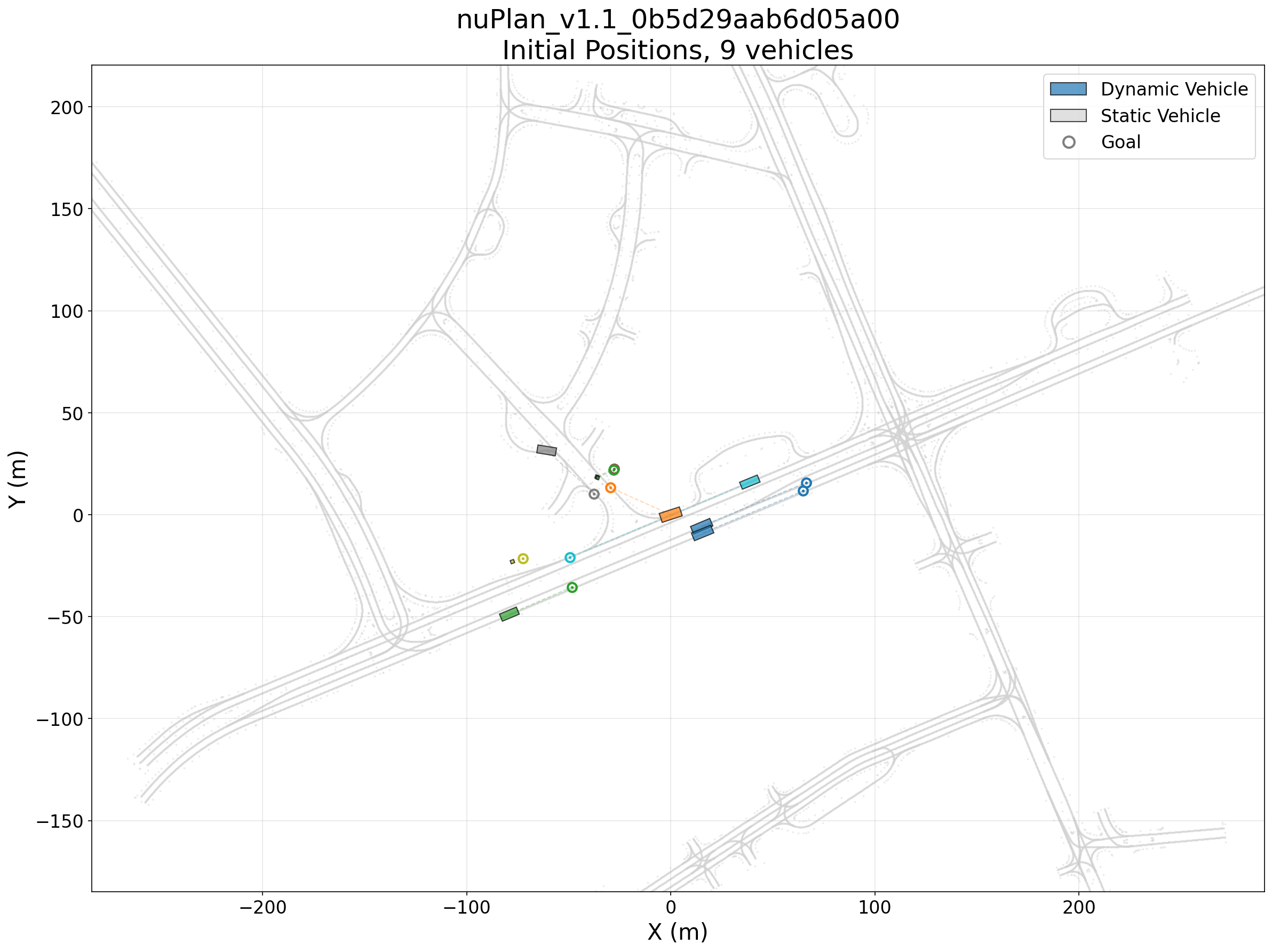}
    \caption{Singapore (mirrored)}
    \label{fig:2b}
  \end{subfigure}

  \caption{\textbf{Visualization of scenarios from different cities in the nuPlan dataset.}
  The rectangles denote vehicles and the circles with the corresponding color denote their goals.
  To mirror the Singapore and make it a right-hand traffic city, we first flip the scenario across x-axis, then we change the orientation of road edges while keeping the orientation of road lanes.
  }
  \label{fig:cities}
\end{figure*}

We visualize logged scenarios from different cities in \cref{fig:cities}.

\subsection{Generated Scenarios}

\begin{figure*}[h!]
  \centering
  \begin{subfigure}[b]{0.45\textwidth}
    \centering
    \includegraphics[width=\linewidth]{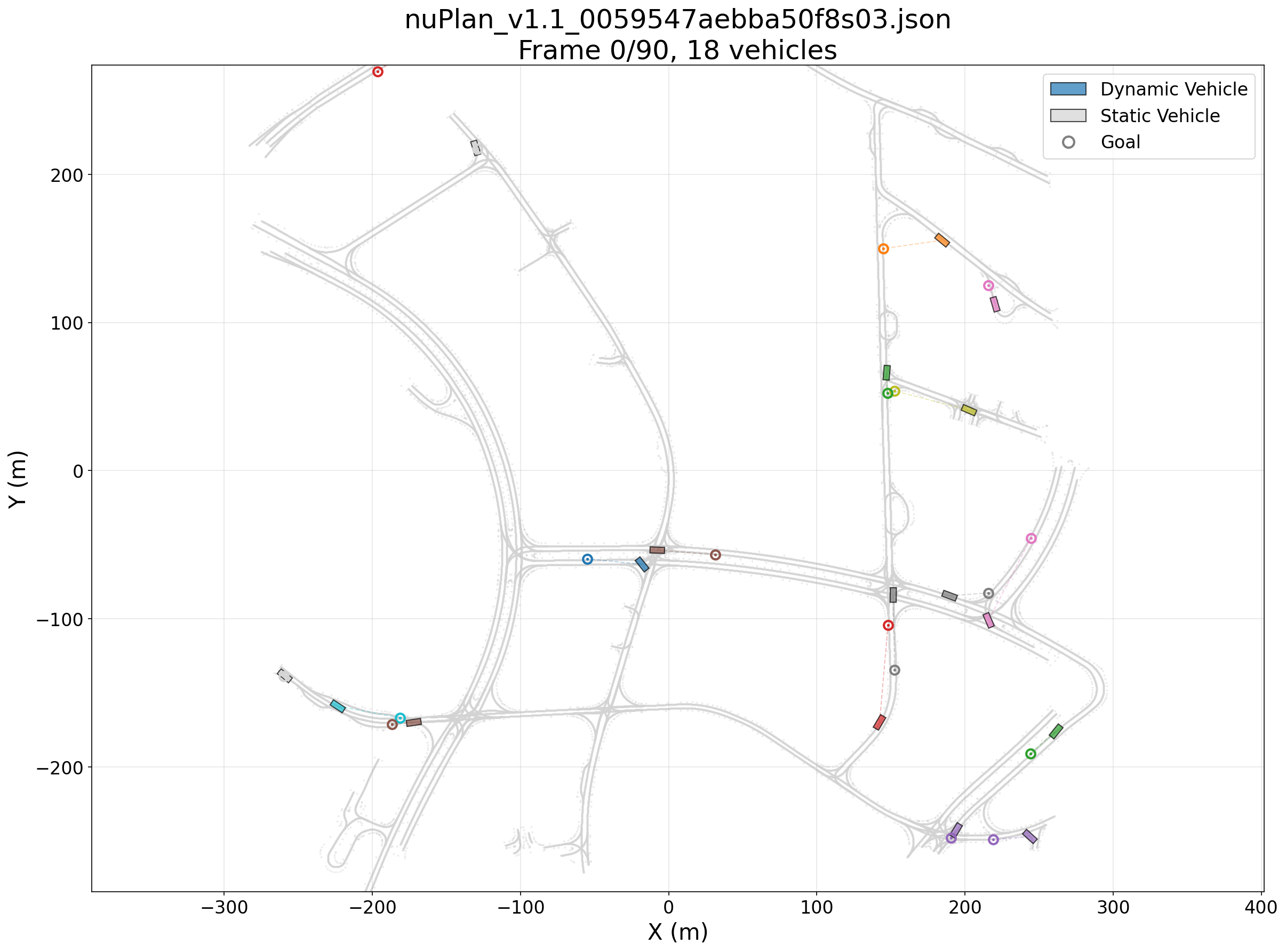}
    \caption{Map 1 - Generated scenario 1}
    \label{fig:1a}
  \end{subfigure}
  \hfill
  \begin{subfigure}[b]{0.45\textwidth}
    \centering
    \includegraphics[width=\linewidth]{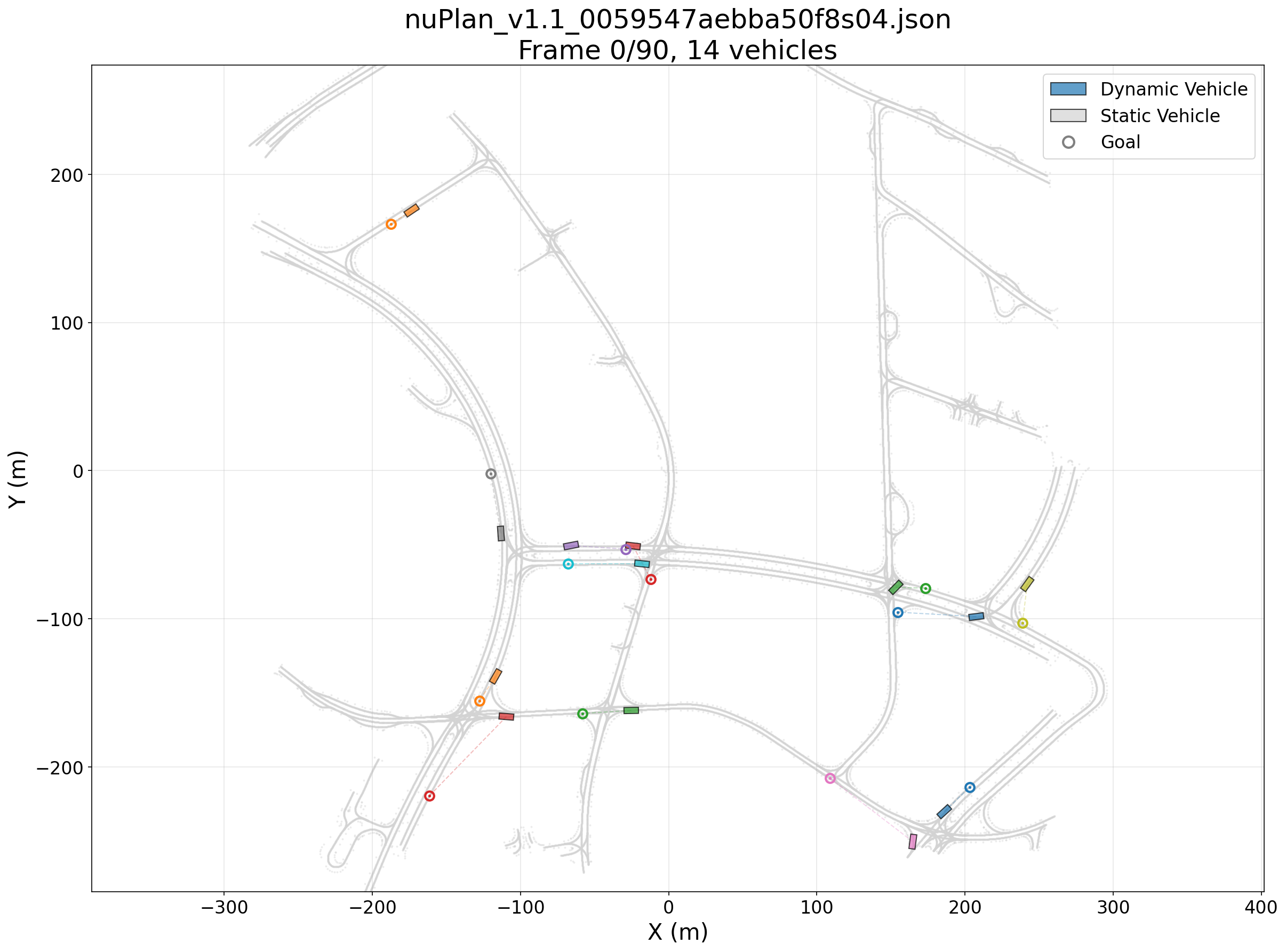}
    \caption{Map 1 - Generated scenario 2}
    \label{fig:1b}
  \end{subfigure}

  \vspace{1ex}

  \begin{subfigure}[b]{0.45\textwidth}
    \centering
    \includegraphics[width=\linewidth]{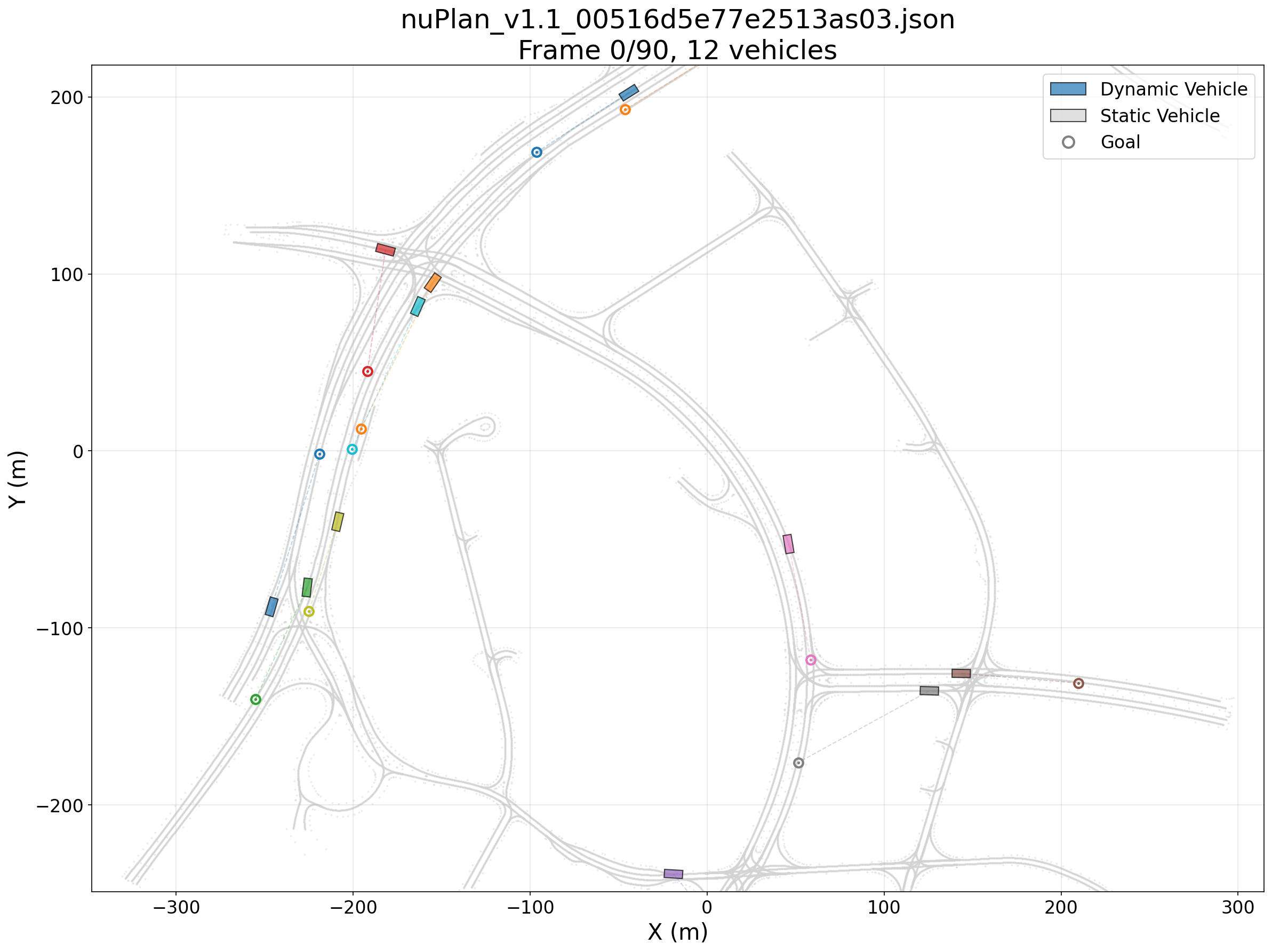}
    \caption{Map 2 - Generated scenarios 1}
    \label{fig:2a}
  \end{subfigure}
  \hfill
  \begin{subfigure}[b]{0.45\textwidth}
    \centering
    \includegraphics[width=\linewidth]{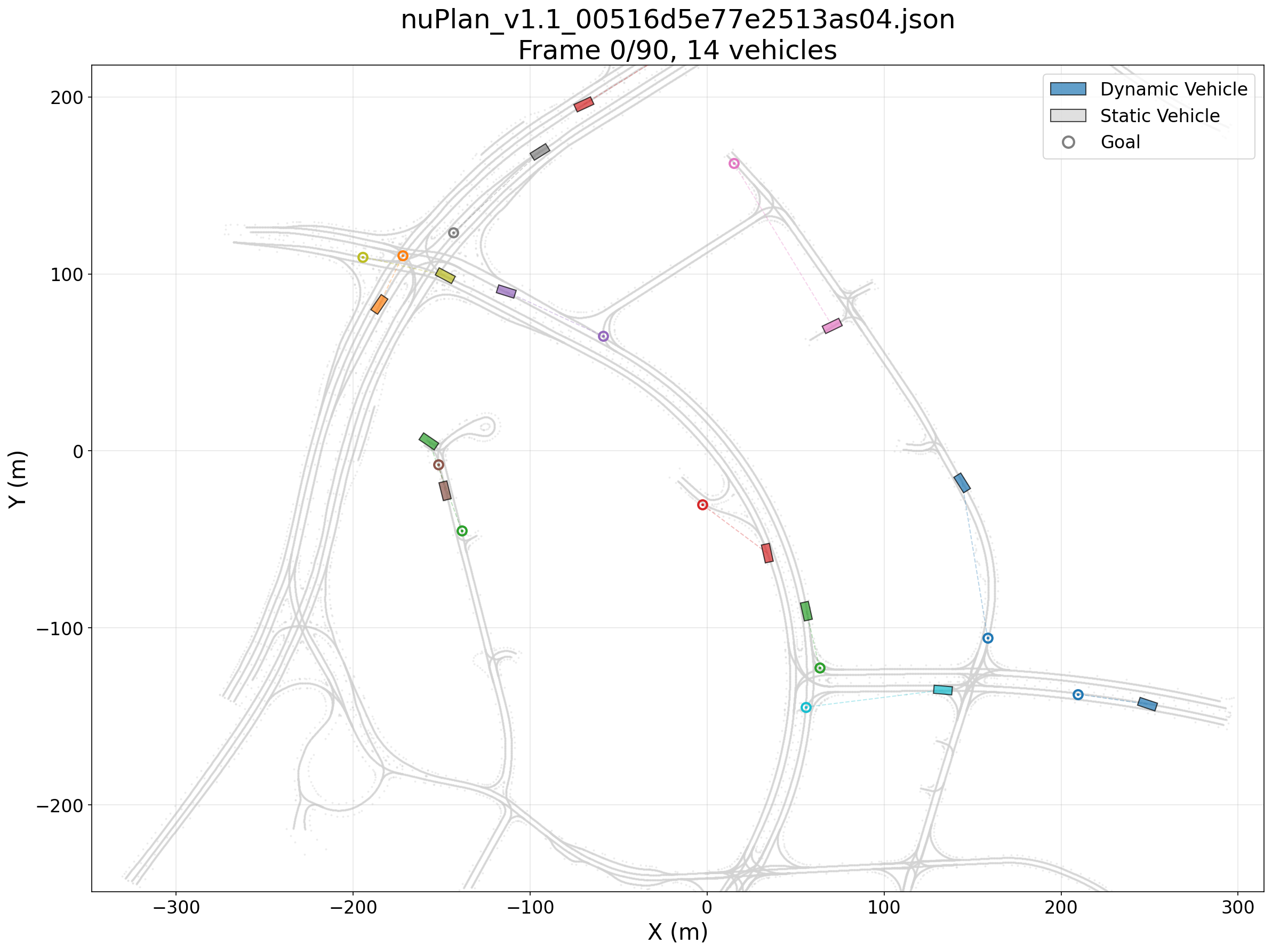}
    \caption{Map 2 - Generated scenario 2}
    \label{fig:2b}
  \end{subfigure}

  \caption{Examples of generated scenarios on Singapore map for two different sampled map segments.
  }
  \label{fig:scene_gen_ex}
\end{figure*}

\cref{fig:scene_gen_ex} demonstrates examples of generated scenarios on two sampled map segments from Singapore. Vehicles are depicted by colored rectangles, and their assigned goal points are illustrated using a circle with the same color. As it can be seen, vehicles are generated on drivable areas on all scenarios, and their corresponding goals are also generated on drivable areas at a reachable distance. Moreover, there are no overlapping vehicles on the map or vehicles with unrealistic headways. The illustration of two scenarios from the same map segment aims to demonstrate the effect of random generation of scenarios.

%% file: appendices/DetailedMetrics.tex
\section{Metric Definitions}\label{sec:extendedmetrics}

\subsection{WOSAC Realism Evaluation Metric}

The Waymo Open Sim Agents Challenge (WOSAC)~\citep{montali2023waymo} is a benchmark for evaluating the realism of simulation agents in autonomous driving. The core idea behind WOSAC evaluation is that realistic simulation agents should produce behavior distributions that match the actual distribution of real-world driving scenarios observed in logged data. To quantify this, WOSAC computes the approximate negative log-likelihood (NLL) of logged human trajectories under the distribution induced by the simulation agent. Lower NLL (equivalently, higher likelihood) indicates that the simulated behaviors better match the distribution of real human driving. 
For evaluation, WOSAC requires agents to generate $K$ rollouts per scenario. 

\subsubsection{Component Metrics.} 

WOSAC decomposes realism into nine interpretable component metrics organized into three categories rather than computing a single holistic likelihood over full trajectories. These three elements largely overlap with comfort/kinematics, TTC/collisions, and drivable-area compliance categories in nuPlan planning challenge. 
We favor WOSAC over nuPlan's evaluation metrics because nuPlan’s metrics are predominantly threshold- and constraint-based validation and safety checks, whereas WOSAC provides a distributional measure of \textit{humanlikeness} that aligns with our notion of realism.

The three components of the WOSAC metrics are detailed as follows:

\textbf{Kinematic Metrics} capture the motion dynamics of individual agents, measuring whether simulated vehicles move with human-like speed and acceleration profiles. Let $p_t = [p_t^x, p_t^y, p_t^z]^\top$ denote the position at timestep $t$, and $\theta_t$ denote the heading angle:
\begin{enumerate}
    \item Linear Speed: $\|v_t\| = \|(p_{t+1} - p_t)/\Delta t\|_2$. This measures the instantaneous speed of the vehicle. 
    
    \item Linear Acceleration: $(\|v_{t+1}\| - \|v_t\|)/\Delta t$. This captures how aggressively the vehicle speeds up or slows down. Jerky or unnaturally aggressive acceleration patterns indicate unrealistic behavior.
    
    \item Angular Speed: $\omega_t = d(\theta_{t+1}, \theta_t)/\Delta t$, where $d(\cdot, \cdot)$ computes the minimal angular difference on the unit circle. This measures how quickly the vehicle is turning.
    
    \item Angular Acceleration: $d(\omega_{t+1}, \omega_t)/\Delta t$. This captures the smoothness of steering behavior.
\end{enumerate}

\textbf{Interaction Metrics} capture multi-agent dynamics, measuring whether simulated vehicles interact with other road users in human-like ways:
\begin{enumerate}
    \setcounter{enumi}{4}
    \item Distance to Nearest Object: The signed distance to the nearest other object, computed via the GJK (Gilbert-Johnson-Keerthi) algorithm on bounding box polygons. Positive values indicate separation, while negative values indicate overlap (collision). unrealistic agents may follow too closely or maintain unnaturally large gaps.
    
    \item Collisions: A binary indicator $\mathbf{1}\{\text{signed distance} < 0\}$ for whether a collision has occurred. While the distance metric captures the full distribution of proximity to other objects, collisions are weighted more heavily in the final score because they represent safety-critical failures. 
    
    \item Time-to-Collision (TTC): The estimated time before a collision would occur, assuming both vehicles maintain their current velocities. TTC captures the temporal safety margin that human drivers maintain. 
\end{enumerate}

\textbf{Map-based Metrics} measure whether simulated vehicles respect road boundaries and lane structures:
\begin{enumerate}
    \setcounter{enumi}{7}
    \item Distance to Road Edge: The signed distance to the nearest road boundary. Positive values indicate the vehicle is within the drivable area, while negative values indicate the vehicle has departed the road.
    
    \item Road Departures (Offroad): A binary indicator for whether the vehicle has left the drivable area. Like collisions, road departures are weighted more heavily.
\end{enumerate}

\subsubsection{Time-Series NLL Computation} 

For each component metric $j$, WOSAC computes a score by comparing the distribution of feature values from the logged human trajectory against the empirical distribution from $K$ simulated rollouts. The score is computed as an average over the time axis with validity masking to handle agents that may enter or exit the scene. Let $\mathcal{V}_t \in \{0, 1\}$ denote the validity indicator at timestep $t$ (indicating whether the agent is active):
\begin{equation}
    m_j = \exp\left(-\frac{1}{\sum_{t=0}^{H-1} \mathbbm{1}(\mathcal{V}_t)} \sum_{t=0}^{H-1} \mathbbm{1}(\mathcal{V}_t) \cdot \text{NLL}_{j,t}\right),
\end{equation}
where $H$ is the horizon length. The per-timestep $\text{NLL}_{j,t}$ is estimated by constructing a histogram-based categorical distribution over the $K$ sampled feature values, with Laplace smoothing to avoid infinite NLL for out-of-distribution logged values. The exponential transformation maps the average NLL to the $[0, 1]$ range, where higher values indicate greater realism (i.e., the logged human behavior is more likely under the simulated distribution).

\subsubsection{Final Realism Meta Score -- Composite Metric} 

Let $|\mathcal{D}|$ denote the number of test scenarios and $M=9$ the number of component metrics. The final realism meta score aggregates all component metrics via a weighted average:
\begin{equation}
    \mathcal{M}^K = \frac{1}{|\mathcal{D}| \cdot M} \sum_{i=1}^{|\mathcal{D}|} \sum_{j=1}^{M} w_j \cdot m^K_{i,j}, \quad \text{where} \quad \sum_{j=1}^{M} w_j = 1,
\end{equation}
where $w_j$ denotes the weight for metric $j$. Following WOSAC, we set the weights for collision and road departure metrics to be twice as large as other components. This weighting emphasizes safety-critical behaviors, reflecting the fact that collisions and road departures are unambiguous indicators of unrealistic driving that human drivers almost never exhibit. 

The WOSAC composite metric $\mathcal{M}^K$ provides a tractable approximation to our realism meta score $R(\pi, C)$ defined in Equation~\ref{eq:realism}: rather than estimating the full trajectory likelihood $\log q_\pi(\tau|\xi)$ directly, WOSAC decomposes it into interpretable feature-level NLLs computed by comparing logged human trajectories $\tau \sim p(\cdot|\xi)$ against $K$ rollouts from the policy $q_\pi(\cdot|\xi)$. Throughout the paper, we use ``realism meta score'' and ``WOSAC realism meta score'' interchangeably.

\subsubsection{Realism Score is City-Dependent}\label{sec:wosaccitydependent}

In addition, we investigate whether the realism meta score between different cities is comparable by evaluating the realism of human trajectories in different cities.
The results are reported in \cref{tab:expertpolicy}, where $\pi^{\text{expert}}_c$ denotes continuous inferred actions from logged demonstrations, which provide an empirical upper bound on realism, and $\pi^{\text{expert}}_d$ denotes discretized inferred actions used to train behavior cloning (BC) policies.
The different realism meta scores achieved by $\pi^{\text{expert}}_c$ across cities indicate that realism is inherently city-dependent and should therefore be interpreted comparatively within the same city rather than in absolute terms across different cities.

\begin{table}[htbp]
\caption{\textbf{Performance of expert policies in different cities.}
}
\label{tab:expertpolicy}
\centering
\begin{tabular}{@{}cccc@{}}
\toprule
\textbf{Policy} & \textbf{Scenarios}   & \textbf{Realism Meta Score} & \textbf{Success Rate} \\ \hline
\multirow{2}{*}{$\pi^\text{expert}_{c}$} & Boston & 0.8311 & 100.0\% \\
& Singapore           &   0.8581                     &  100.0\%                     \\
\hline
\multirow{2}{*}{$\pi^\text{expert}_{d}$} & Boston & 0.7814 & 85.47\% \\
& Singapore           &   0.8056                     &  88.26\%                     \\
\bottomrule
\end{tabular}
\end{table}

\paragraph{Local driving conventions.}
WOSAC compares simulated behavior with human reference trajectories from the target city, which allows the selected feature distributions to partially reflect local driving patterns. For instance, speed and acceleration features characterize aspects of vehicle motion, while proximity and time-to-collision features provide information about interactions with surrounding vehicles.
These measurements capture aspects of local conventions that are expressed through trajectories.
Informal signaling, horn use, and courtesy gestures are not directly represented, although their effects on vehicle motion may influence the measured features.
We therefore complement the realism score with component-level results, collision and off-road rates, and the failure cases in \cref{sec:failurecasestudy}. Therefore, trajectory-based evaluation may not fully reflect the social context and expectations underlying local driving conventions.

\subsection{Additional Metrics}
We report collision rate and offroad rate separately, while they are already captured within the WOSAC interaction and map-based component metrics, respectively.
Beyond the WOSAC realism meta score, we report the Average Displacement Error (ADE) as a supplementary metric.
We compute ADE as the mean L2 distance between the position of the simulated agent and the position of the logged human trajectory at corresponding timesteps:
\begin{equation}
    \text{ADE} = \frac{1}{|\mathcal{D}|} \sum_{i=1}^{|\mathcal{D}|} \frac{1}{H} \sum_{t=1}^{H} \|p_t^{\pi,(i)} - p_t^{\text{log},(i)}\|_2,
\end{equation}
where $p_t^{\pi,(i)}$ denotes the position of the agent controlled by policy $\pi$ at timestep $t$ in scenario $i$, and $p_t^{\text{log},(i)}$ denotes the corresponding position from the logged human trajectory.

Unlike open-loop trajectory prediction tasks, where ADE measures prediction accuracy against a fixed ground truth, in closed-loop evaluation, ADE should be interpreted as a measure of behavioral similarity rather than prediction accuracy. This means that lower ADE indicates that the policy produces trajectories that remain closer to what a human driver executed in the same scenario, even though the two trajectories evolve under different closed-loop dynamics. We include ADE as a complementary reference metric, while recognizing that the WOSAC realism meta score provides a more principled distributional measure of human-likeness.

%% file: appendices/ExtendedMetricAnalysis.tex
\section{Sub-Metric Analysis}
\label{sec:extendedmetricanalysis}

\begin{figure*}[htbp]
  \centering
  \begin{subfigure}[b]{0.48\textwidth}
    \centering
    \includegraphics[width=\linewidth]{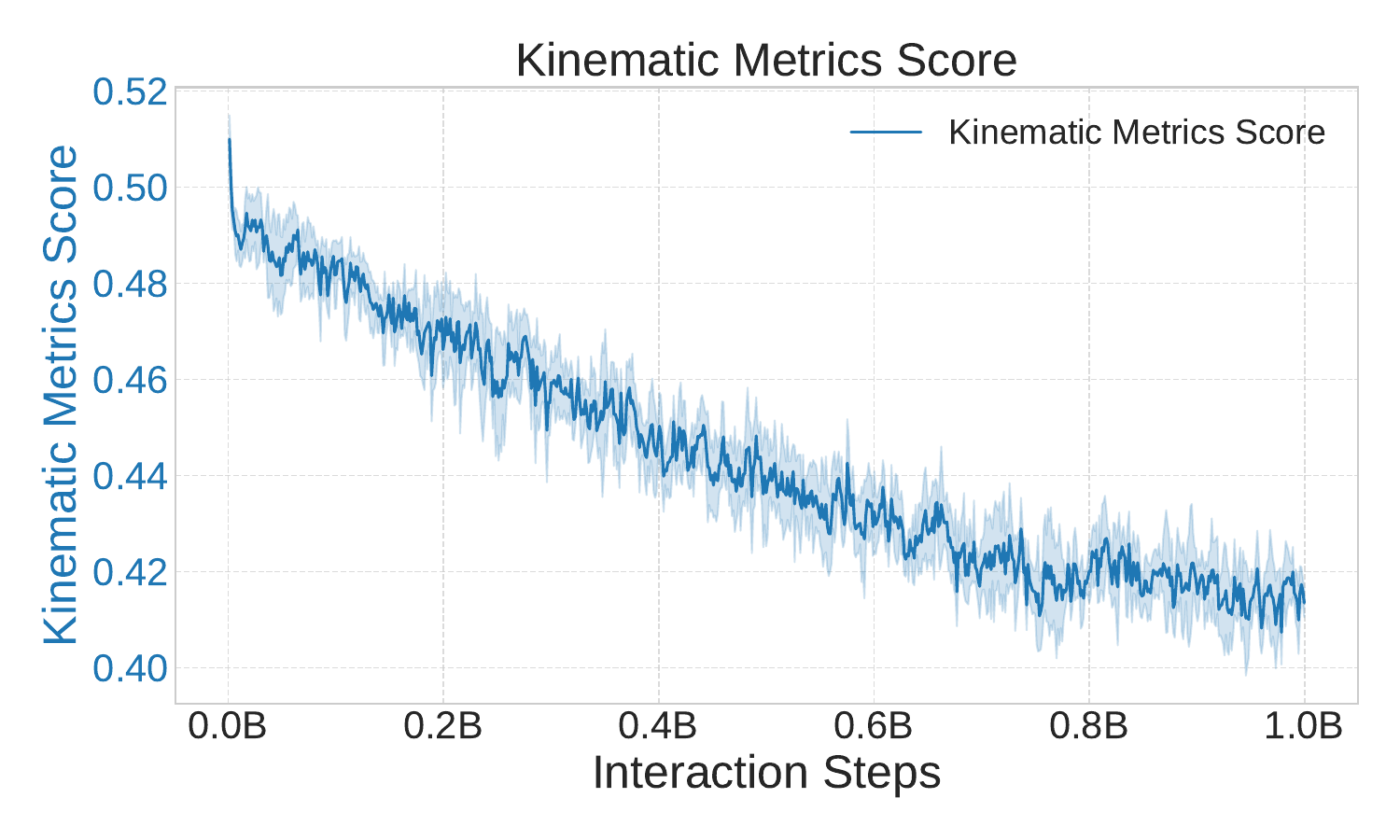}
    \caption{Kinematics metric}
    \label{fig:1a}
  \end{subfigure}
  \hfill
  \begin{subfigure}[b]{0.48\textwidth}
    \centering
    \includegraphics[width=\linewidth]{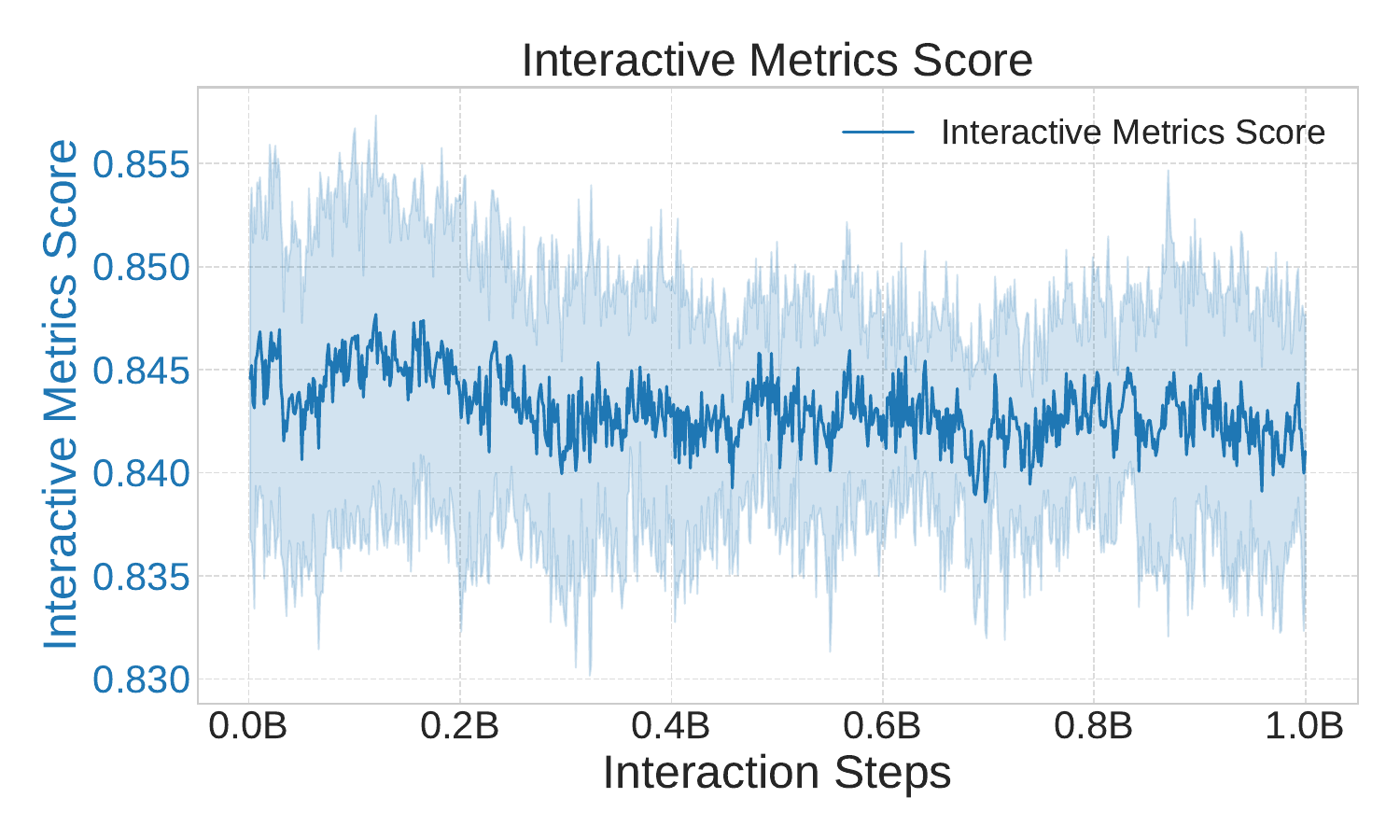}
    \caption{Interactive metric}
    \label{fig:1b}
  \end{subfigure}

  \vspace{1ex}

  \begin{subfigure}[b]{0.48\textwidth}
    \centering
    \includegraphics[width=\linewidth]{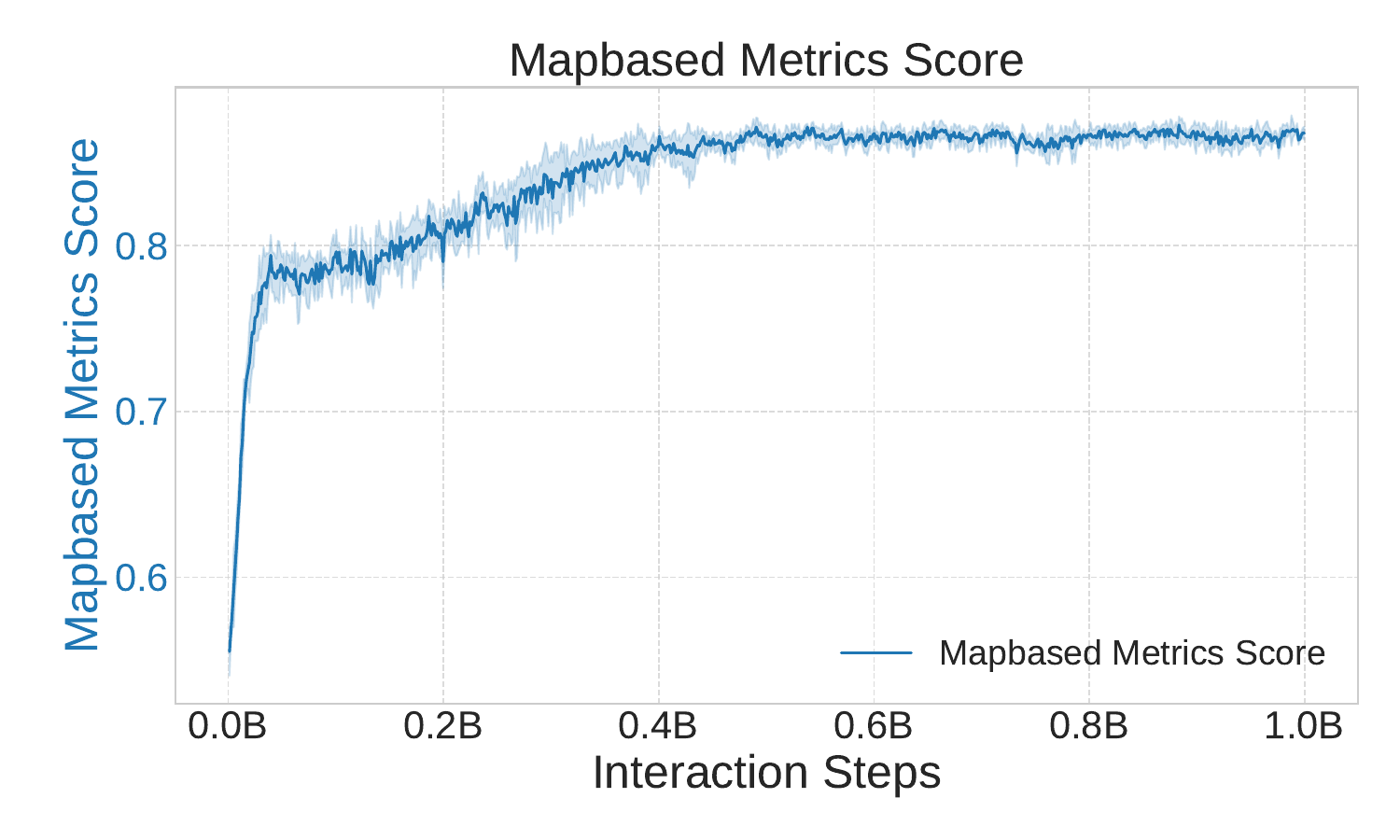}
    \caption{Map-based metric}
    \label{fig:2a}
  \end{subfigure}
  \hfill
  \begin{subfigure}[b]{0.48\textwidth}
    \centering
    \includegraphics[width=\linewidth]{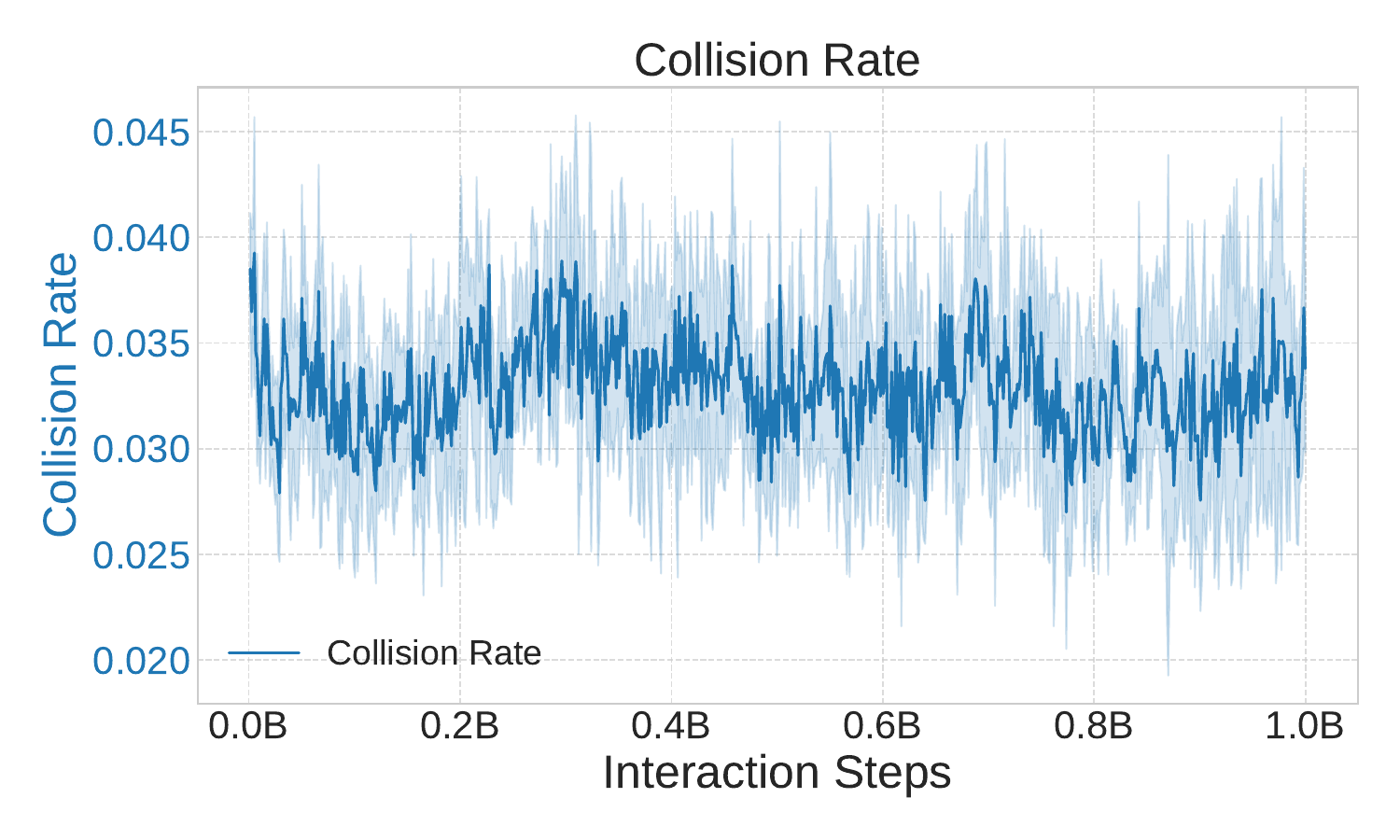}
    \caption{Collision Rate}
    \label{fig:2b}
  \end{subfigure}

  \vspace{1ex}

  \begin{subfigure}[b]{0.48\textwidth}
    \centering
    \includegraphics[width=\linewidth]{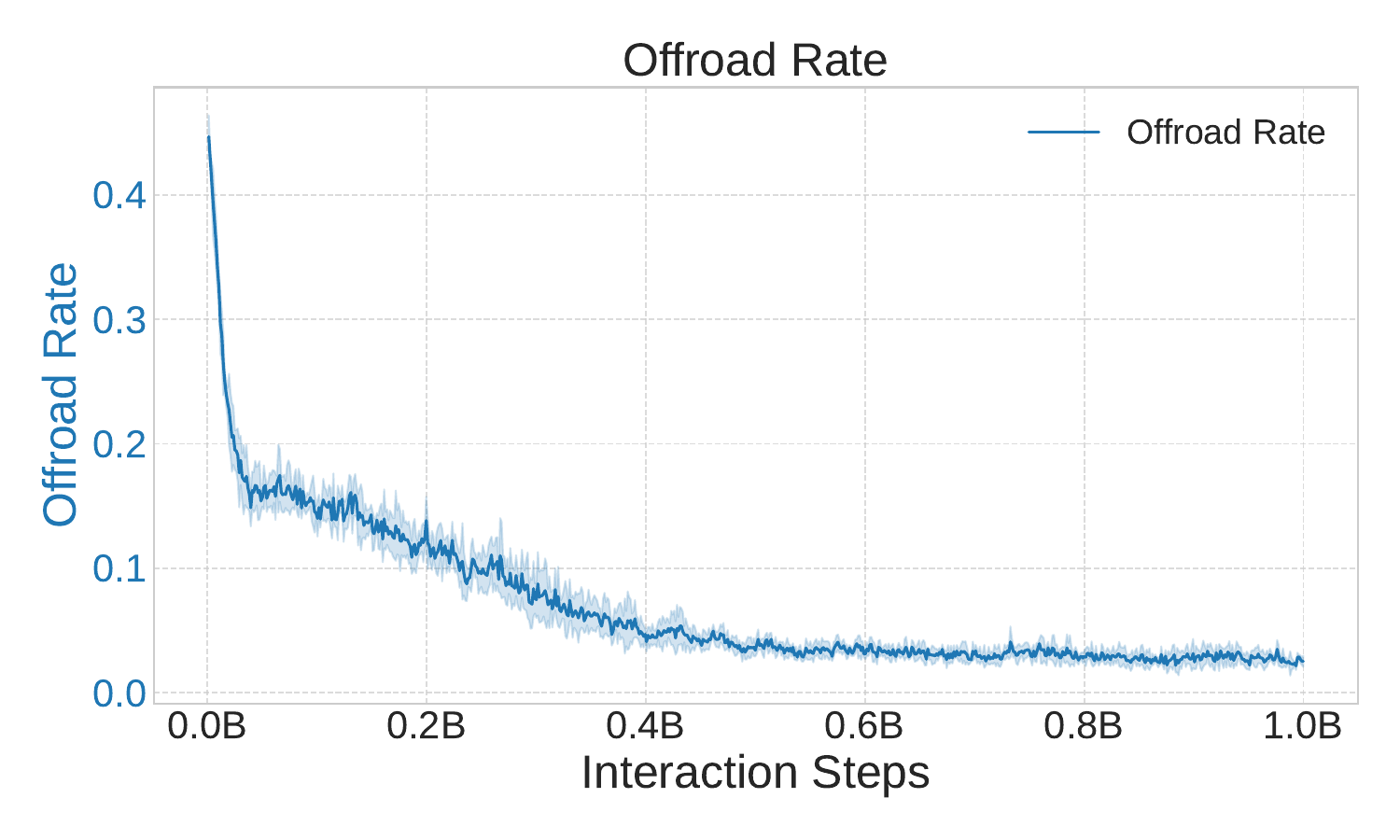}
    \caption{Offroad Rate}
    \label{fig:3b}
  \end{subfigure}
    \hfill
  \begin{subfigure}[b]{0.48\textwidth}
    \centering
    \includegraphics[width=\linewidth]{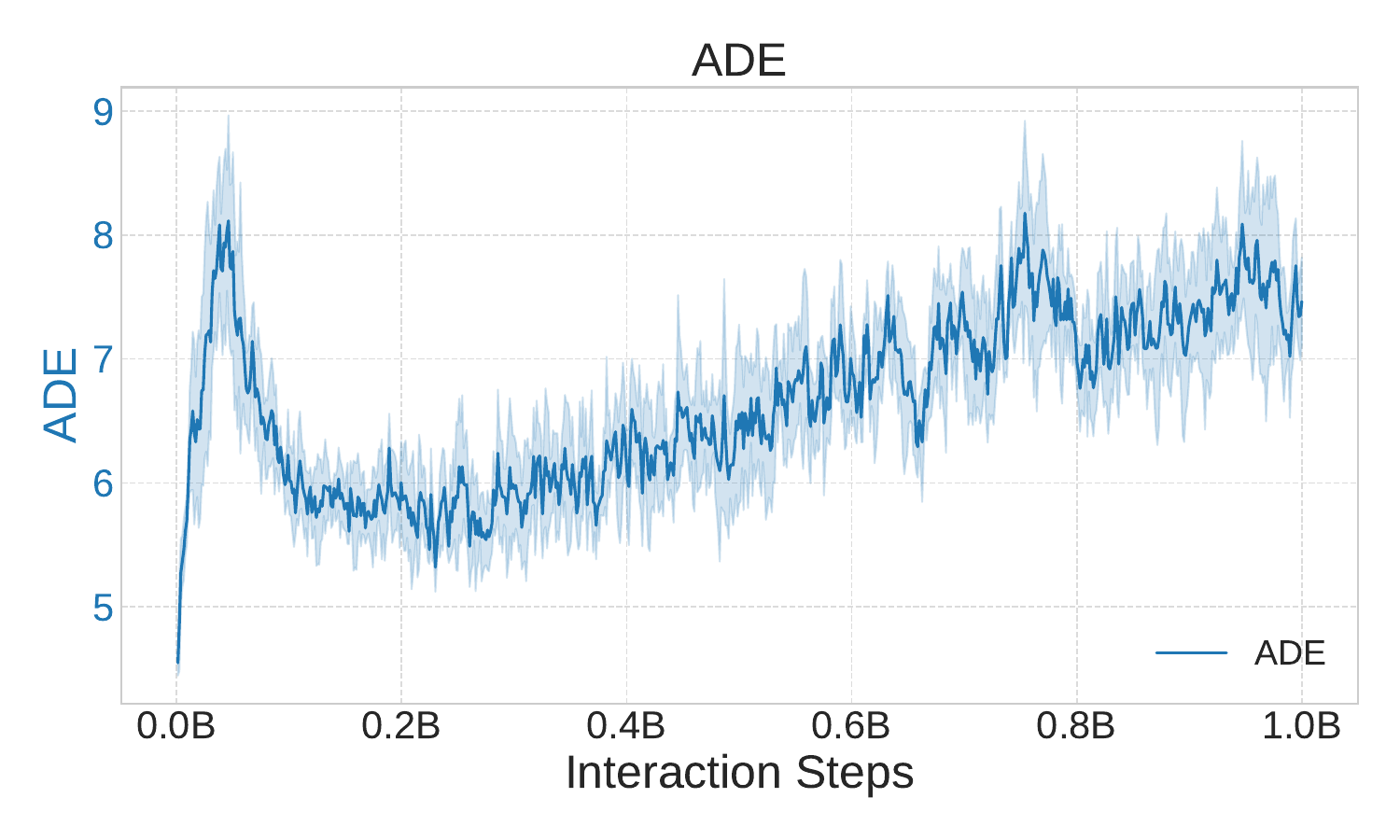}
    \caption{Average displacement error (ADE)}
    \label{fig:3b}
  \end{subfigure}

  \caption{\textbf{Detailed sub-metrics in realism metric and ADE in the experiment of transferring from Boston to Singapore.}
  Although the BC policy (at the beginning of training) achieves reasonable kinematic smoothness, it frequently violates map constraints (e.g., off-road driving), leading to a poor map-based realism meta score.
  In our experiments, behavior cloning serves as a canonical source-city planner. 
  While its absolute performance depends on dataset coverage, our focus is on the performance degradation induced by the lack of supervised finetuning in the target city and the ability of NOMAD to mitigate it without demonstrations.
  }
  \label{fig:detailed_metrics_boston_to_sin}
\end{figure*}

Following \cref{sec:extendedmetrics}, and taking the transfer from Boston to Singapore as an example, \cref{fig:detailed_metrics_boston_to_sin} decomposes the realism meta score into its constituent components: kinematic metrics, interactive metrics, and map-based metrics. The kinematic score, which captures velocity and acceleration profiles, exhibits a declining trend as training progresses, decreasing from approximately 0.51 to 0.42. This suggests that self-play improvements in success rate and the overall realism meta score come with some degradation in kinematics-related metrics. 

In contrast, the interactive metrics, measuring behaviors such as time-to-collision and collision avoidance, remain relatively stable throughout training, indicating that the adapted policy preserves safe interaction patterns learned from the source city. Most notably, the map-based metrics show substantial improvement, rising from approximately 0.55 to above 0.87, which reflects the policy's increasing ability to comply with Singapore's distinct road geometries and lane structures. Finally, the average displacement error (ADE) improves substantially from approximately 9 to around 7. Lower ADE values indicate that the planned trajectory stays closer to the ground-truth trajectory on average over time. Note that ADE is not a good metric for closed-loop evaluation, since completely imitating expert trajectories is not our goal. 

Overall, these results suggest that effective city adaptation via map-based self-play improves the realism scores in most of the realism sub-metrics. However, this decomposition also highlights a trade-off: while NOMAD improves the overall realism meta score and reduces ADE, it comes with a modest degradation in kinematics metrics. This suggests that matching the target-city trajectory structure is easier than fully recovering speed and acceleration profiles. These kinematics gaps could potentially be mitigated through more targeted reward design, which is beyond the scope of the study.

\begin{figure*}[htbp]
  \centering
  \begin{subfigure}[b]{0.48\textwidth}
    \centering
    \includegraphics[width=\linewidth]{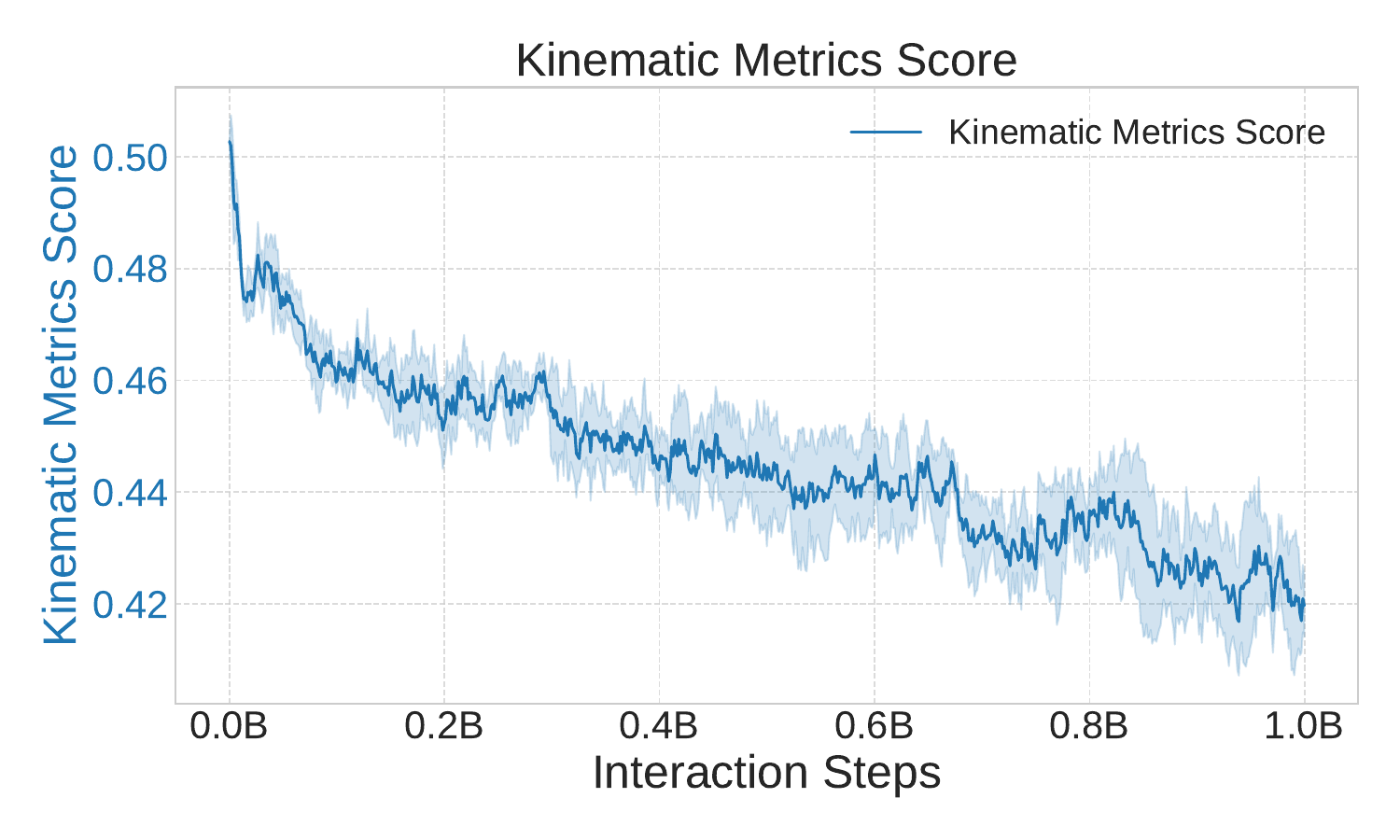}
    \caption{Kinematics metric}
    \label{fig:1a}
  \end{subfigure}
  \hfill
  \begin{subfigure}[b]{0.48\textwidth}
    \centering
    \includegraphics[width=\linewidth]{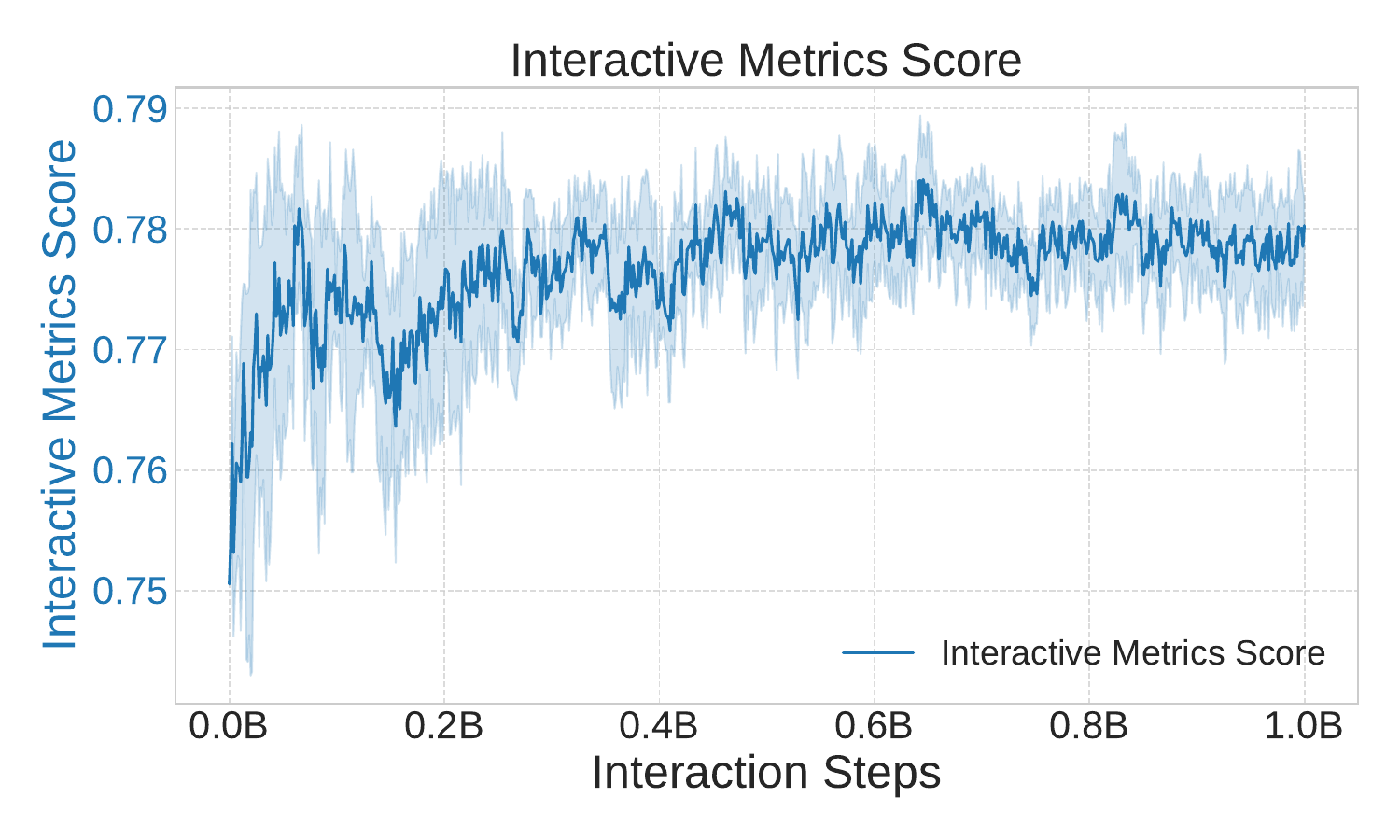}
    \caption{Interactive metric}
    \label{fig:1b}
  \end{subfigure}

  \vspace{1ex}

  \begin{subfigure}[b]{0.48\textwidth}
    \centering
    \includegraphics[width=\linewidth]{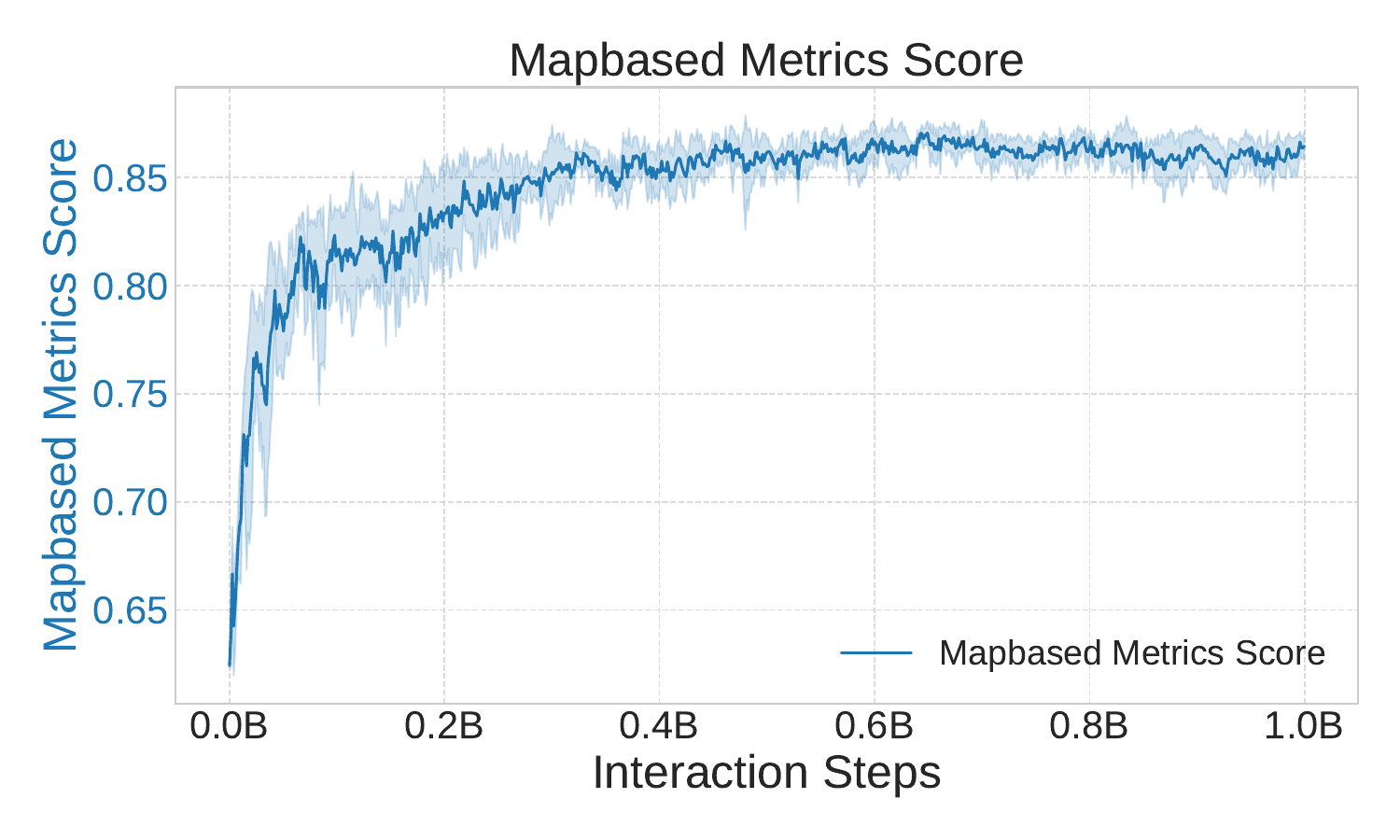}
    \caption{Map-based metric}
    \label{fig:2a}
  \end{subfigure}
  \hfill
  \begin{subfigure}[b]{0.48\textwidth}
    \centering
    \includegraphics[width=\linewidth]{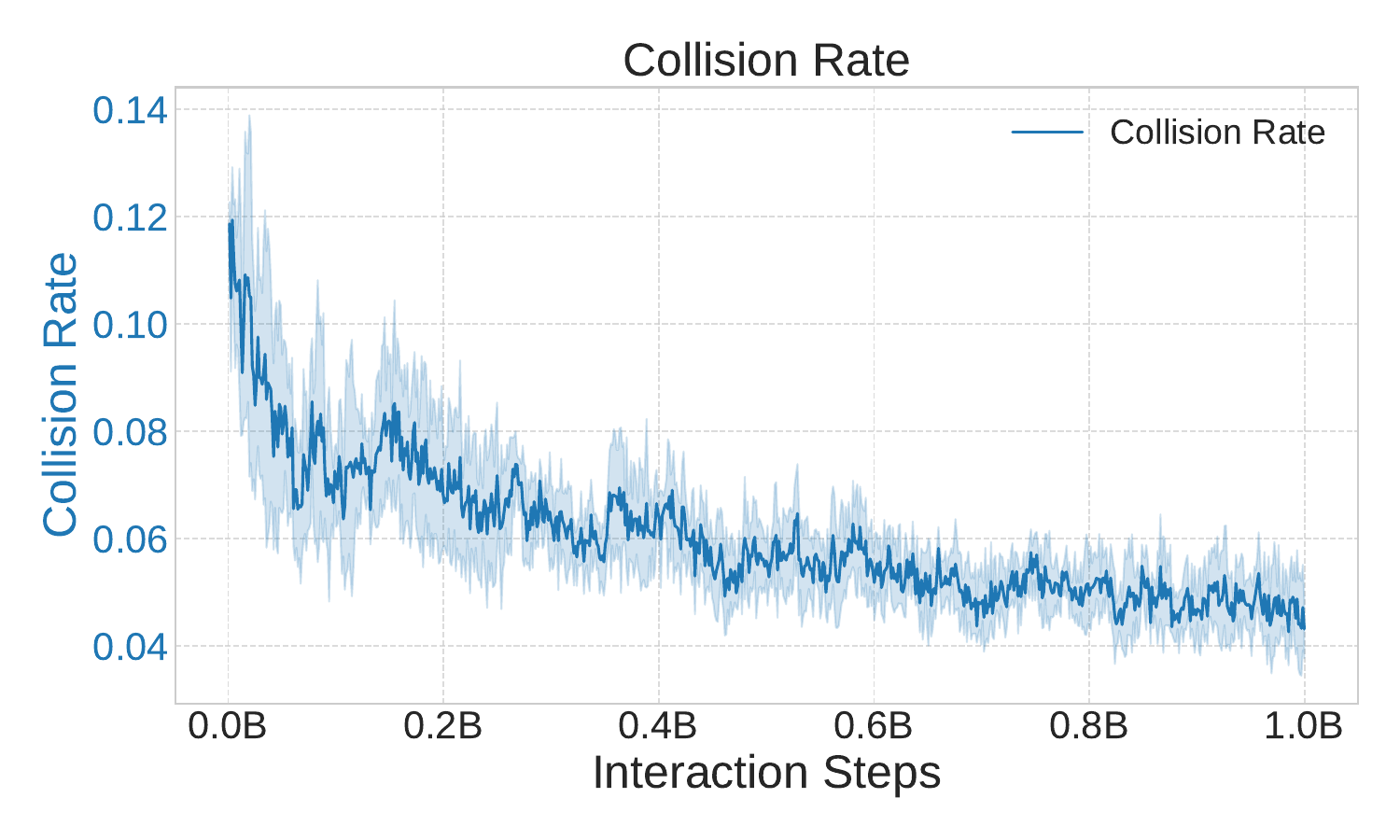}
    \caption{Collision rate}
    \label{fig:2b}
  \end{subfigure}

  \vspace{1ex}

  \begin{subfigure}[b]{0.48\textwidth}
    \centering
    \includegraphics[width=\linewidth]{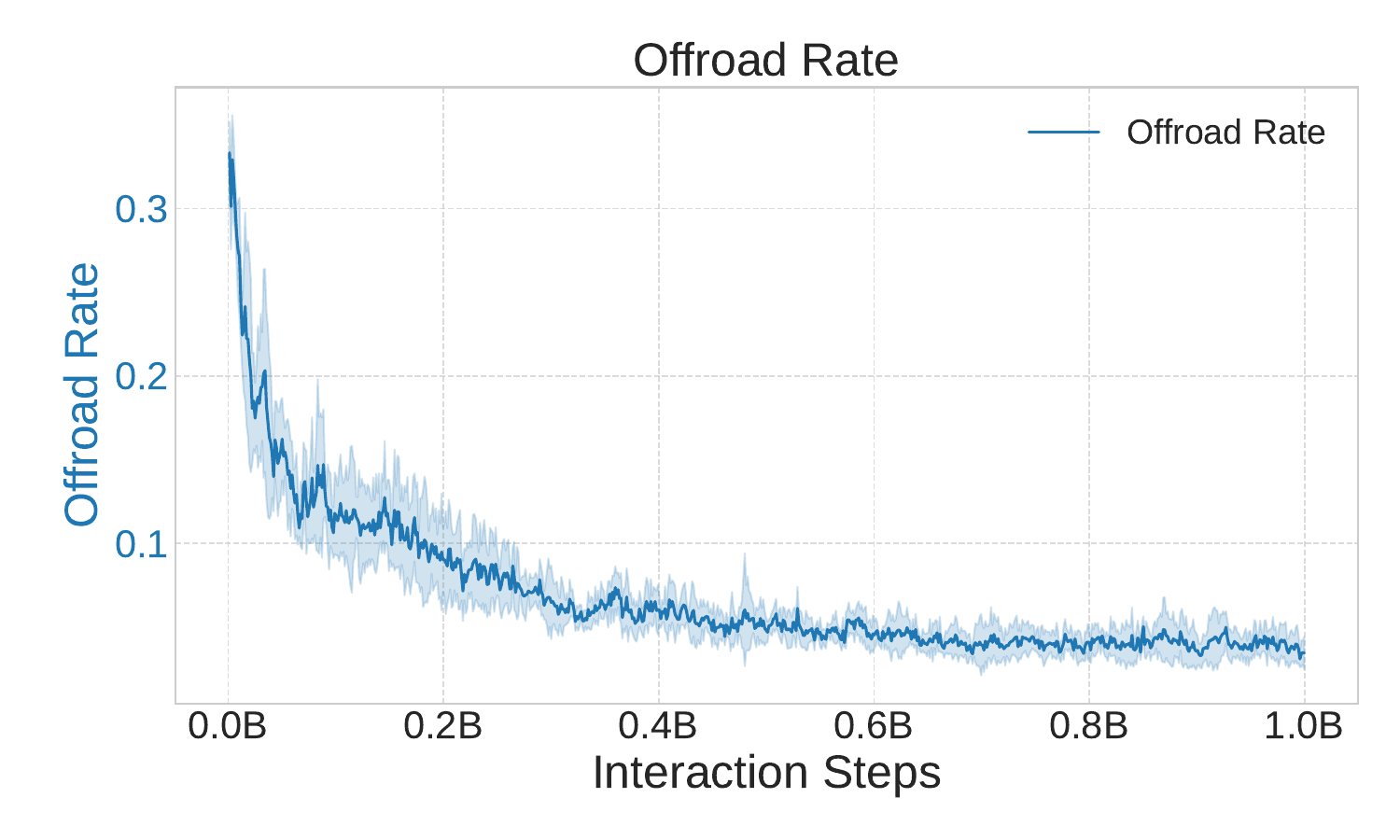}
    \caption{Offroad rate}
    \label{fig:3a}
 \end{subfigure}
  \hfill
  \begin{subfigure}[b]{0.48\textwidth}
    \centering
    \includegraphics[width=\linewidth]{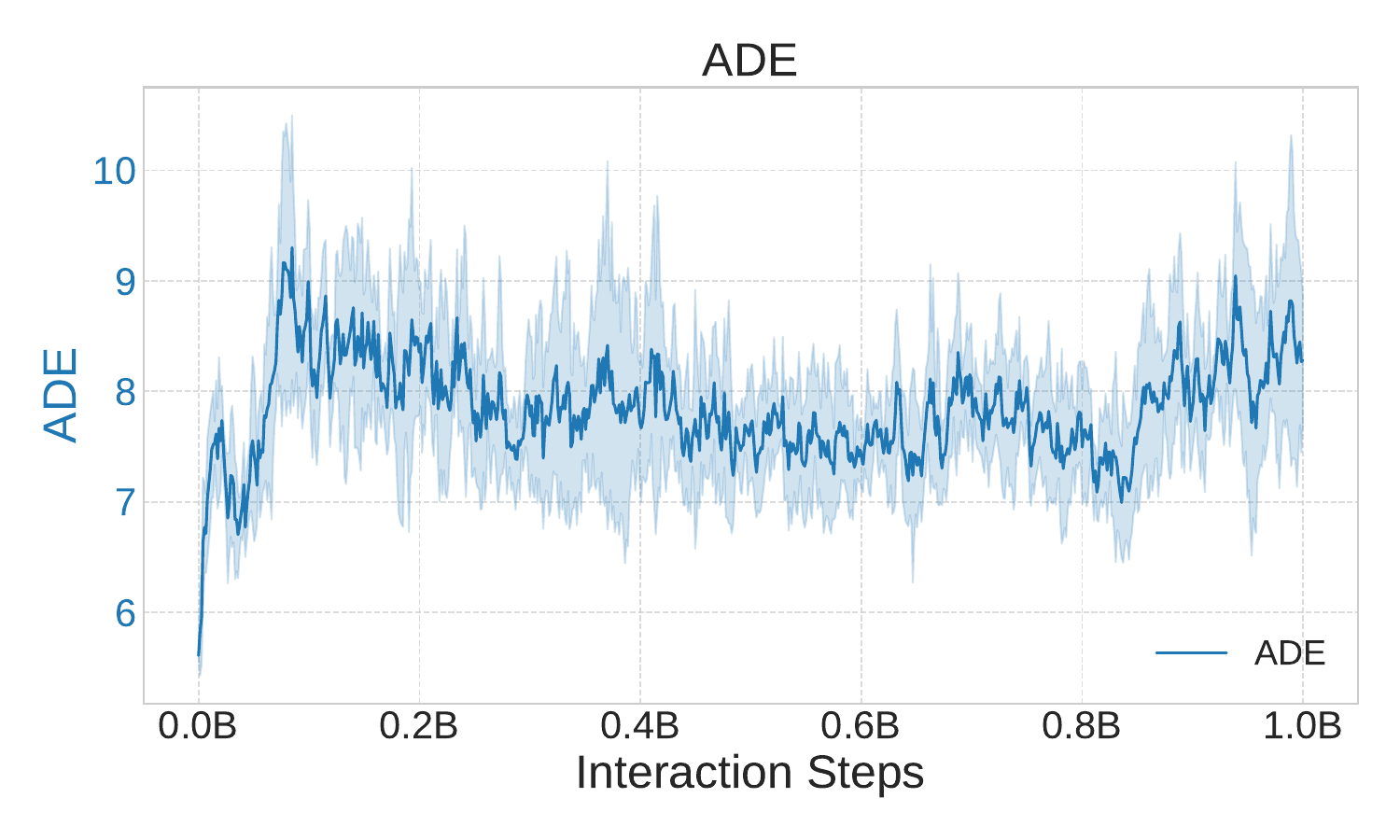}
    \caption{Average displacement error (ADE)}
    \label{fig:3b}
  \end{subfigure}

  \caption{Detailed sub-metrics in realism metric and ADE in the experiment of transferring from Singapore to Boston.}
  \label{fig:detailed_metrics_sin_to_bos}
\end{figure*}

\begin{figure*}[htbp]
  \centering
  \begin{subfigure}[b]{0.48\textwidth}
    \centering
    \includegraphics[width=\linewidth]{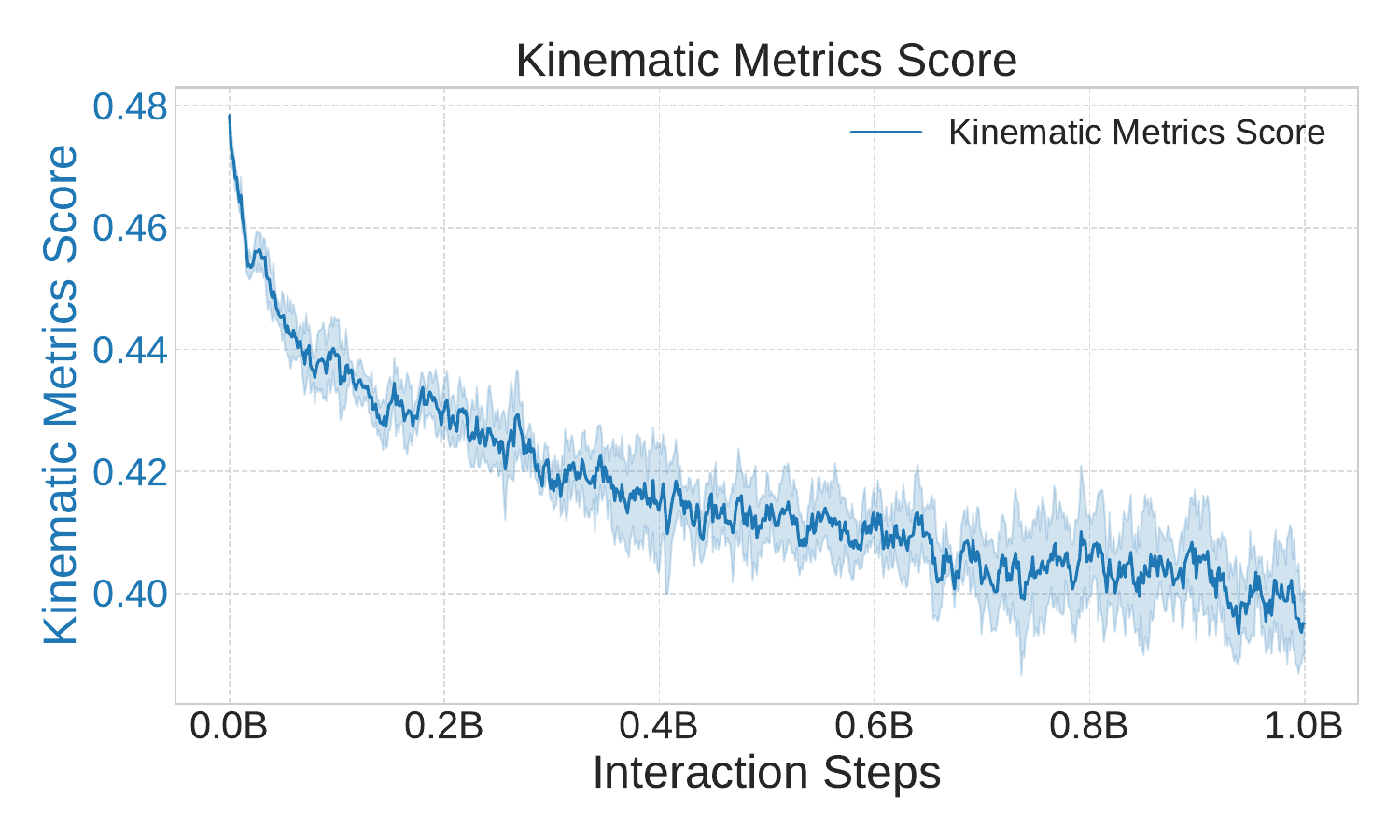}
    \caption{Kinematics metric}
    \label{fig:1a}
  \end{subfigure}
  \hfill
  \begin{subfigure}[b]{0.48\textwidth}
    \centering
    \includegraphics[width=\linewidth]{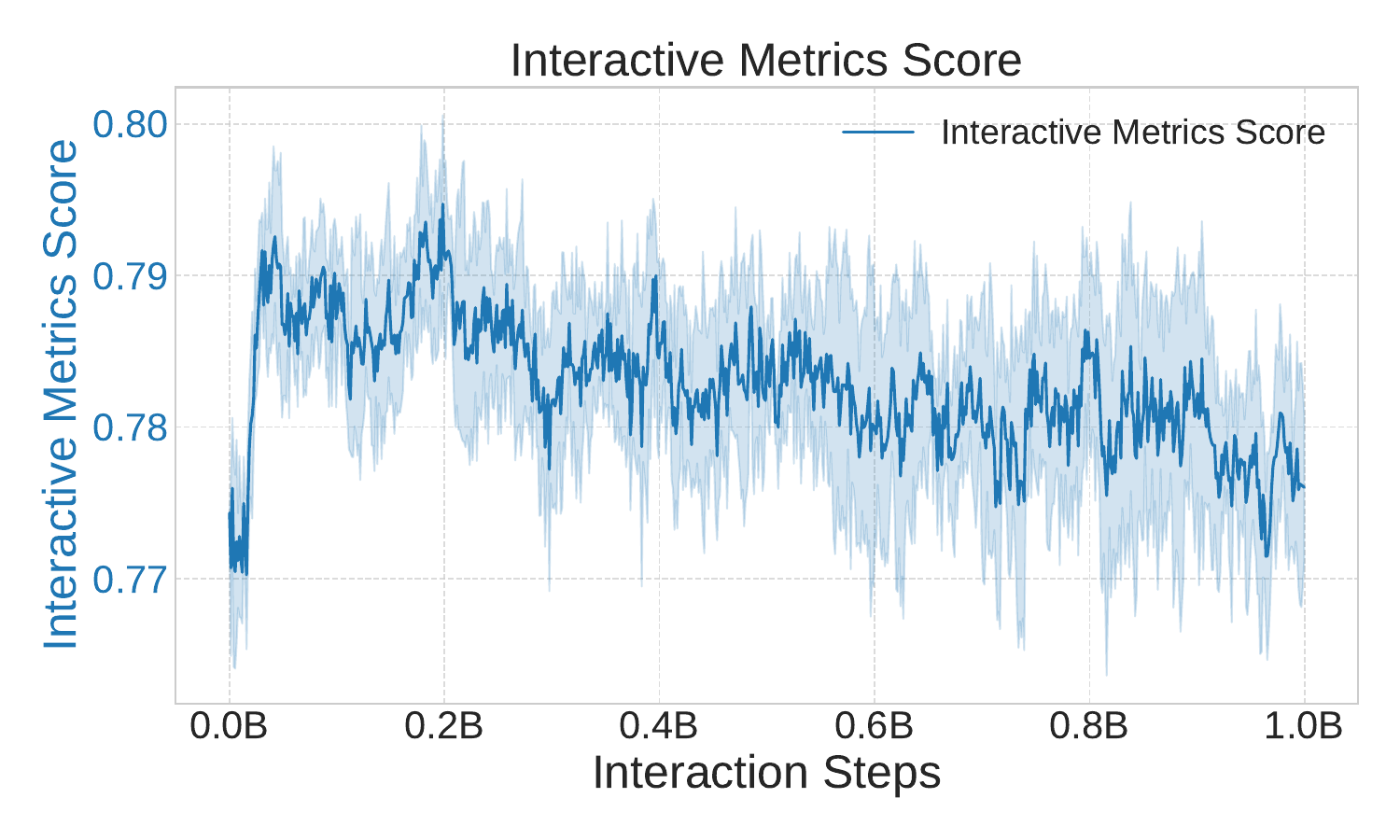}
    \caption{Interactive metric}
    \label{fig:1b}
  \end{subfigure}

  \vspace{1ex}

  \begin{subfigure}[b]{0.48\textwidth}
    \centering
    \includegraphics[width=\linewidth]{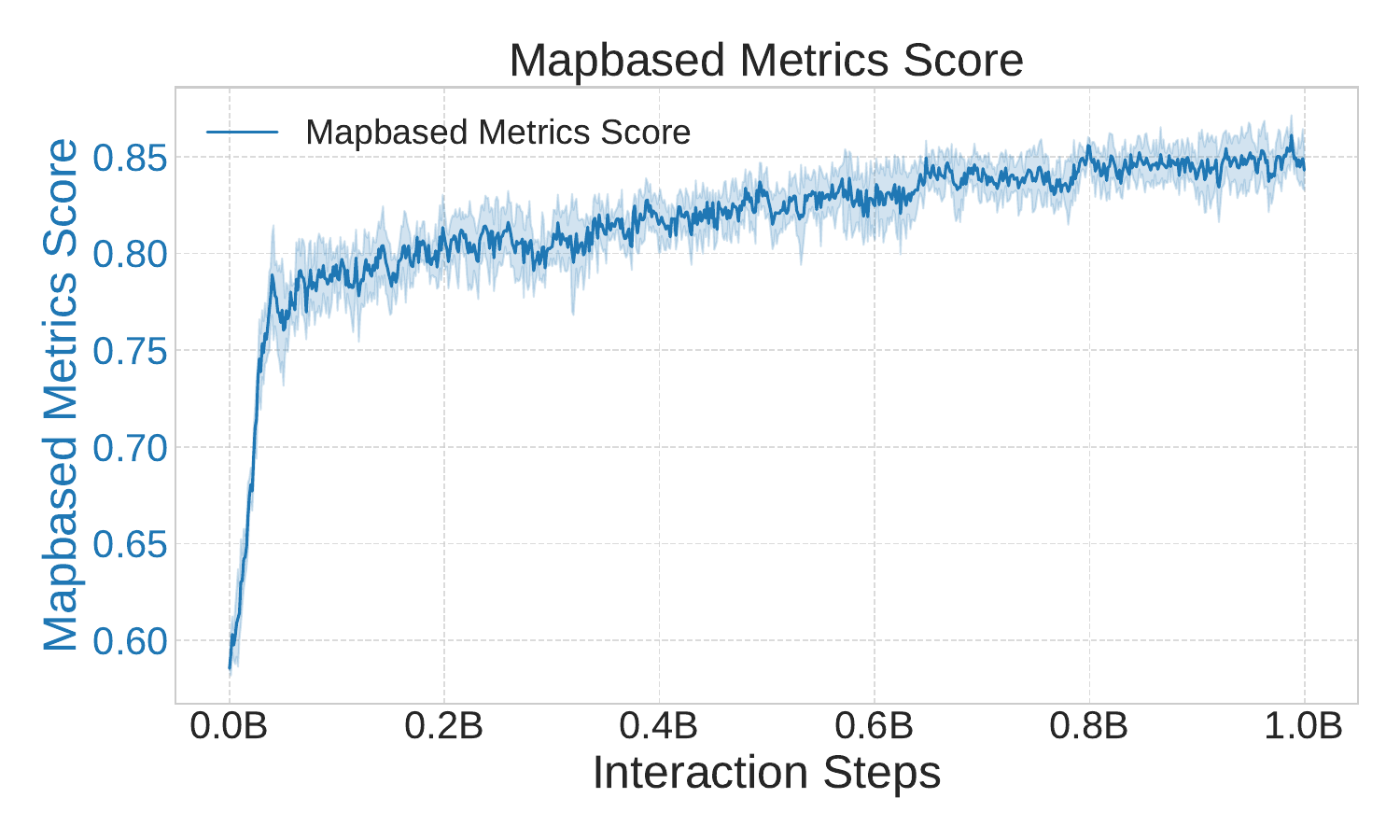}
    \caption{Map-based metric}
    \label{fig:2a}
  \end{subfigure}
  \hfill
  \begin{subfigure}[b]{0.48\textwidth}
    \centering
    \includegraphics[width=\linewidth]{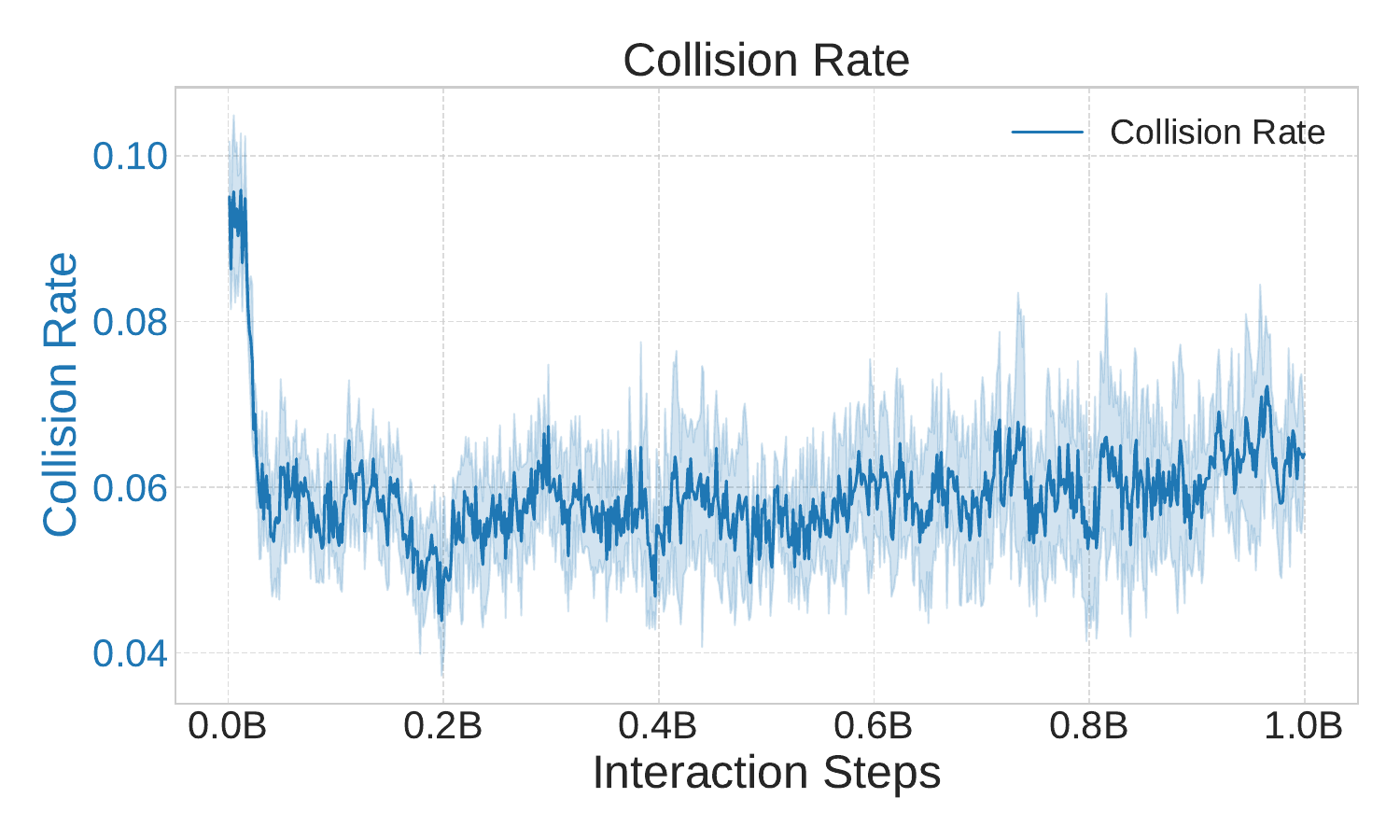}
    \caption{Collision rate}
    \label{fig:2b}
  \end{subfigure}

  \begin{subfigure}[b]{0.48\textwidth}
    \centering
    \includegraphics[width=\linewidth]{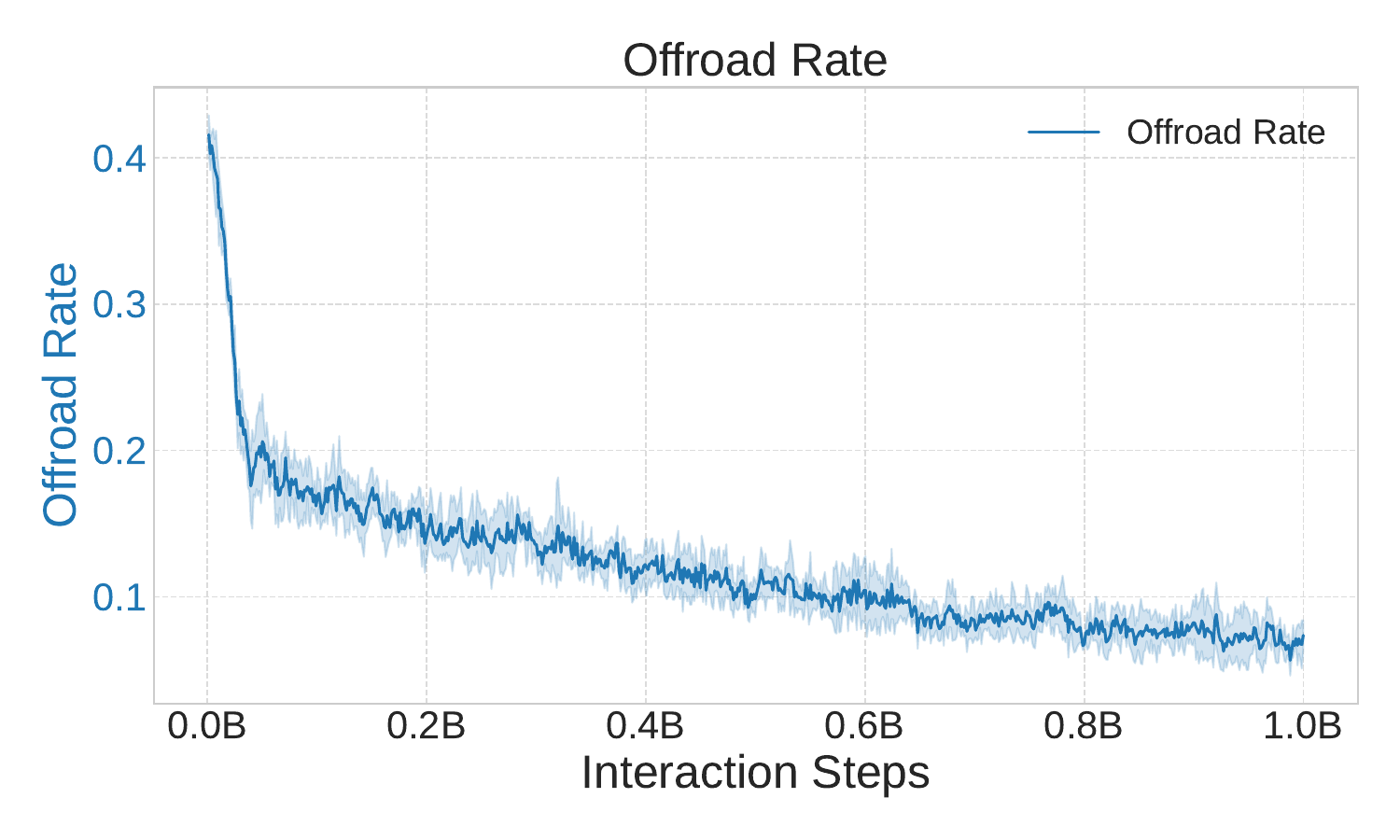}
    \caption{Offroad rate}
    \label{fig:3a}
  \end{subfigure}
  \hfill
  \begin{subfigure}[b]{0.48\textwidth}
    \centering
    \includegraphics[width=\linewidth]{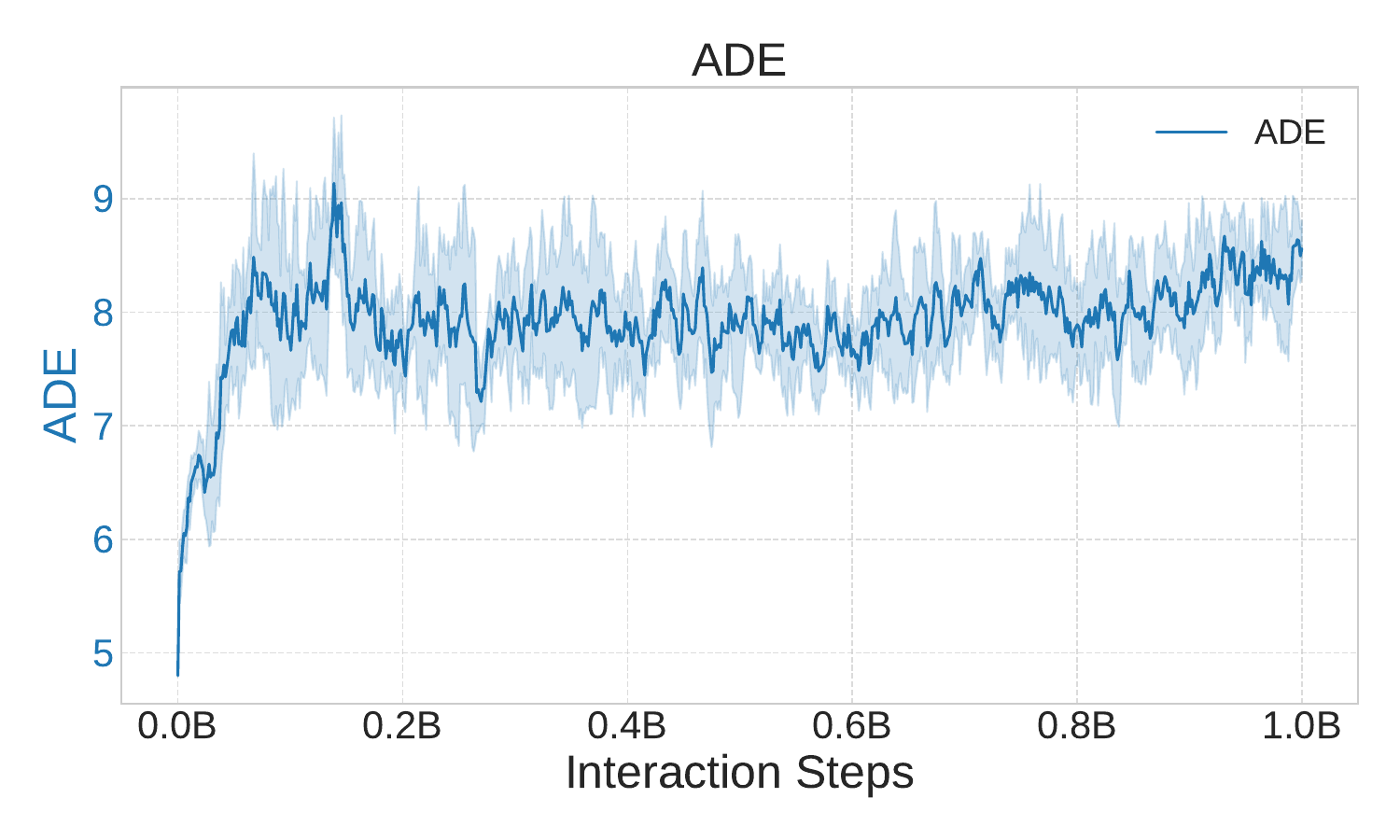}
    \caption{Average displacement error (ADE)}
    \label{fig:3b}
  \end{subfigure}

  \caption{Detailed sub-metrics in realism metric and ADE in the experiment of transferring from Singapore to Pittsburgh.}
  \label{fig:detailed_metrics_sin_to_pitts}
\end{figure*}

\begin{figure*}[htbp]
  \centering
  \begin{subfigure}[b]{0.48\textwidth}
    \centering
    \includegraphics[width=\linewidth]{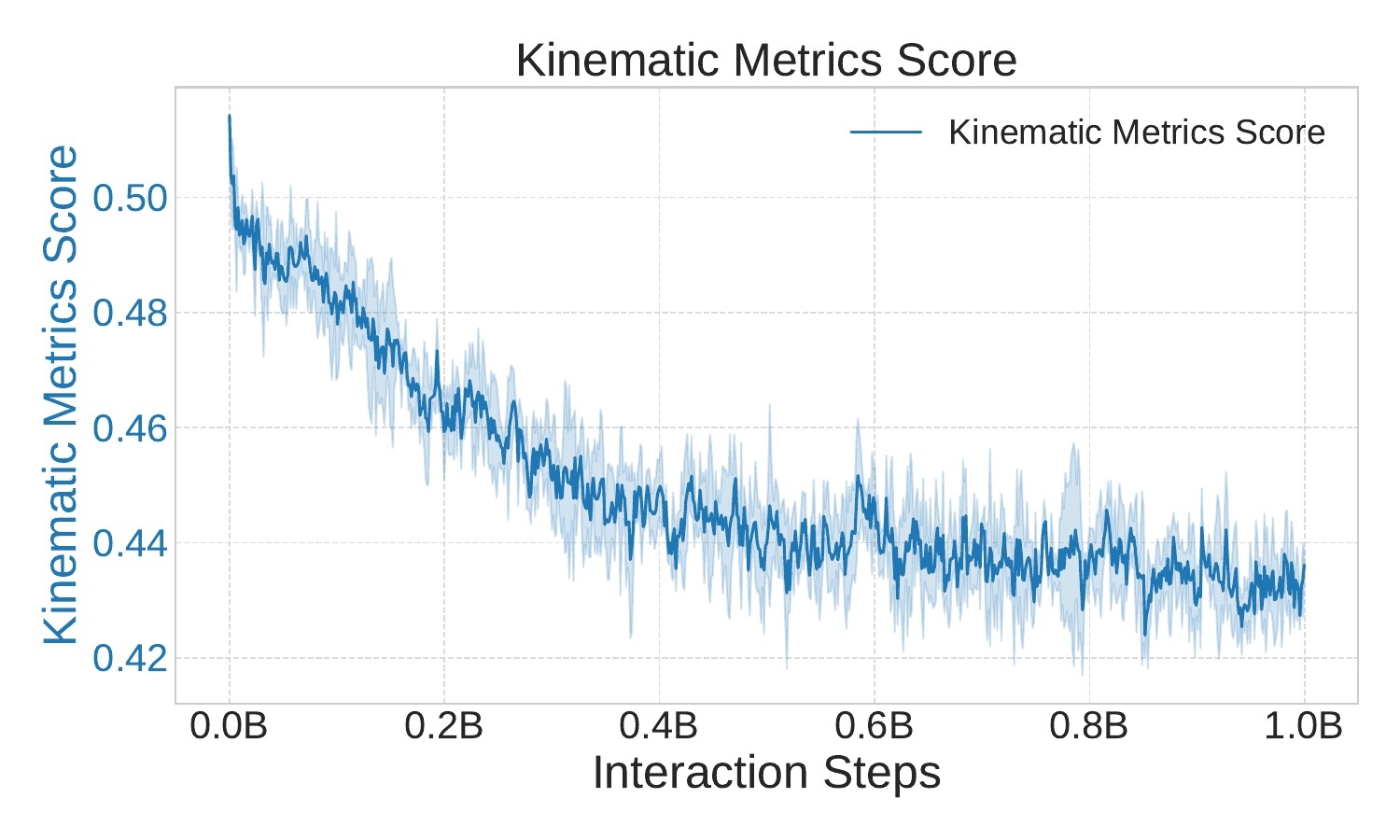}
    \caption{Kinematics metric}
    \label{fig:1a}
  \end{subfigure}
  \hfill
  \begin{subfigure}[b]{0.48\textwidth}
    \centering
    \includegraphics[width=\linewidth]{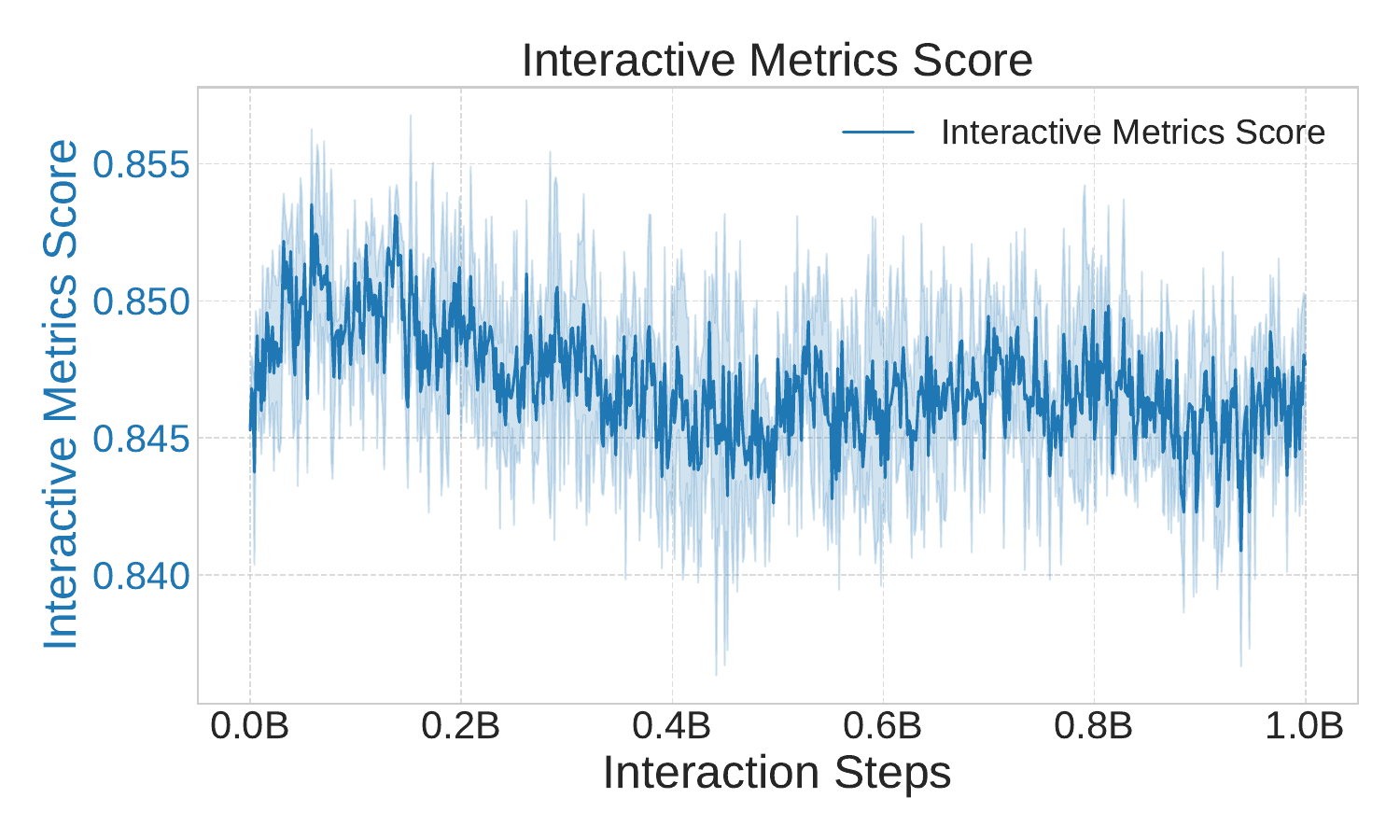}
    \caption{Interactive metric}
    \label{fig:1b}
  \end{subfigure}

  \vspace{1ex}

  \begin{subfigure}[b]{0.48\textwidth}
    \centering
    \includegraphics[width=\linewidth]{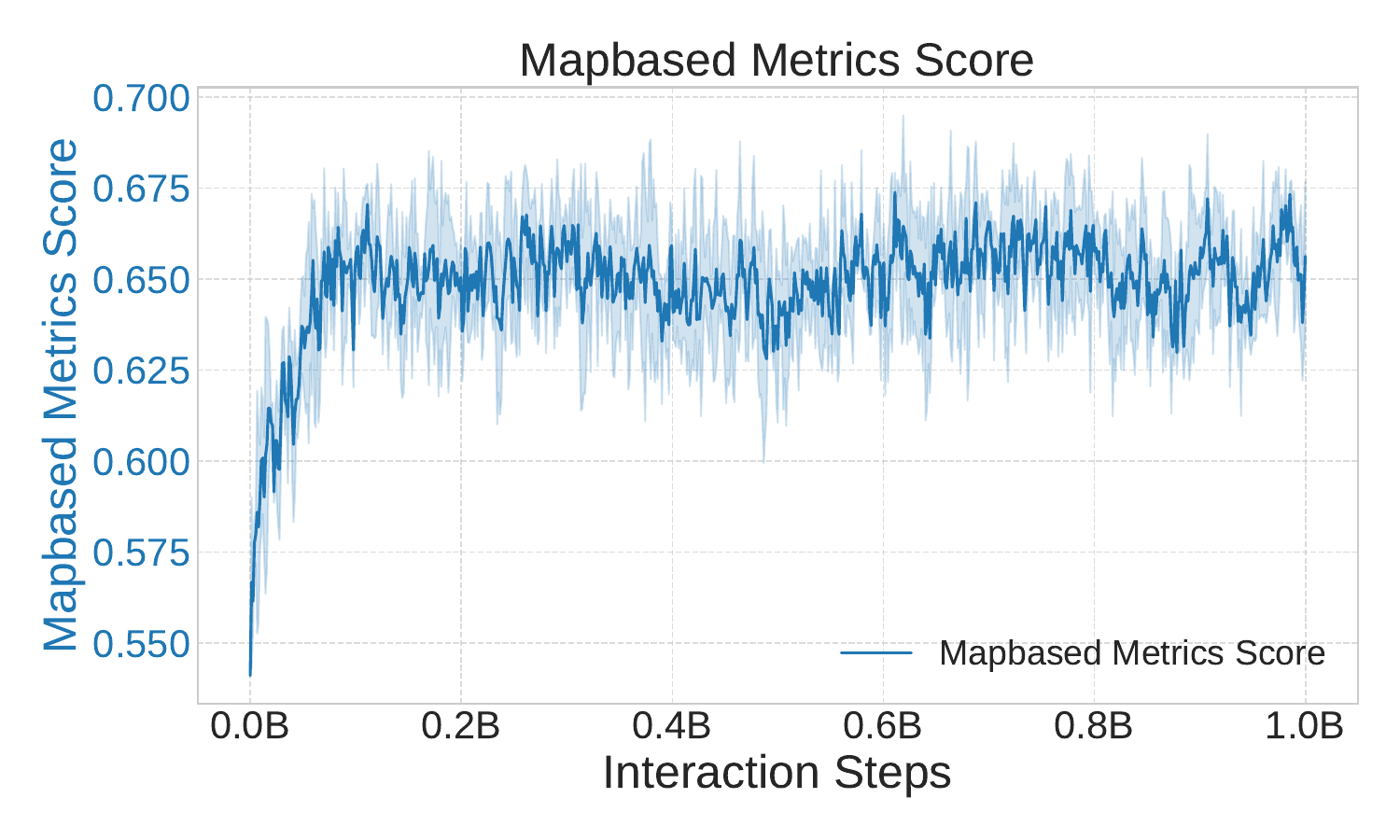}
    \caption{Map-based metric}
    \label{fig:2a}
  \end{subfigure}
  \hfill
  \begin{subfigure}[b]{0.48\textwidth}
    \centering
    \includegraphics[width=\linewidth]{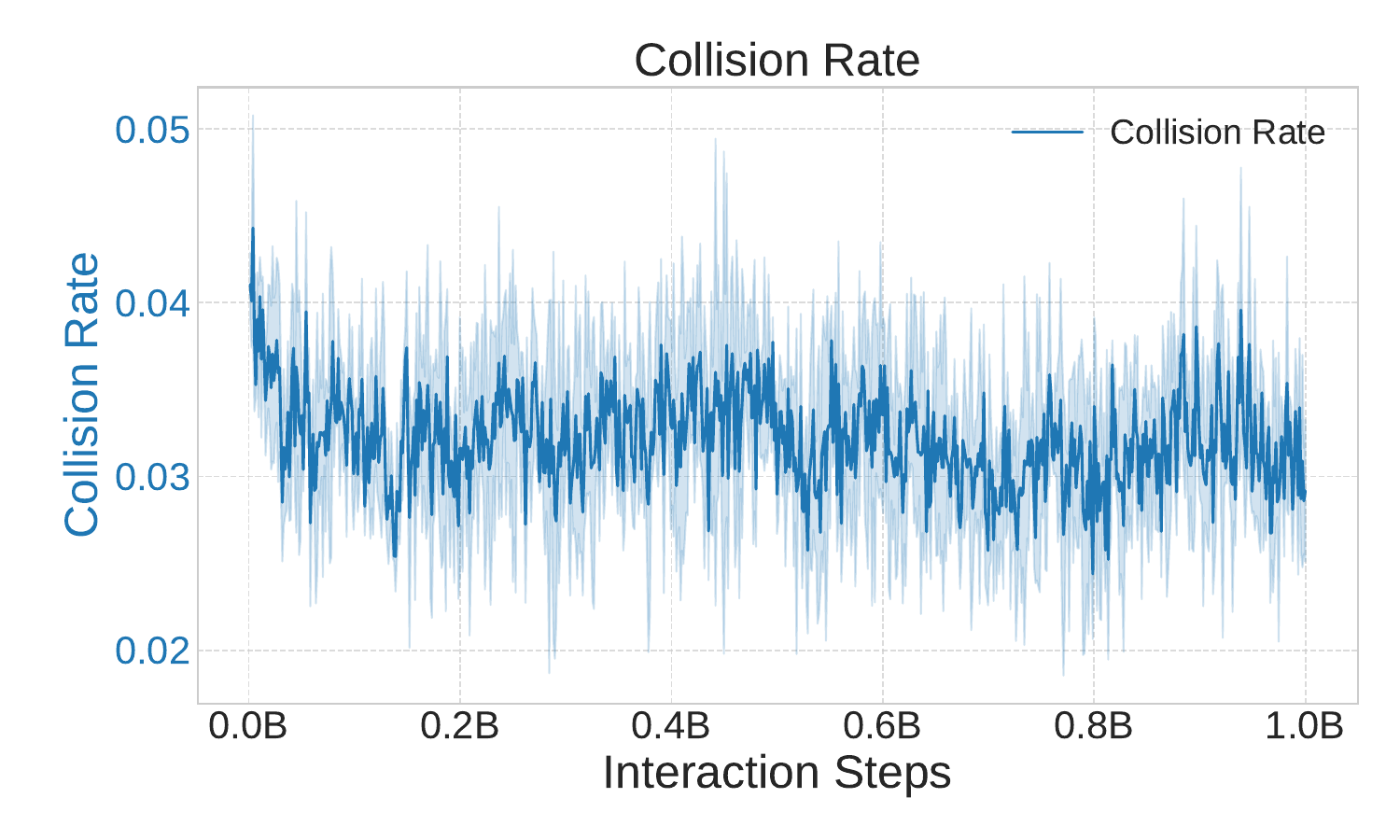}
    \caption{Collision rate}
    \label{fig:2a}
  \end{subfigure}

    \begin{subfigure}[b]{0.48\textwidth}
    \centering
    \includegraphics[width=\linewidth]{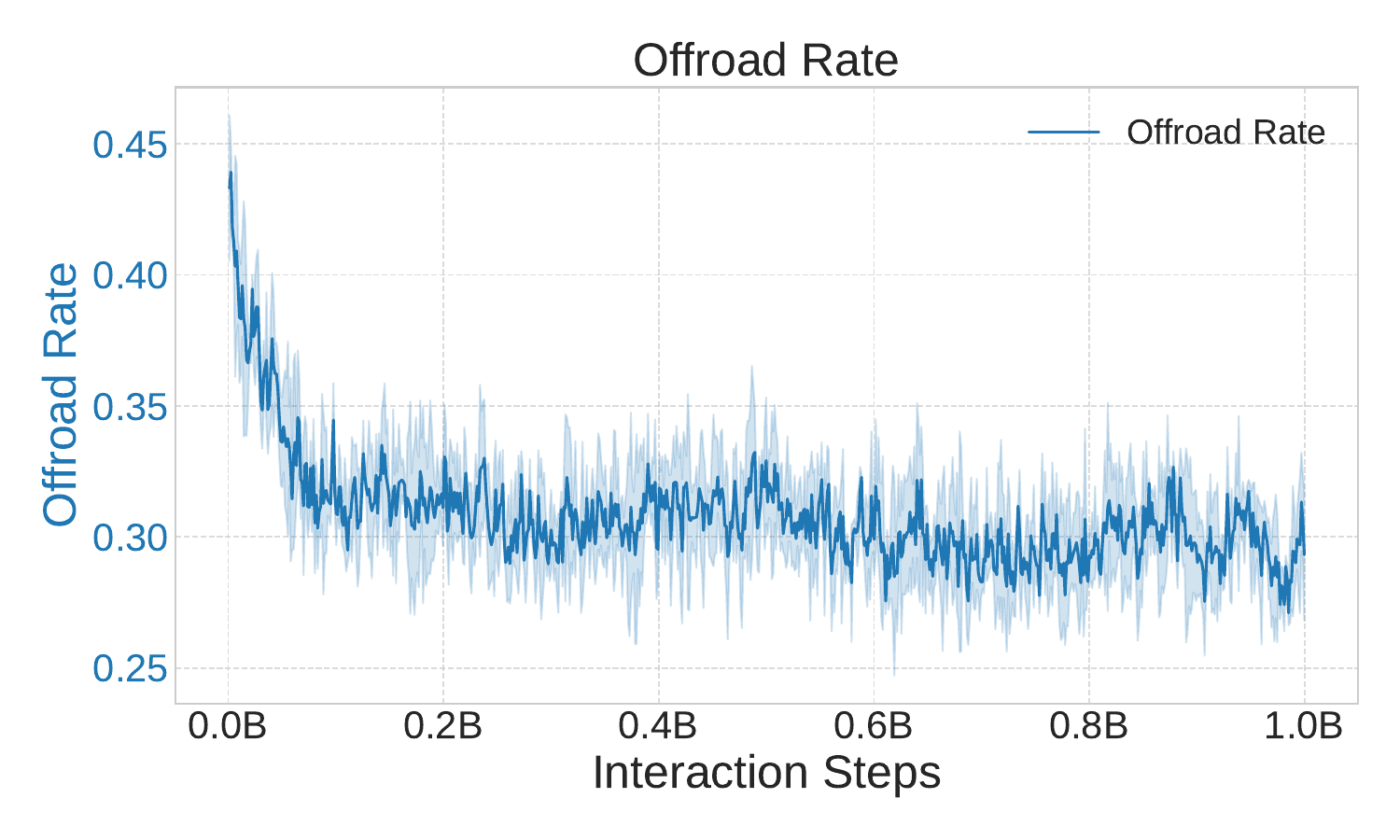}
    \caption{Offroad rate}
    \label{fig:3a}
  \end{subfigure}
  \hfill
  \begin{subfigure}[b]{0.48\textwidth}
    \centering
    \includegraphics[width=\linewidth]{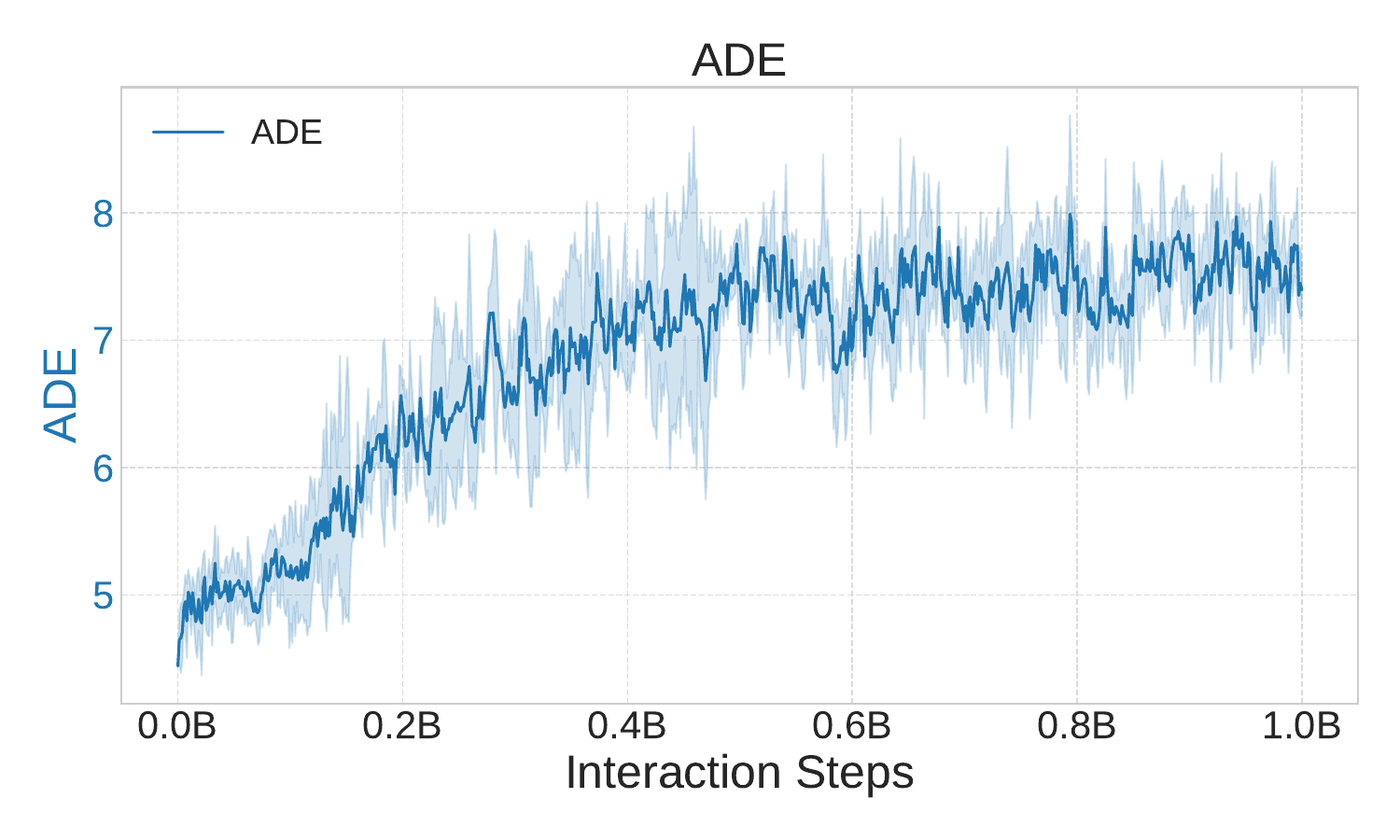}
    \caption{Average displacement error (ADE)}
    \label{fig:3b}
  \end{subfigure}

  \caption{Detailed realism sub-metrics and ADE for behavior cloning followed by self-play MARL in Boston, evaluated zero-shot in Singapore.}
  \label{fig:detailed_metrics_bos_to_bos}
\end{figure*}

\clearpage
\section{Influence of Behavior Priors on Sub-Metrics}

\begin{figure*}[htbp]
  \centering
  \begin{subfigure}[b]{0.48\textwidth}
    \centering
    \includegraphics[width=\linewidth]{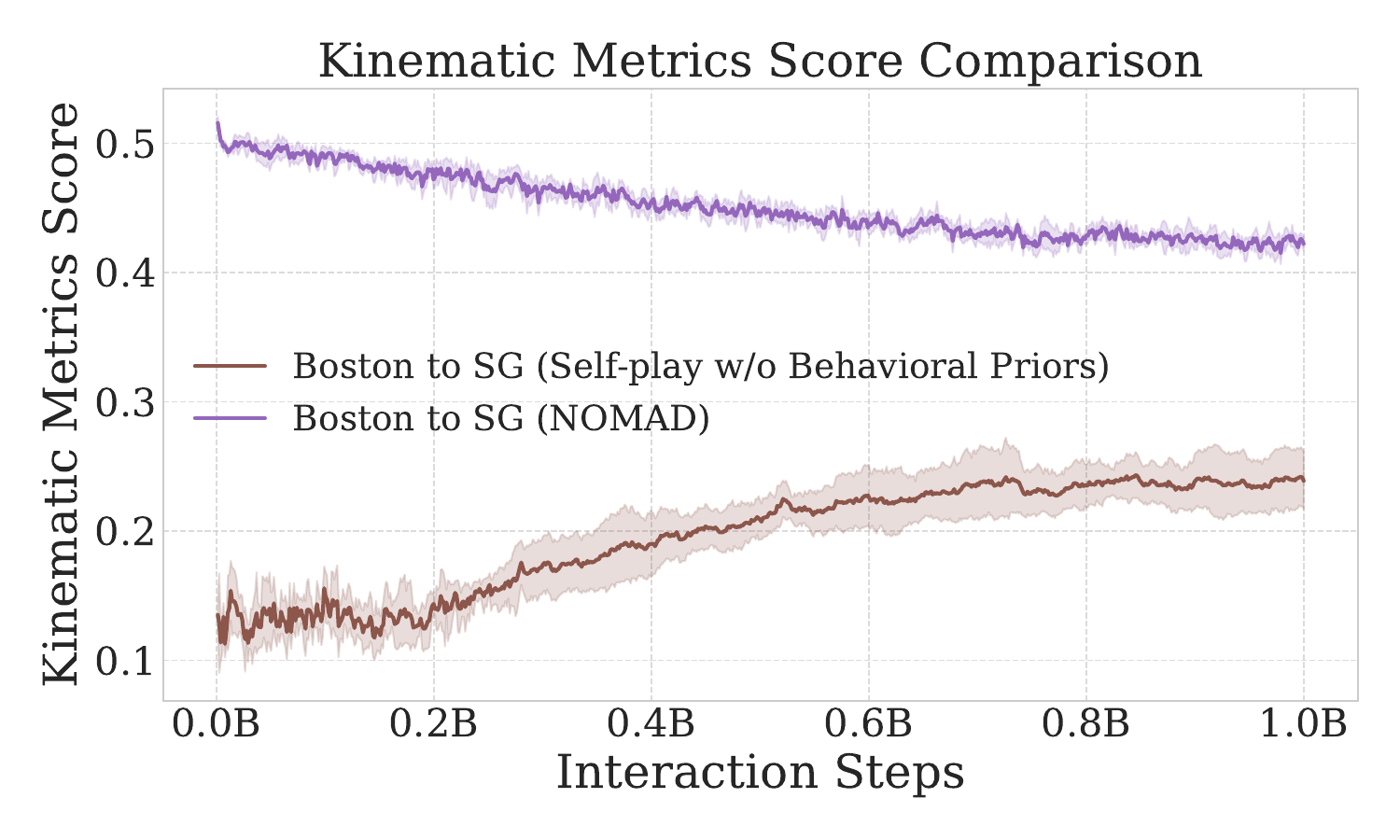}
    \caption{Kinematics metric}
    \label{fig:1a}
  \end{subfigure}
  \hfill
  \begin{subfigure}[b]{0.48\textwidth}
    \centering
    \includegraphics[width=\linewidth]{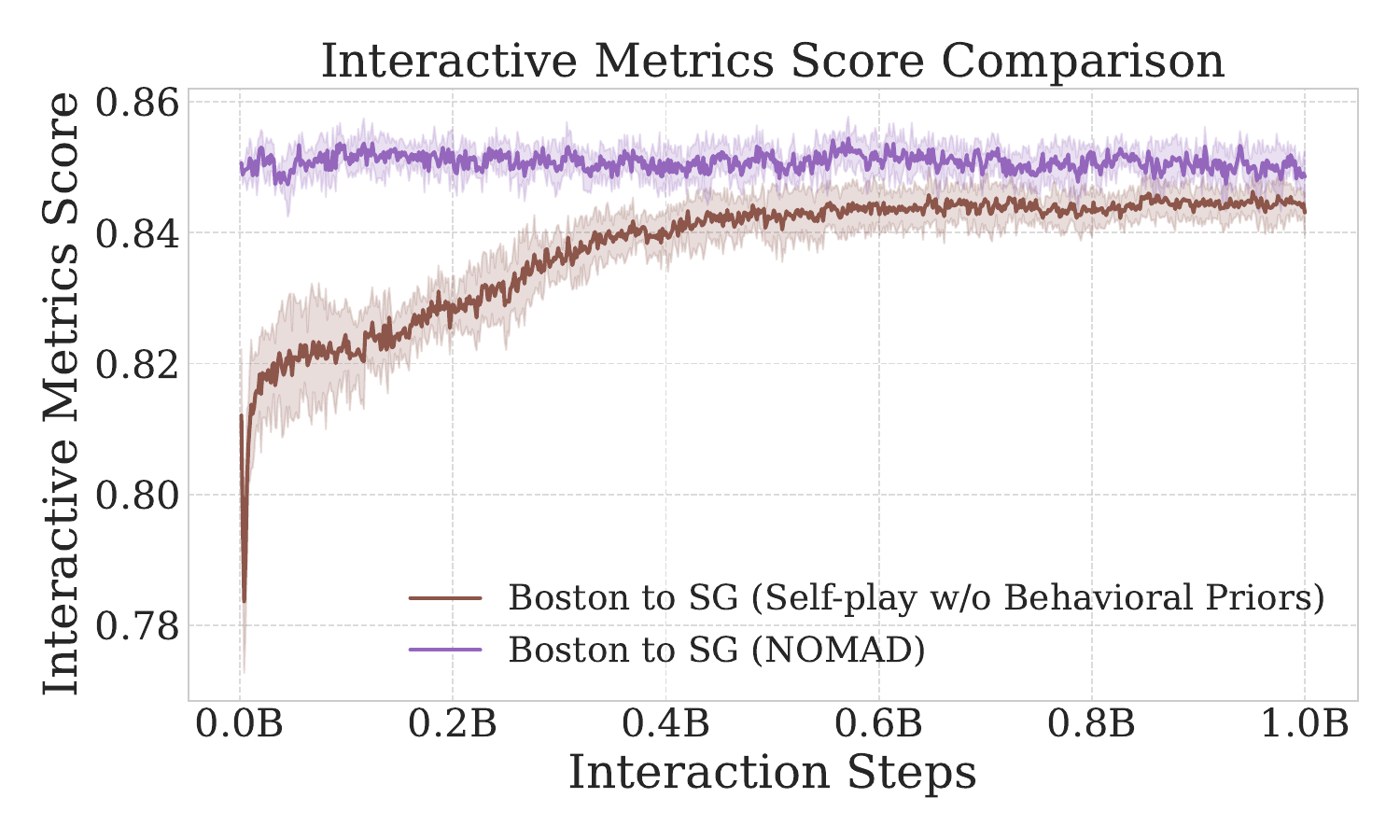}
    \caption{Interactive metric}
    \label{fig:1b}
  \end{subfigure}

  \vspace{1ex}

  \begin{subfigure}[b]{0.48\textwidth}
    \centering
    \includegraphics[width=\linewidth]{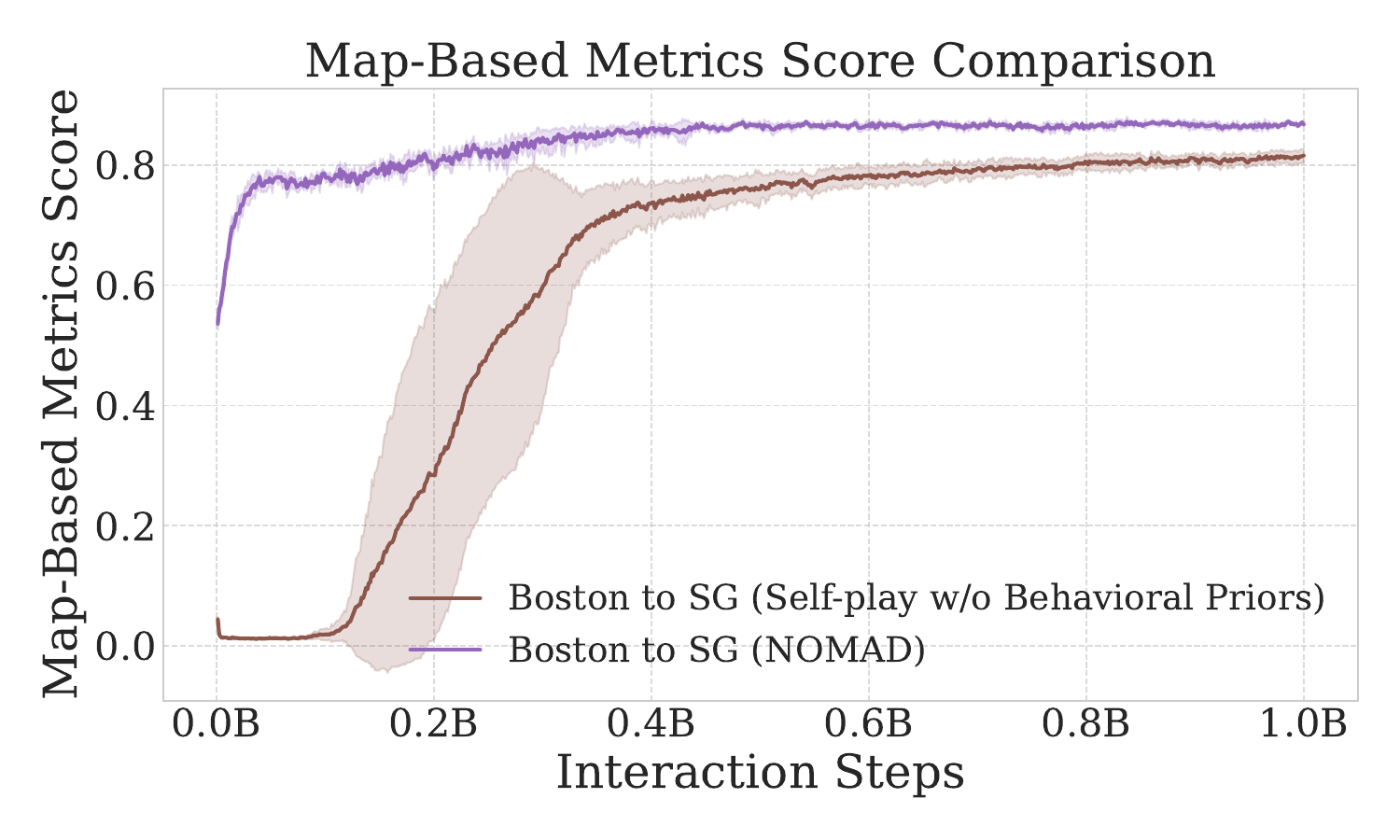}
    \caption{Map-based metric}
    \label{fig:2a}
  \end{subfigure}
  \hfill
  \begin{subfigure}[b]{0.48\textwidth}
    \centering
    \includegraphics[width=\linewidth]{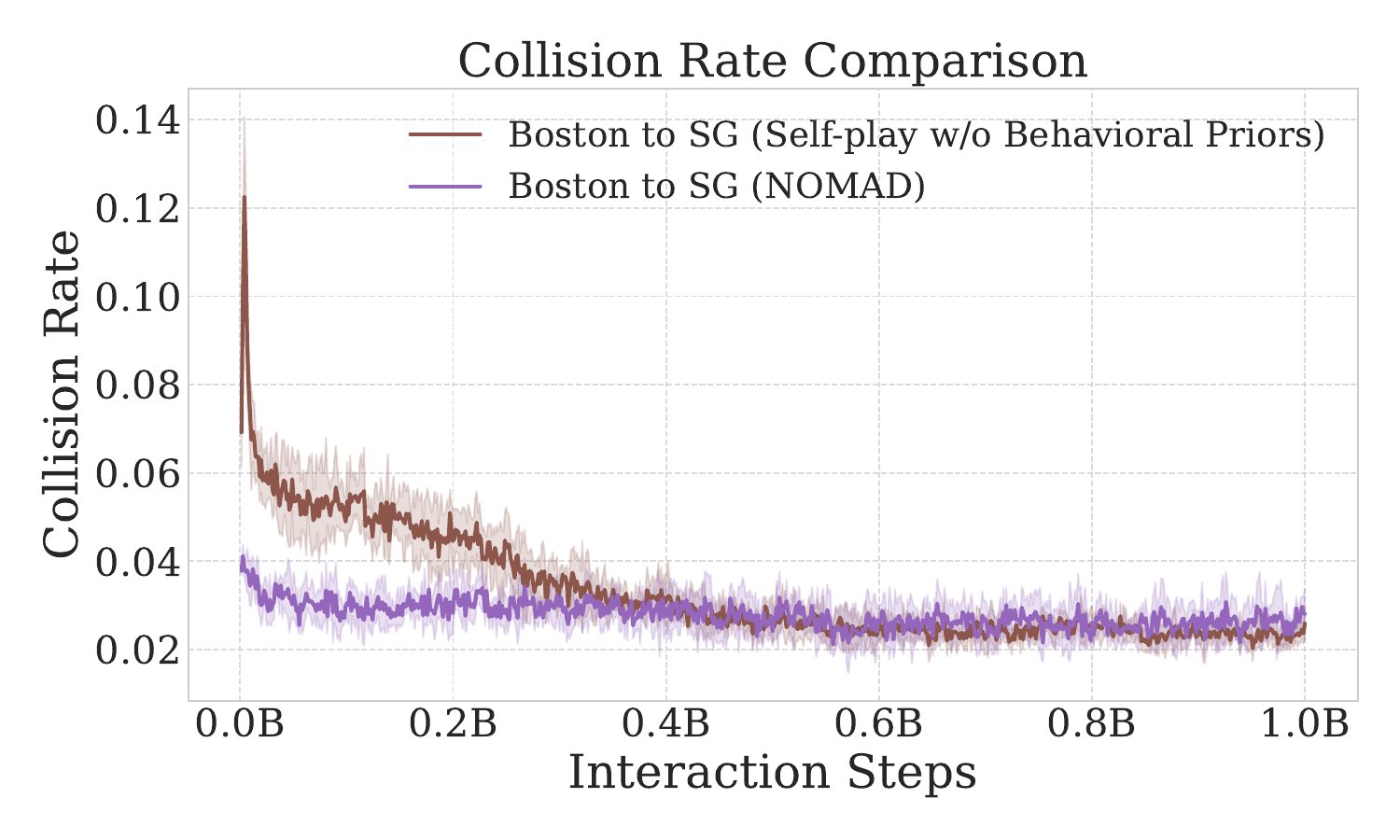}
    \caption{Collision rate}
    \label{fig:2a}
  \end{subfigure}

    \begin{subfigure}[b]{0.48\textwidth}
    \centering
    \includegraphics[width=\linewidth]{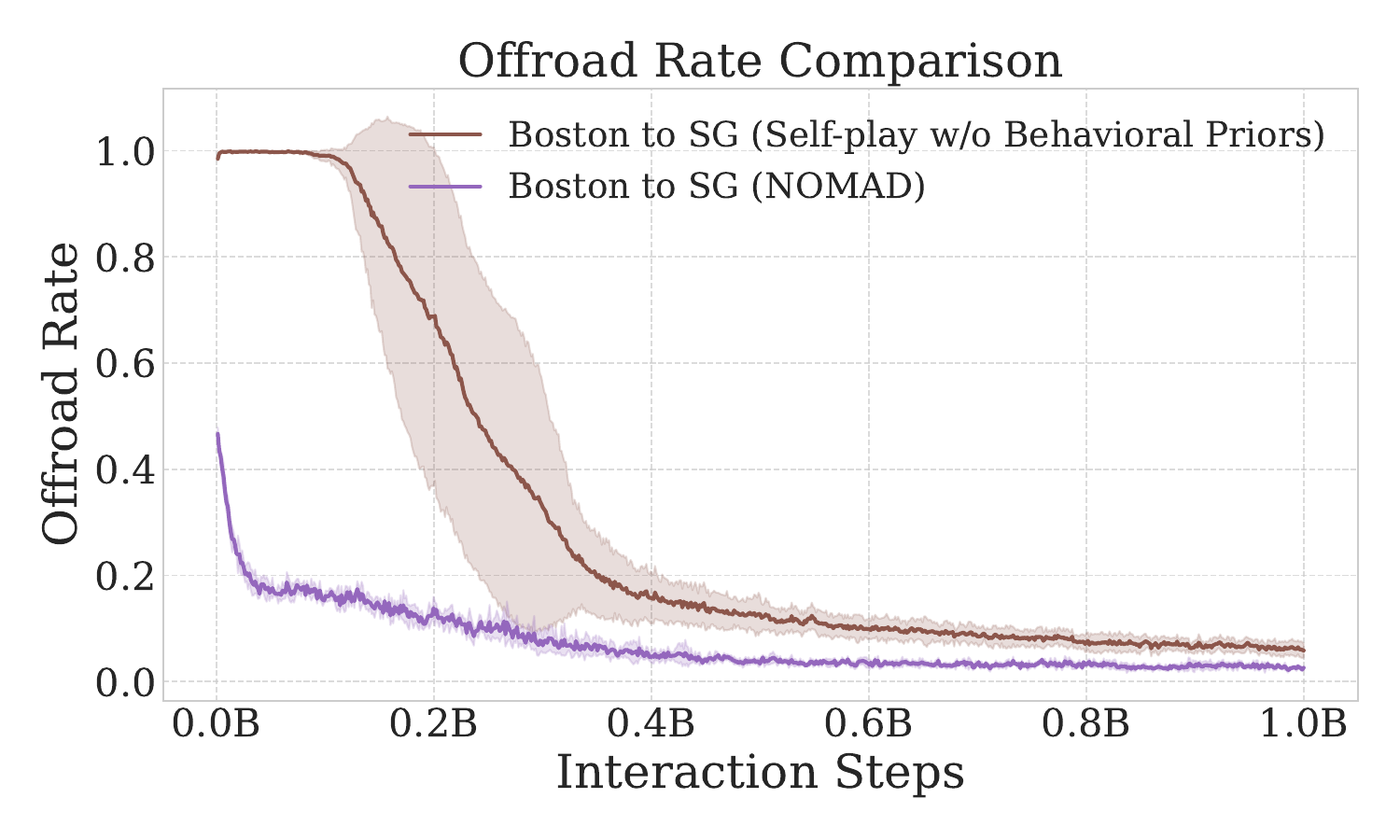}
    \caption{Offroad rate}
    \label{fig:3a}
  \end{subfigure}
  \hfill
  \begin{subfigure}[b]{0.48\textwidth}
    \centering
    \includegraphics[width=\linewidth]{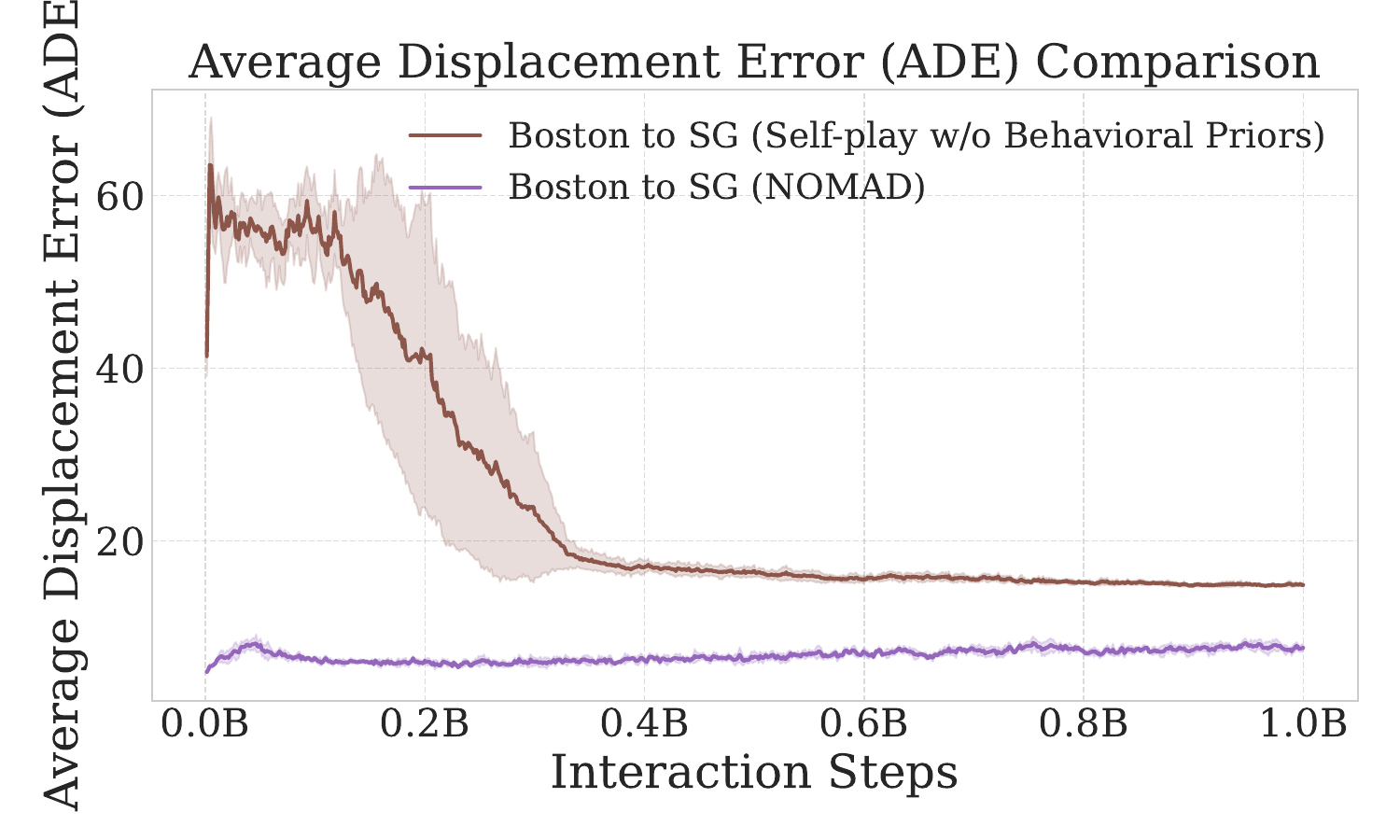}
    \caption{Average displacement error (ADE)}
    \label{fig:3b}
  \end{subfigure}

  \caption{\textbf{Comparison of all metrics between self-play with and without behavioral priors.}
  Behavioral priors play a critical role in stabilizing self-play training and guiding exploration toward human-like driving under the supervision of a simple reward function.
  }
  \label{fig:detailed_metrics_comparison}
\end{figure*}

\cref{fig:detailed_metrics_comparison} provides a detailed breakdown of training dynamics across individual evaluation metrics for map-based self-play with and without behavioral priors. 
While pure self-play (without BC initialization and regularization) is eventually able to improve interactive metrics and reduce collision rates, it consistently lags behind NOMAD in all the realism submetrics throughout training, as reflected by substantially lower kinematic and map-based scores.
These results further substantiate that behavioral priors play a critical role in stabilizing self-play training and guiding exploration toward human-like driving behaviors under the supervision of a simple reward function.

%% file: appendices/AdditionalResults.tex
\section{Extended Experimental Results}
\subsection{The Necessity of the Target-City Map}
To assess the necessity of the target-city map and the agent interactions within it, we introduce an additional baseline, where we perform behavior cloning followed by self-play MARL using \textit{logged scenarios} exclusively in the source city (Boston). Then, we evaluate the resulting policy zero-shot in the target city (Singapore). This baseline employs the same BC initialization and self-play training protocol as NOMAD but has no access to target-city maps or scenarios during training.

\begin{figure*}[htpb]
  \centering
  \includegraphics[width=0.8\linewidth]{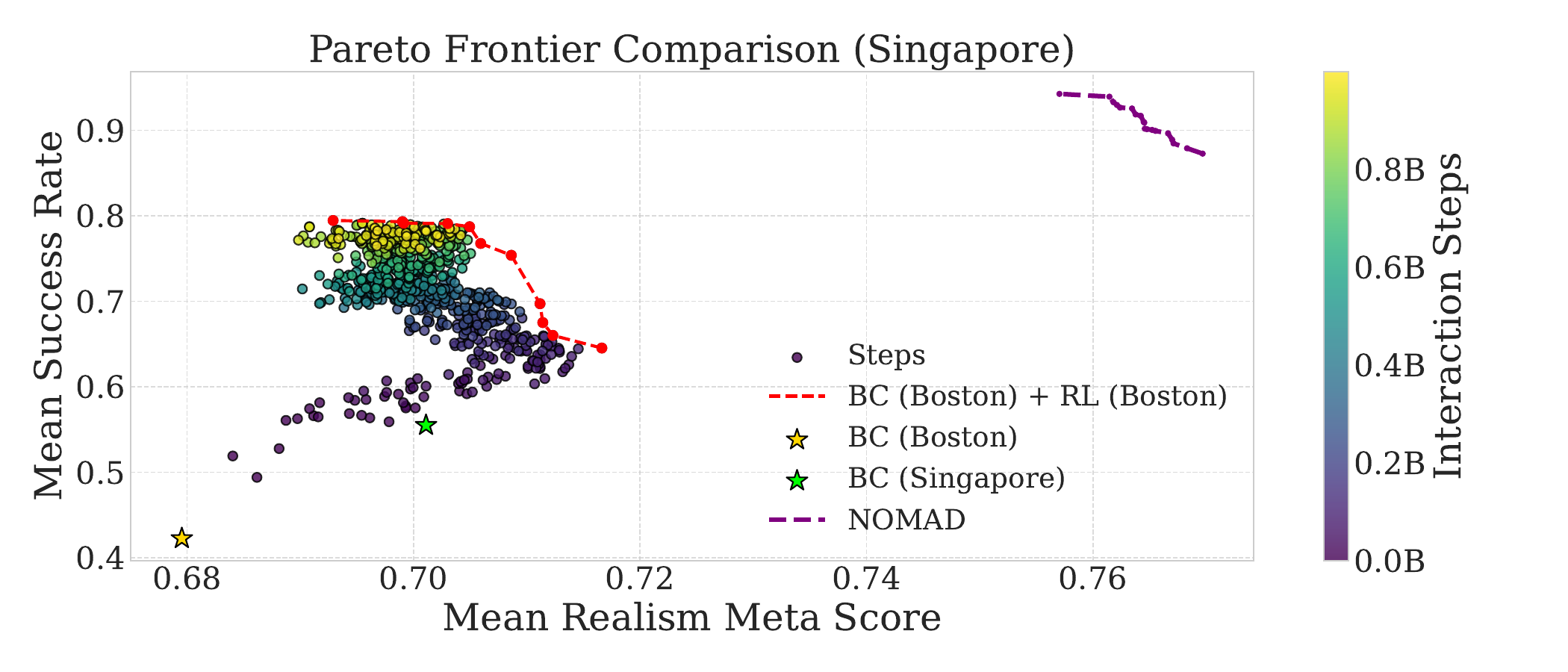}
  
  \caption{
    \textbf{Source-city self-play is beaten by NOMAD.}
    The dashed red curve (BC (Boston) + RL (Boston)) denotes policies trained with behavior cloning and self-play in Boston only, evaluated zero-shot in Singapore.
    The dashed purple curve denotes NOMAD.
     Source-city self-play improves over zero-shot transfer but fails to approach the Pareto frontier achieved by NOMAD, demonstrating that self-play must be grounded in the target-city map to effectively adapt across cities. 
  }
\label{fig:bos_to_bos_frontier}
\end{figure*}

As shown in \cref{fig:bos_to_bos_frontier}, the red Pareto frontier (BC (Boston) + RL (Boston)) surpasses the zero-shot transfer policy but remains substantially below that of NOMAD (BC (Boston) + RL (Singapore)). Self-play (RL) in Boston saturates at considerably lower success rates and yields only marginal realism improvements when evaluated in Singapore.
These results underscore the necessity of map-based self-play in the target city. In fact, effective adaptation requires grounding interaction dynamics in the geometry, topology, and traffic structure of the target city, rather than relying on pure optimization in the source city.

\subsection{NOMAD vs. Demonstration-Based Training}
We also investigate how closely NOMAD can match a policy trained with access to target-city demonstrations. Specifically, we compare NOMAD, which uses neither target-city demonstrations nor logged scenarios, against a data-driven policy trained on 3,200 scenarios with human driving trajectories from Singapore. This policy is pretrained via behavior cloning and subsequently trained using self-play MARL on \textit{both generated and logged scenarios} in the target city. The reward function, training budget, and overall protocol are identical to NOMAD; the only difference is that all training occurs while accessing target-city demonstrations.

\begin{figure*}[htpb]
    \centering\includegraphics[width=0.8\linewidth]{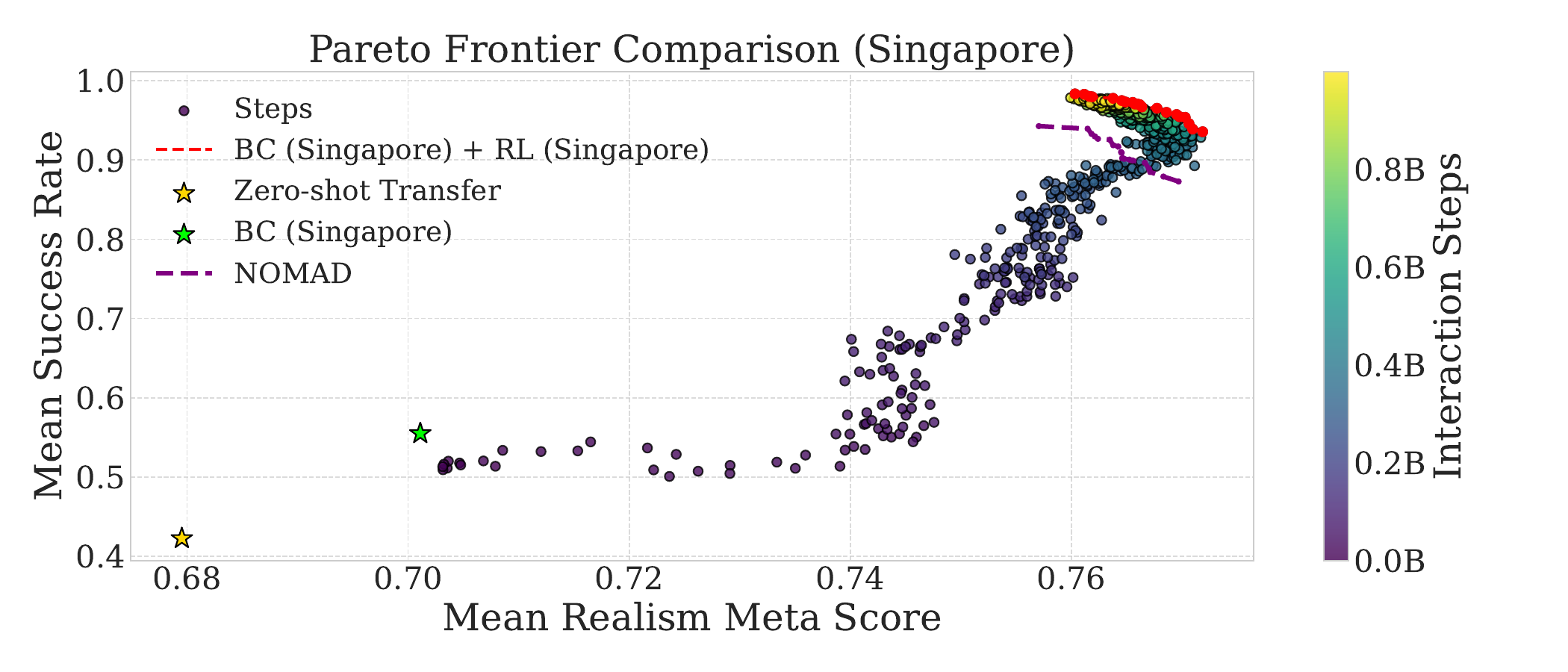}
  \caption{\textbf{NOMAD closes most of the gap between zero-shot transfer and target-city demonstration-accessed policy.}
  We compare zero-shot transfer, NOMAD (RL with generated scenarios), and a policy trained with 3,200 scenarios with logged trajectories in the target-city using the same BC+RL training protocol (BC (Singapore) + RL (Singapore)). 
  While the policy with access to target-city demonstrations yields additional gains over NOMAD, this improvement is modest compared to the substantial gap between zero-shot transfer and NOMAD, highlighting the effectiveness of map-based self-play for city transfer.
  }
  \label{fig:sin_to_sin_frontier}
\end{figure*}

This policy, labeled ``BC (Singapore) + RL (Singapore)'' in \cref{fig:sin_to_sin_frontier}, achieves a modest improvement over NOMAD along the success–realism Pareto frontier. Notably, the performance gap between this policy and NOMAD is considerably smaller than the gap between NOMAD and zero-shot transfer. This suggests that map-based self-play captures most of the benefit typically achievable by target-city demonstrations, and that the primary source of cross-city performance degradation is effectively mitigated through interaction-aware self-play in the target-city simulator rather than direct supervision from target-city data.

\subsection{NOMAD Training Curves}
\begin{figure}[htbp]
  \centering
\includegraphics[width=0.6\linewidth]{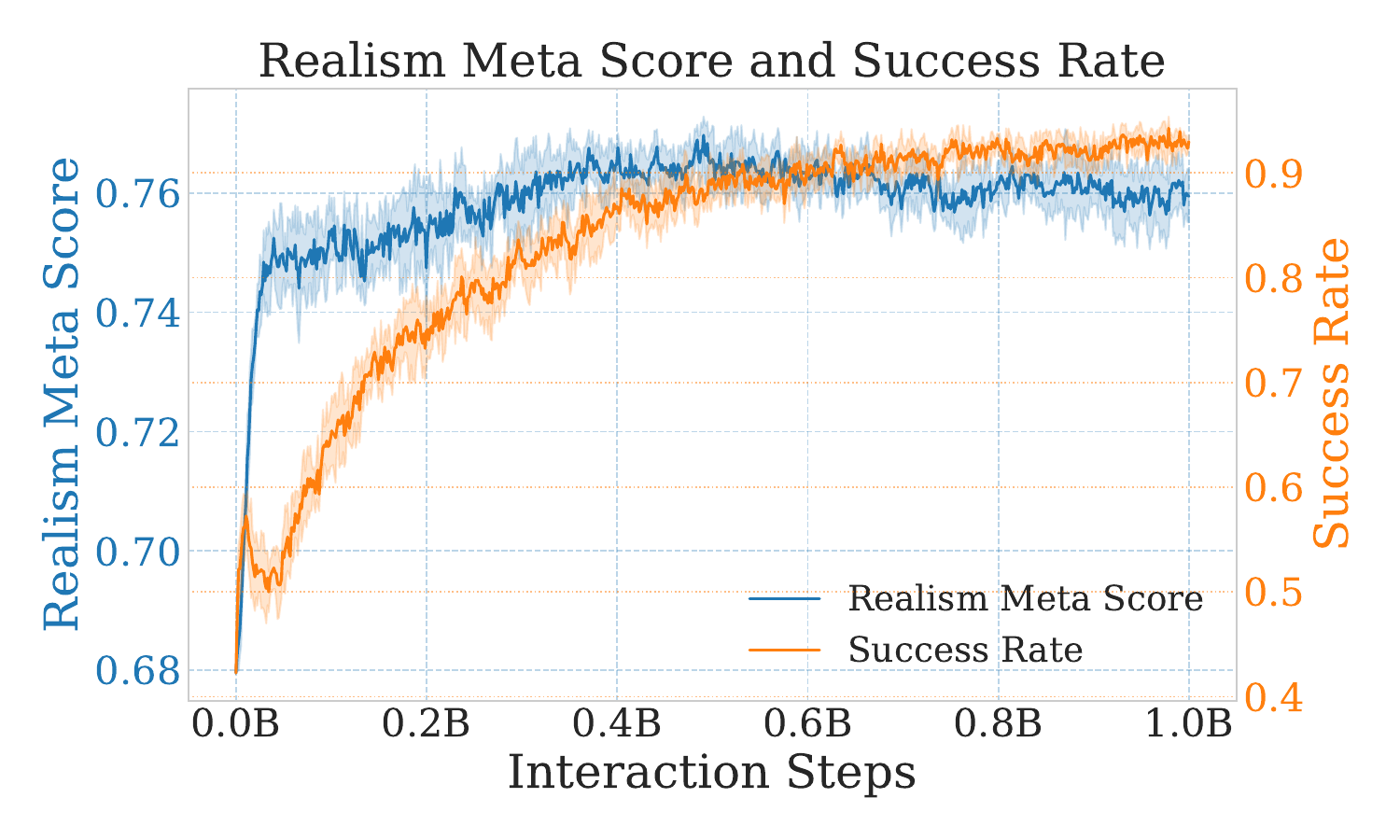}
  \caption{\textbf{Realism meta score and success rate over interaction steps on the Singapore test scenarios.}
  Results are shown with mean and 95\% confidence interval over 5 seeds.
  }
  \label{fig:realismvssuccess}
\end{figure}

\cref{fig:realismvssuccess} shows the training dynamics of NOMAD during adaptation from Boston to Singapore. Both the realism meta score and success rate are evaluated on the Singapore test set over the course of training. The success rate increases rapidly in the early stages of training, rising from approximately 42\% (the zero-shot baseline) to around 90\% within the first 600 million interaction steps, and continues to improve gradually thereafter. The realism meta score exhibits a different trajectory: it increases sharply during the initial phase of training as the policy learns to comply with target-city map constraints, then plateaus and shows a mild decline in later stages as the policy prioritizes goal completion. This pattern reflects the success–realism trade-off discussed in the main text. Results are averaged over 5 independent runs, with shaded regions indicating 95\% confidence intervals.

\subsection{The Role of Behavioral Priors}

\begin{figure}[htbp]
  \centering
\includegraphics[width=0.6\linewidth]{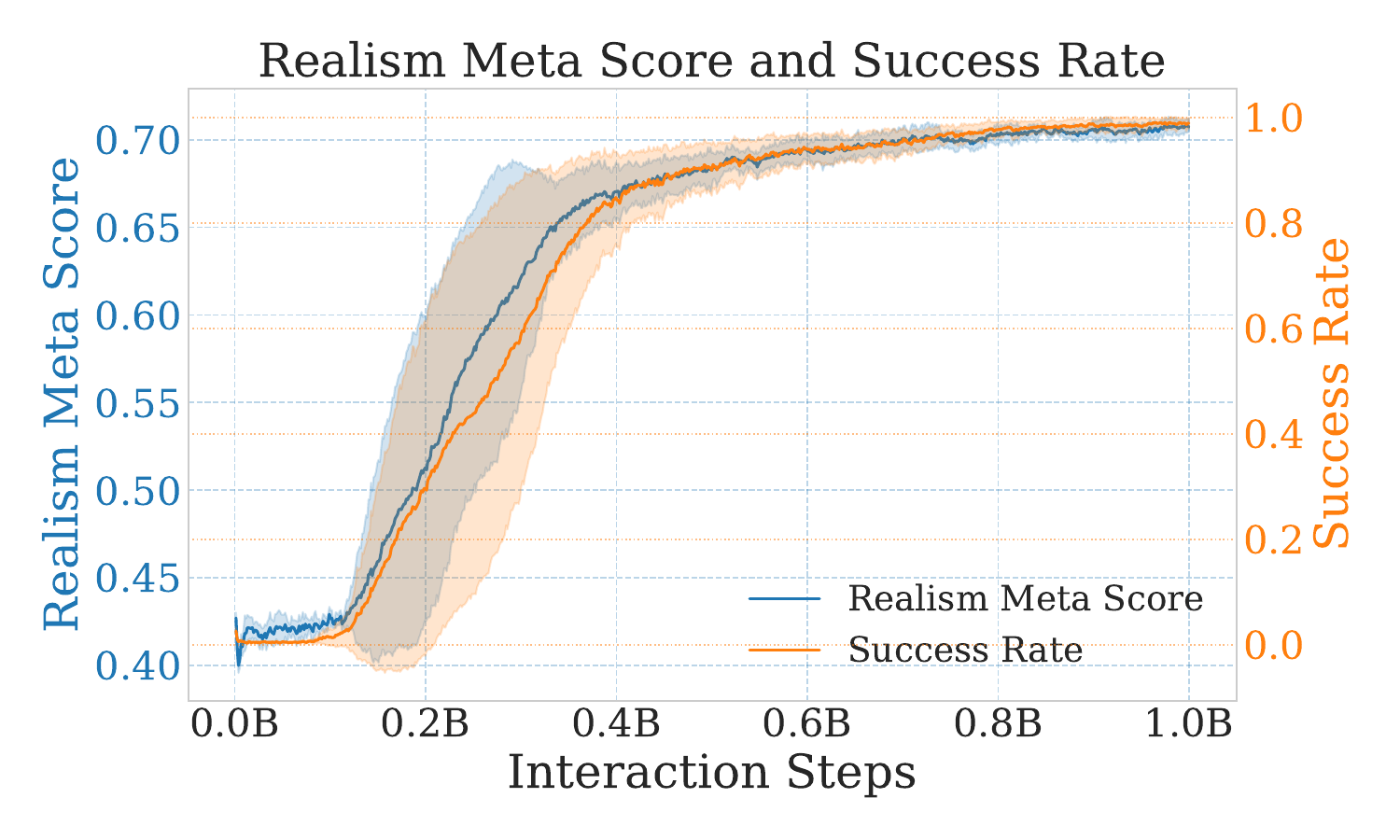}
  \caption{\textbf{Realism meta score and success rate during map-based self-play without behavioral priors in Singapore.}
  Self-play without pretraining and regularization achieves near-perfect success but a lower realism meta score.
  Shaded regions indicate 95\% confidence interval across 5 runs.
  }
  \vspace{-0.5\baselineskip}
  \label{fig:pureselfplay}
\end{figure}

\cref{fig:pureselfplay} shows the training dynamics of mere map-based self-play in Singapore without behavioral priors, corresponding to the ablation study in \cref{sec:pureselfplay}. In this setting, the policy is trained from scratch using only the reward signal, without behavior cloning initialization or KL regularization against a pretrained policy. The success rate starts near zero and increases steadily, eventually reaching near-perfect levels (99\%) after approximately 600 million interaction steps. However, the realism meta score follows a notably different trajectory compared to NOMAD (\cref{fig:realismvssuccess}): it rises from a very low initial value as the policy learns basic driving behavior, but converges to a substantially lower plateau (around 0.65–0.70) compared to NOMAD's final realism scores (around 0.76). This gap highlights the critical role of behavioral priors in maintaining human-like driving patterns when optimizing with a minimal reward function. Results are averaged over 5 independent runs, with shaded regions indicating 95\% confidence intervals.

\subsection{Sensitivity Study on KL Regularization Strength} \label{sec:klstudy}

\begin{figure}[htbp]
  \centering
  \includegraphics[width=0.6\linewidth]{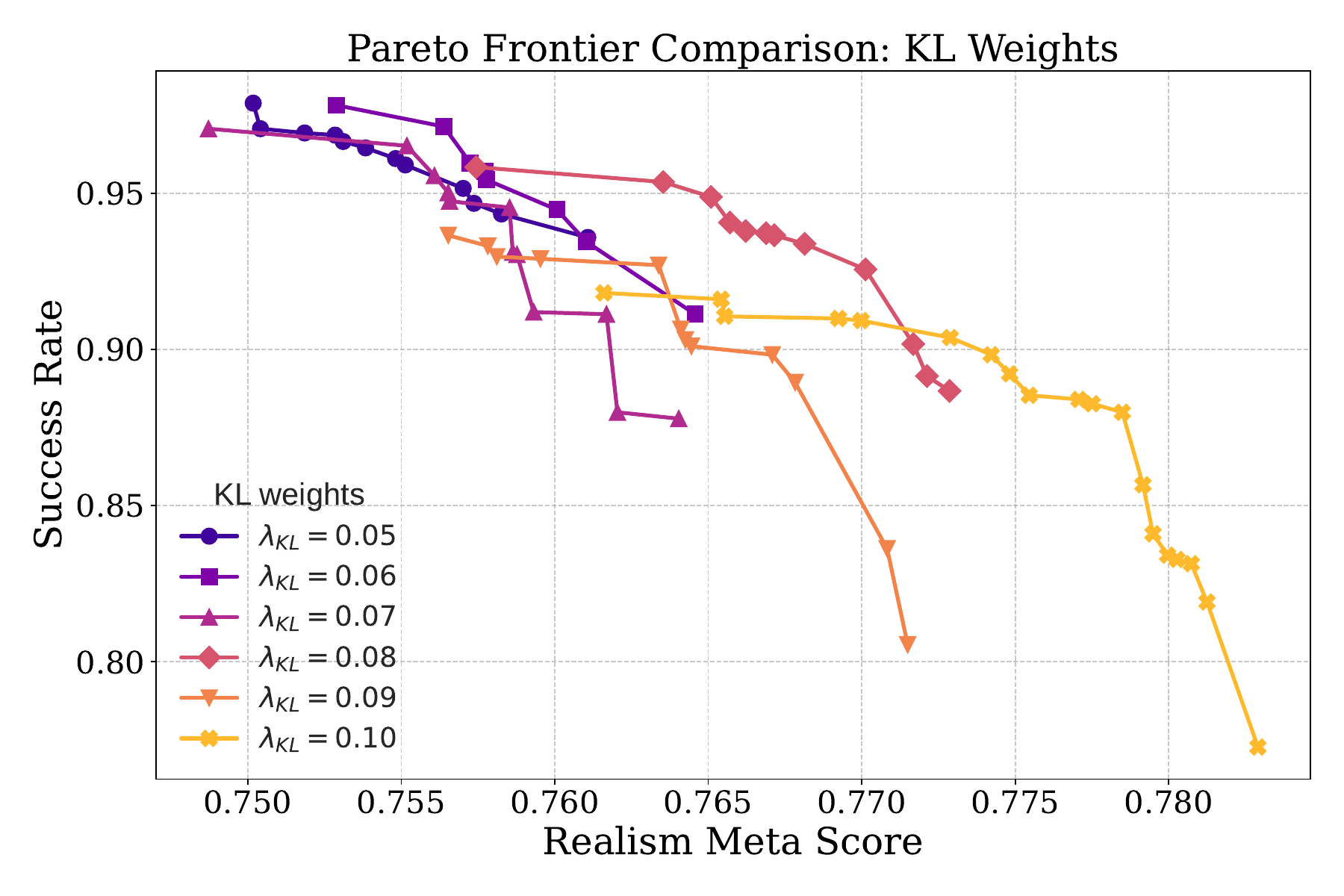}
  \caption{\textbf{The KL regularization weight controls the success--realism trade-off during adaptation.}
  We compare Pareto frontiers with different $\lambda_{\mathrm{KL}}$.
  Smaller KL weights favor higher success at the cost of realism, while larger KL weights encourage more realistic behaviors but constrain success rate.
  }
  \label{fig:kl_ablation}
\end{figure}

We evaluate how KL divergence coefficient $\lambda_{\text{KL}}$ in \cref{eq:kl_reg_obj} influences the balance between success rate and behavior realism during city transfer. 
\cref{fig:kl_ablation} presents Pareto frontiers for different values of $\lambda_{\text{KL}}$ in Boston-to-Singapore transfer.
Higher KL coefficients constrain the policy to remain closer to the source-city prior, resulting in higher realism meta score but limiting the policy's ability to adapt to target-city geometries, manifested as lower success rates. 
On the other hand, lower KL coefficients produce higher success rate but at the cost of realism.
The results show that $\lambda_{\mathrm{KL}}=0.08$ achieves the best balance.
Importantly, even with relatively aggressive or conservative KL regularization choices, NOMAD can still substantially outperform the zero-shot baseline.
In our experiments, we use $\lambda_{\text{KL}}=0.08$ for all city pairs.

%% file: appendices/EvalNonSelfplay.tex
\section{Evaluation Under Non-Self-Play Test-Time Agents}\label{sec:nonselfplay}

Autonomous driving policies are ultimately deployed in traffic with potentially \emph{unseen} agents, raising the concern that multi-agent self-play might overfit to partner behaviors encountered during training (e.g., forming seed-specific conventions)~\citep{hu2020other}.
To address this, we supplement our standard self-play evaluation with two protocol variants that explicitly break the self-play assumption.
First, \emph{log-replay} evaluates the learned ego policy against non-reactive background vehicles that follow logged human trajectories, removing partner responsiveness and directly testing robustness under distribution shift.
Second, \emph{cross-play} evaluates coordination between independently trained policies (different random seeds) at the same Pareto-frontier checkpoints, testing whether performance depends on shared conventions within a single training run.
Together, these experiments provide evidence that \textbf{NOMAD’s gains persist beyond self-play evaluation and are not driven by brittle co-adaptation to a particular set of training partners.}

\subsection{Log-Replay: Removing Co-Adaptation at Test Time}

\begin{figure*}[htpb]
  \centering
  \includegraphics[width=1.0\linewidth]{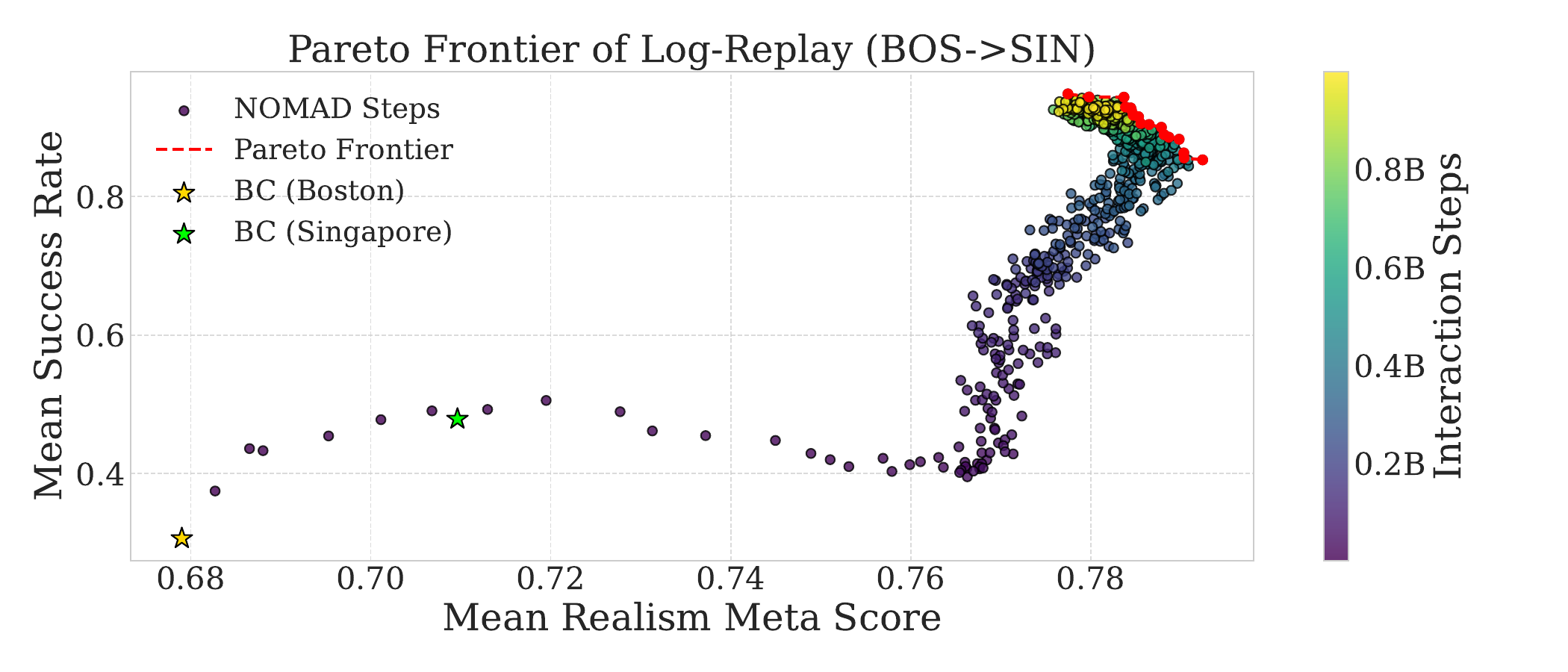}
  \caption{\textbf{Pareto frontier of success rate versus realism meta score across training checkpoints for Singapore with log-replay sim agents.}
  Each point corresponds to a policy checkpoint, colored by the number of interaction steps.
  The yellow star denotes the zero-shot transfer policy $\pi^0$ from Boston, while the lime star shows behavior cloning using Singapore trajectories.
  The red dashed line and points indicate the empirical Pareto frontier on this test set.
  Frontier checkpoints consistently dominate the baselines and exhibit only a mild trade-off between task success and realism.
  }
  \label{fig:bos_to_sin_logreplay}
\end{figure*}

\begin{figure*}[htpb]
  \centering
  \includegraphics[width=1.0\linewidth]{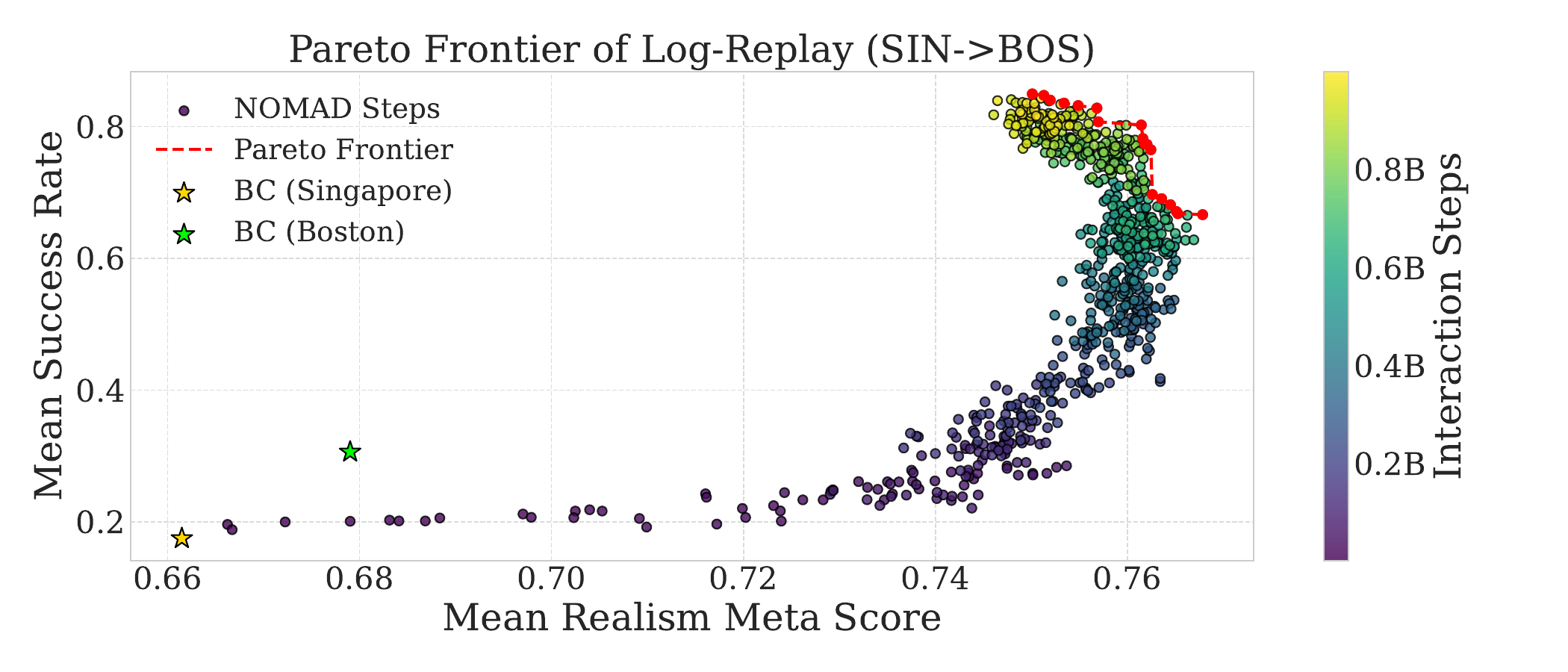}
  \caption{\textbf{Pareto frontier of success rate versus realism meta score across training checkpoints for Boston with log-replay sim agents.}
  Each point corresponds to a policy checkpoint, colored by the number of interaction steps.
  The yellow star denotes the zero-shot transfer policy $\pi^0$ from Singapore, while the lime star shows behavior cloning using Boston trajectories.
  The red dashed line and points indicate the empirical Pareto frontier on this test set.
  Frontier checkpoints consistently dominate the baselines and exhibit only a mild trade-off between task success and realism.
  }
  \label{fig:sin_to_bos_logreplay}
\end{figure*}

\begin{figure*}[htpb]
  \centering
  \includegraphics[width=1.0\linewidth]{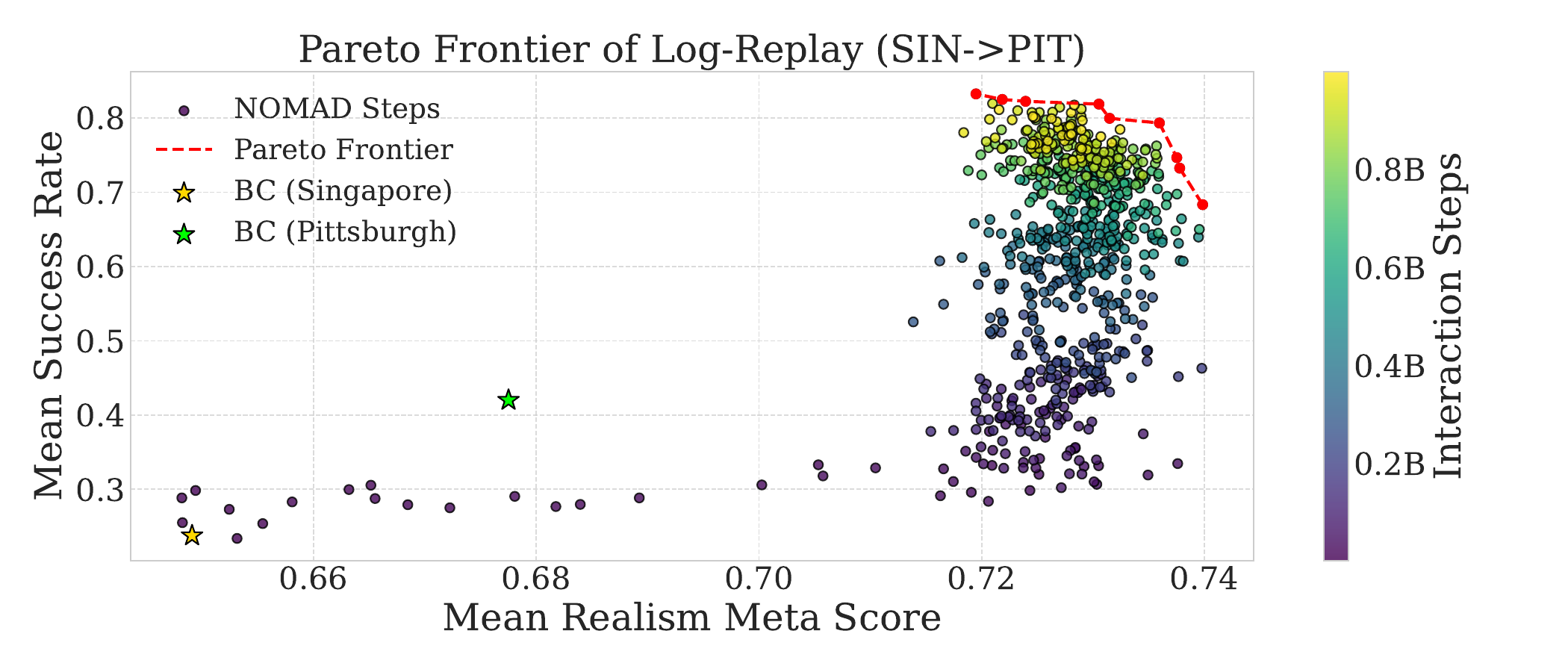}
  \caption{\textbf{Pareto frontier of success rate versus realism meta score across training checkpoints for Pittsburgh with log-replay sim agents.}
  Each point corresponds to a policy checkpoint, colored by the number of interaction steps.
  The yellow star denotes the zero-shot transfer policy $\pi^0$ from Singapore, while the lime star shows behavior cloning using Pittsburgh trajectories.
  The red dashed line and points indicate the empirical Pareto frontier on this test set.
  Frontier checkpoints consistently dominate the baselines and exhibit only a mild trade-off between task success and realism.
  }
  \label{fig:sin_to_pitts_logreplay}
\end{figure*}

To probe whether NOMAD’s self-play improvements rely on co-adaptation among learned agents, we perform a log-replay evaluation in which only the ego vehicle is controlled by the learned policy, while all other vehicles follow their logged trajectories.
Figures~\ref{fig:bos_to_sin_logreplay}, \ref{fig:sin_to_bos_logreplay}, and~\ref{fig:sin_to_pitts_logreplay} show the results on Boston-to-Singapore, Singapore-to-Boston, and Singapore-to-Pittsburgh, respectively.
Relative to self-play evaluation (\cref{fig:frontier}), we observe that the success rate decreases while realism meta score increases.

We attribute the lower success rate compared to self-play evaluation to the factor of convention losing: during training, the ego vehicle interacts with responsive self-controlled agents, whereas at test time log-replay agents do not follow the same interaction conventions; behaviors that succeed under self-play coordination (e.g., merges/yields that assume reciprocity) may fail when partners do not respond. In contrast, the higher realism may be induced by human-like interaction context: when surrounding vehicles follow human trajectories, the ego’s best-response actions tend to imitate typical human reactions (e.g., more conservative gap acceptance, yielding, and speed profiles), which are rewarded by our realism metric.

Importantly, these result differences do not affect our main claim. 
Although log-replay evaluation yields slightly different absolute success/realism values, \textbf{NOMAD consistently improves over the zero-shot policy by a large margin across all city pairs, indicating robust benefits that persist even when the ego is evaluated against non-reactive logged traffic rather than self-play partners.}

\subsection{Cross-Play: Convention Consistency Across Independently Trained Agents}

A common concern with multi-agent self-play is that independently trained agents may converge to seed-specific conventions that work well only when paired with themselves, but fail under \emph{cross-play} with a novel partner~\citep{hu2020other}.
To test this, for each city-transfer setting we select $\mathrm{n}=5$ independent instances (different random seeds) for each checkpoint on the empirical success--realism Pareto frontier.
We then evaluate all $\mathrm{n}^2$ pairings, where diagonal entries correspond to \emph{self-play} (same seed paired with itself) and off-diagonal entries correspond to \emph{cross-play} (different seeds paired together).

\begin{figure}[htpb]
  \centering
  \includegraphics[width=1.0\linewidth]{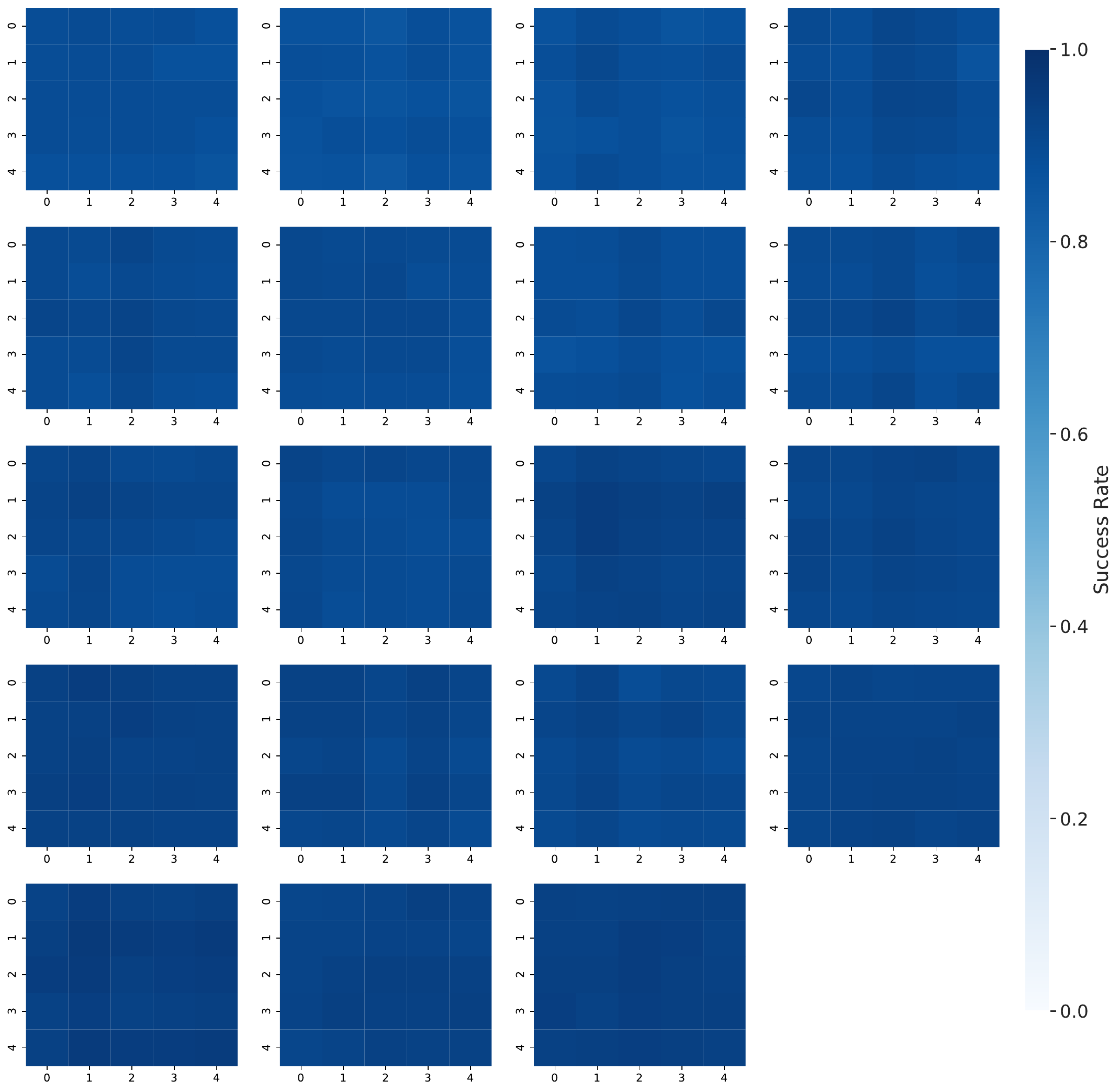}
  \caption{\textbf{Cross-play success matrices across Pareto-frontier checkpoints (Boston$\rightarrow$Singapore).}
  Each panel corresponds to one selected checkpoint on the success--realism Pareto frontier.
  Within each panel, entry $(i,j)$ reports the ego success rate when pairing seed $i$ with seed $j$ ($\mathrm{n}=5$ total seeds).
  Diagonal entries are self-play (same seed paired with itself), while off-diagonals are cross-play (different seeds).
  Off-diagonal performance is comparable to diagonal performance across checkpoints, indicating minimal sensitivity to partner seed.}
  \label{fig:xp_bos_to_sin_sr}
\end{figure}

\begin{figure}[htpb]
  \centering
  \includegraphics[width=1.0\linewidth]{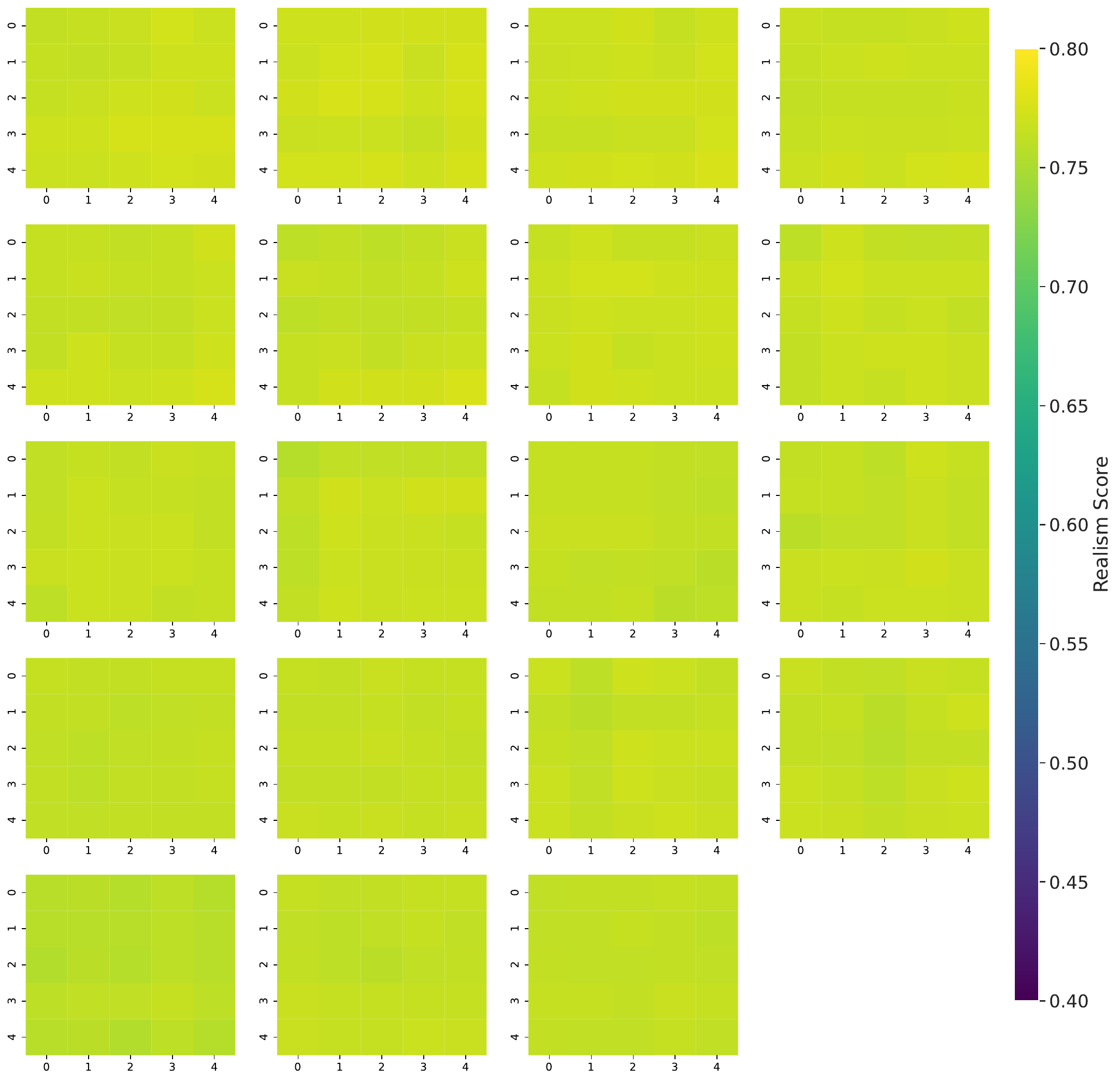}
  \caption{
\textbf{
Cross-play realism matrices across Pareto-frontier checkpoints (Boston$\rightarrow$Singapore).}
  Each panel corresponds to one selected checkpoint on the success--realism Pareto frontier.
  Within each panel, entry $(i,j)$ reports the ego realism meta score when pairing seed $i$ with seed $j$ ($\mathrm{n}=5$ total seeds).
  Diagonal entries are self-play (same seed paired with itself), while off-diagonals are cross-play (different seeds).
  Off-diagonal performance is comparable to diagonal performance across checkpoints, indicating minimal sensitivity to partner seed.
  }
  \label{fig:xp_bos_to_sin_r}
\end{figure}

\begin{figure}[htpb]
  \centering
  \includegraphics[width=1.0\linewidth]{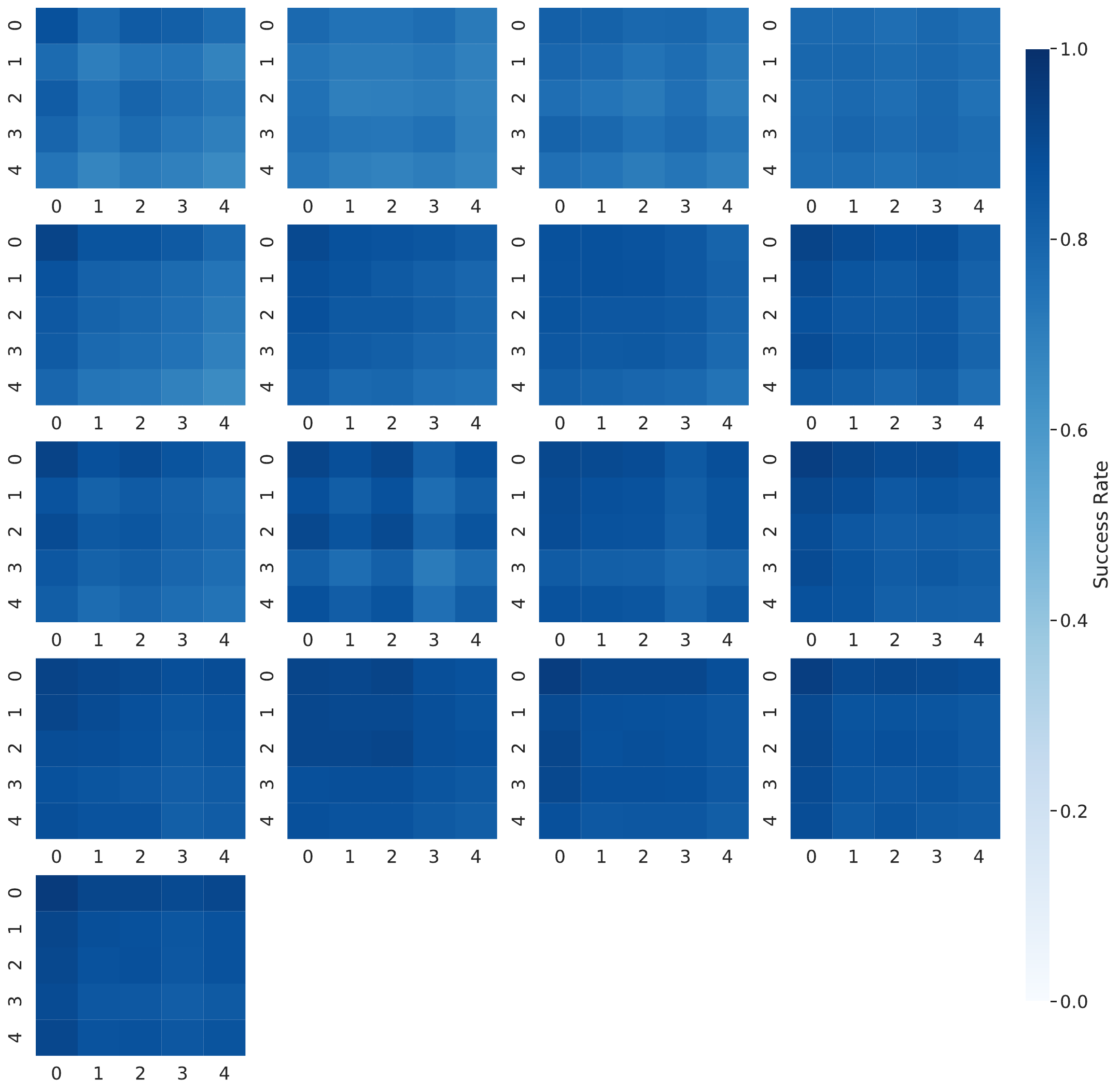}
  \caption{
\textbf{Cross-play success matrices across Pareto-frontier checkpoints (Singapore$\rightarrow$Boston).}
  Each panel corresponds to one selected checkpoint on the success--realism Pareto frontier.
  Within each panel, entry $(i,j)$ reports the ego success rate when pairing seed $i$ with seed $j$ ($\mathrm{n}=5$ total seeds).
  Diagonal entries are self-play (same seed paired with itself), while off-diagonals are cross-play (different seeds).
  Off-diagonal performance is comparable to diagonal performance across checkpoints, indicating minimal sensitivity to partner seed.
  }
  \label{fig:xp_sin_to_bos_sr}
\end{figure}

\begin{figure}[htpb]
  \centering
  \includegraphics[width=1.0\linewidth]{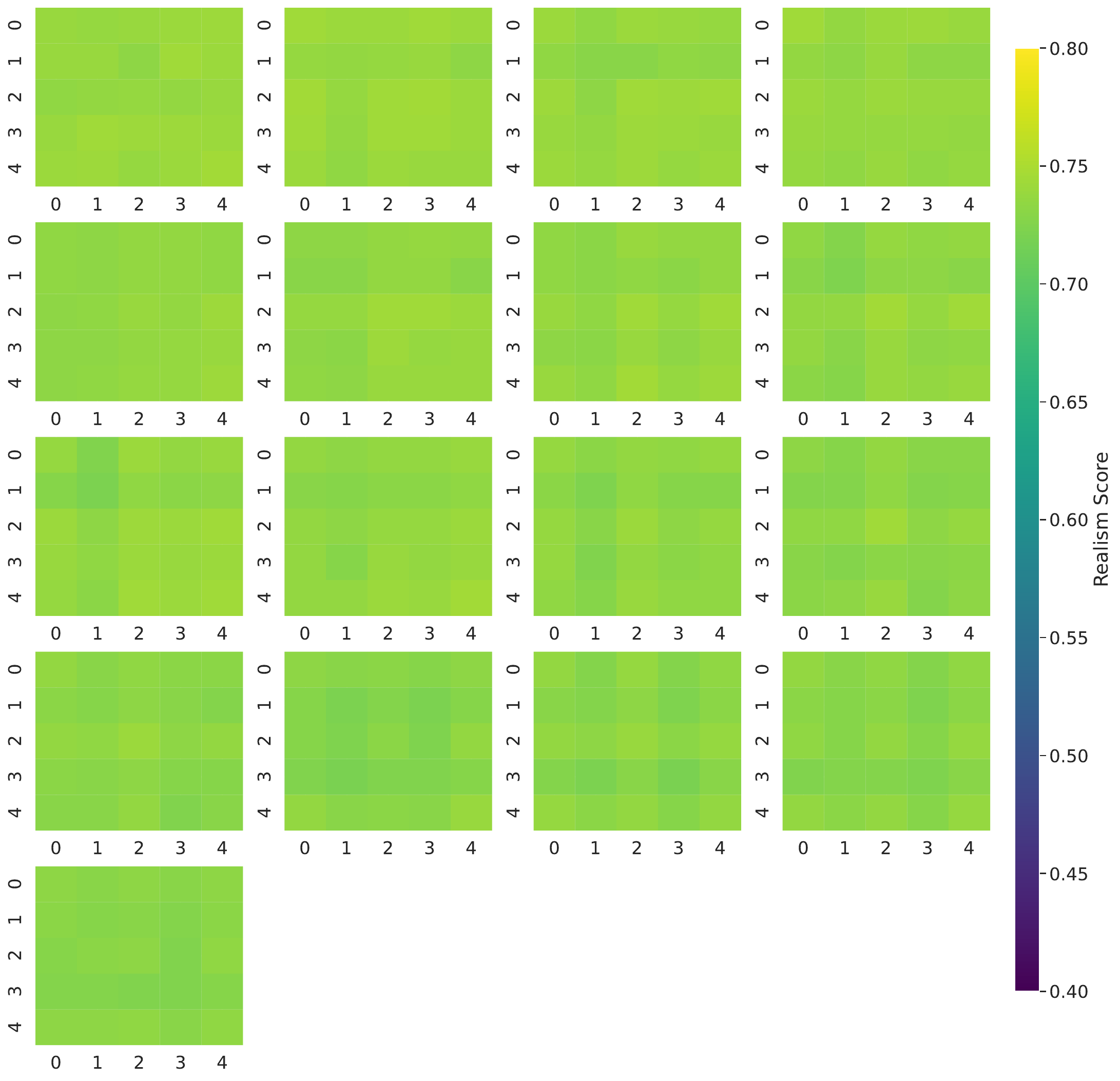}
  \caption{
    \textbf{Cross-play realism matrices across Pareto-frontier checkpoints (Singapore$\rightarrow$Boston).}
  Each panel corresponds to one selected checkpoint on the success--realism Pareto frontier.
  Within each panel, entry $(i,j)$ reports the ego realism meta score when pairing seed $i$ with seed $j$ ($\mathrm{n}=5$ total seeds).
  Diagonal entries are self-play (same seed paired with itself), while off-diagonals are cross-play (different seeds).
  Off-diagonal performance is comparable to diagonal performance across checkpoints, indicating minimal sensitivity to partner seed.
  }
  \label{fig:xp_sin_to_bos_r}
\end{figure}

\begin{figure}[htpb]
  \centering
  \includegraphics[width=1.0\linewidth]{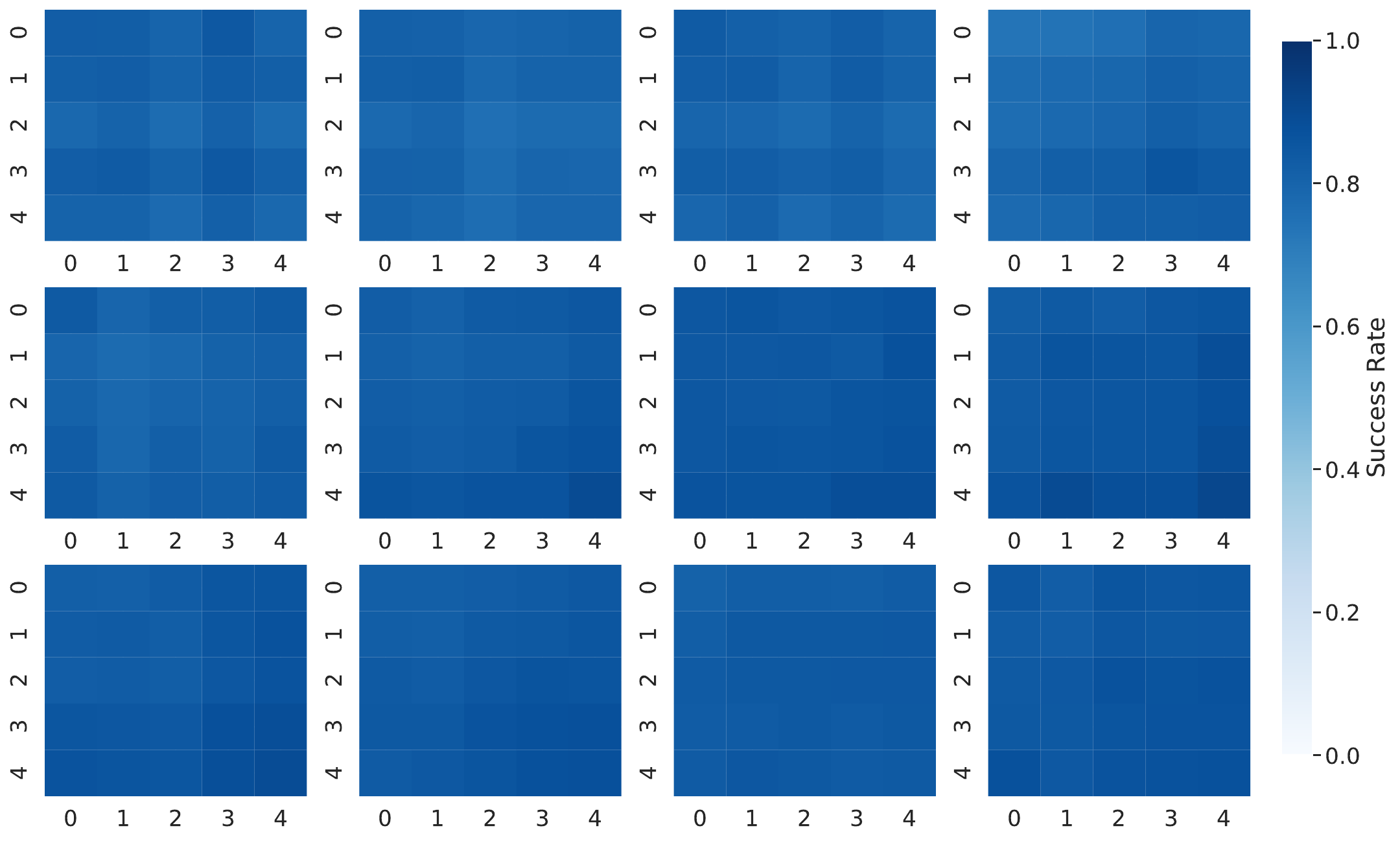}
  \caption{
    \textbf{Cross-play success matrices across Pareto-frontier checkpoints (Singapore$\rightarrow$\allowbreak Pittsburgh).}
    Each panel corresponds to one selected checkpoint on the success--realism Pareto frontier.
    Within each panel, entry $(i,j)$ reports the ego success rate when pairing seed $i$ with seed $j$ ($\mathrm{n}=5$ total seeds).
    Diagonal entries are self-play (same seed paired with itself), while off-diagonals are cross-play (different seeds).
    Off-diagonal performance is comparable to diagonal performance across checkpoints, indicating minimal sensitivity to partner seed.
  }
  \label{fig:xp_sin_to_pitts_sr}
\end{figure}

\begin{figure}[htpb]
  \centering
  \includegraphics[width=1.0\linewidth]{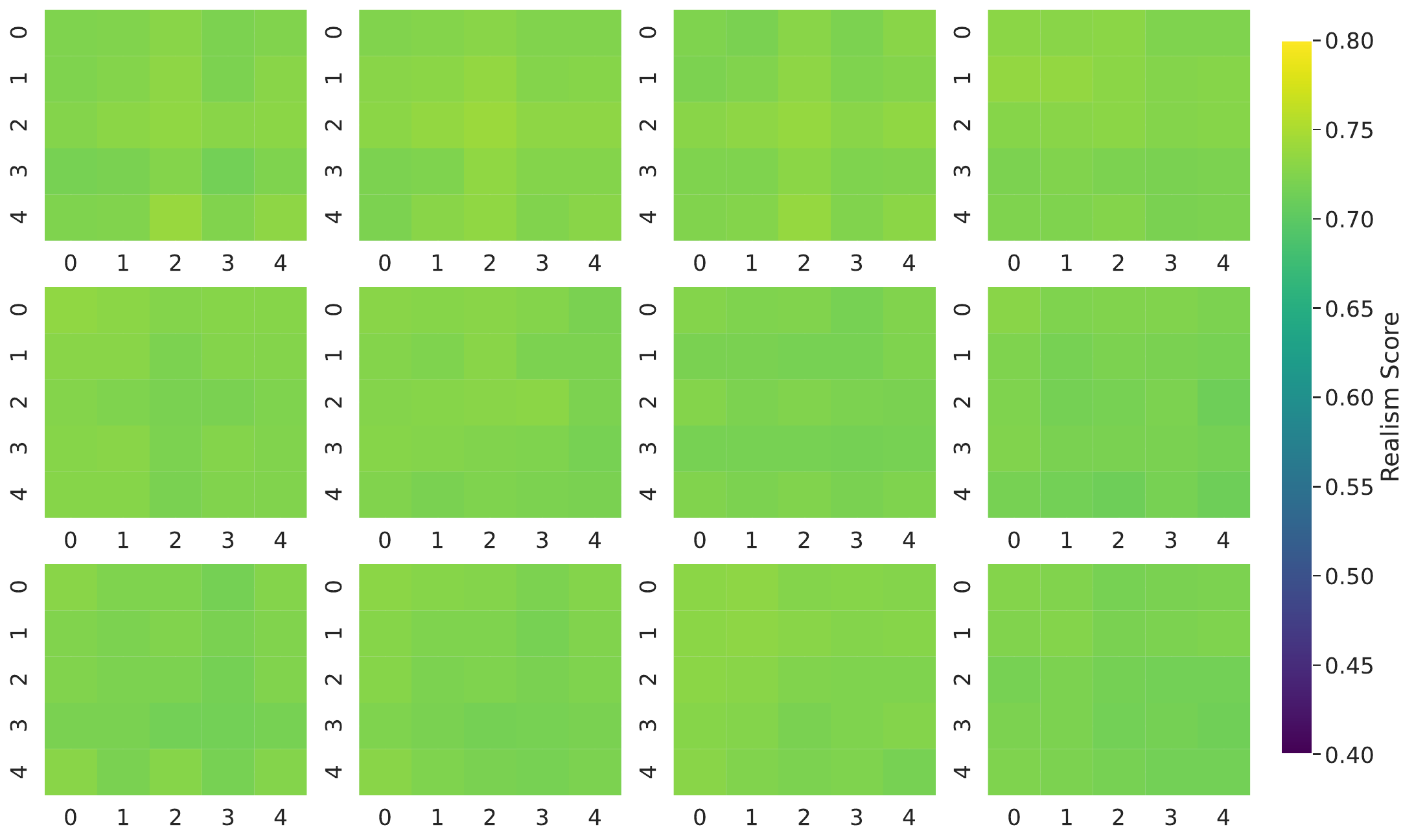}
  \caption{
    \textbf{Cross-play realism matrices across Pareto-frontier checkpoints (Singapore$\rightarrow$\allowbreak Pittsburgh).}
  Each panel corresponds to one selected checkpoint on the success--realism Pareto frontier.
  Within each panel, entry $(i,j)$ reports the ego realism meta score when pairing seed $i$ with seed $j$ ($\mathrm{n}=5$ total seeds).
  Diagonal entries are self-play (same seed paired with itself), while off-diagonals are cross-play (different seeds).
  Off-diagonal performance is comparable to diagonal performance across checkpoints, indicating minimal sensitivity to partner seed.
  }
  \label{fig:xp_sin_to_pitts_r}
\end{figure}

Across all evaluated city pairs and frontier checkpoints, cross-play performance closely matches self-play.
Figures~\ref{fig:xp_bos_to_sin_sr}--\ref{fig:xp_sin_to_pitts_r} visualize this by plotting the $5\times 5$ cross-play matrices for each frontier checkpoint; we observe no systematic degradation in off-diagonal entries.
Tables~\ref{tab:xp_bos_to_sin}--\ref{tab:xp_sin_to_pitts} further summarize the results, showing that cross-play success and realism meta score remain within the variability observed under self-play at the same checkpoint.
Overall, these results suggest NOMAD does not rely on fragile, seed-specific coordination conventions.

We attribute this stability to the strong behavioral priors and regularization used during adaptation: initializing from a pretrained behavior cloning policy and regularizing updates with KL regularization anchor learning to a shared, human-like driving mode, reducing the chance of divergent conventions across seeds.

\begin{table}[h]
\caption{
\textbf{Self-play vs cross-play summary at Pareto-frontier checkpoints (Boston$\rightarrow$Singapore).}
We evaluate each selected frontier checkpoint under both self-play (paired with the same-seed partner) and cross-play (paired with different-seed partners) using $\mathrm{n}=5$ independently trained seeds.
We report mean $\pm$ standard deviation across pairings.
Cross-play success and realism closely match self-play at the same checkpoint, indicating that NOMAD does not rely on seed-specific conventions.
}
\centering
\label{tab:xp_bos_to_sin}
\resizebox{\linewidth}{!}{%
\begin{tabular}{@{}ccccc@{}}
\toprule
\multicolumn{1}{l}{\textbf{Checkpoint}} & \multicolumn{1}{l}{\textbf{Realism Self-Play}} & \multicolumn{1}{l}{\textbf{Realism Cross-Play}} & \multicolumn{1}{l}{\textbf{Success Rate Self-Play}} & \multicolumn{1}{l}{\textbf{Success Rate Cross-Play}} \\ \midrule
3690 & 0.768 $\pm$ 0.004 & 0.769 $\pm$ 0.003 & 0.885 $\pm$ 0.011 & 0.885 $\pm$ 0.008 \\
3710 & 0.771 $\pm$ 0.003 & 0.771 $\pm$ 0.003 & 0.873 $\pm$ 0.011 & 0.872 $\pm$ 0.010 \\
3740 & 0.770 $\pm$ 0.003 & 0.769 $\pm$ 0.002 & 0.879 $\pm$ 0.017 & 0.880 $\pm$ 0.011 \\
3930 & 0.768 $\pm$ 0.003 & 0.767 $\pm$ 0.002 & 0.897 $\pm$ 0.015 & 0.894 $\pm$ 0.013 \\
3980 & 0.766 $\pm$ 0.004 & 0.766 $\pm$ 0.003 & 0.900 $\pm$ 0.014 & 0.902 $\pm$ 0.010 \\
4010 & 0.766 $\pm$ 0.005 & 0.766 $\pm$ 0.003 & 0.901 $\pm$ 0.011 & 0.899 $\pm$ 0.009 \\
4300 & 0.768 $\pm$ 0.003 & 0.768 $\pm$ 0.002 & 0.889 $\pm$ 0.012 & 0.888 $\pm$ 0.011 \\
4610 & 0.767 $\pm$ 0.004 & 0.766 $\pm$ 0.002 & 0.899 $\pm$ 0.016 & 0.898 $\pm$ 0.011 \\
4670 & 0.766 $\pm$ 0.002 & 0.765 $\pm$ 0.002 & 0.909 $\pm$ 0.016 & 0.907 $\pm$ 0.012 \\
4700 & 0.766 $\pm$ 0.005 & 0.765 $\pm$ 0.003 & 0.902 $\pm$ 0.012 & 0.903 $\pm$ 0.010 \\
4800 & 0.764 $\pm$ 0.002 & 0.763 $\pm$ 0.002 & 0.927 $\pm$ 0.013 & 0.927 $\pm$ 0.010 \\
5000 & 0.765 $\pm$ 0.003 & 0.765 $\pm$ 0.003 & 0.917 $\pm$ 0.009 & 0.917 $\pm$ 0.008 \\
6040 & 0.763 $\pm$ 0.001 & 0.763 $\pm$ 0.001 & 0.933 $\pm$ 0.005 & 0.934 $\pm$ 0.005 \\
6150 & 0.765 $\pm$ 0.001 & 0.765 $\pm$ 0.001 & 0.919 $\pm$ 0.017 & 0.921 $\pm$ 0.010 \\
6160 & 0.766 $\pm$ 0.004 & 0.766 $\pm$ 0.003 & 0.910 $\pm$ 0.013 & 0.910 $\pm$ 0.011 \\
6570 & 0.765 $\pm$ 0.004 & 0.764 $\pm$ 0.003 & 0.926 $\pm$ 0.007 & 0.924 $\pm$ 0.005 \\
7400 & 0.758 $\pm$ 0.003 & 0.758 $\pm$ 0.002 & 0.943 $\pm$ 0.011 & 0.943 $\pm$ 0.008 \\
7480 & 0.763 $\pm$ 0.003 & 0.764 $\pm$ 0.002 & 0.930 $\pm$ 0.009 & 0.930 $\pm$ 0.007 \\
7490 & 0.762 $\pm$ 0.002 & 0.763 $\pm$ 0.001 & 0.939 $\pm$ 0.004 & 0.939 $\pm$ 0.004 \\
\bottomrule
\end{tabular}%
}
\end{table}

\begin{table}[h]
\caption{
\textbf{Self-play vs cross-play summary at Pareto-frontier checkpoints (Singapore$\rightarrow$Boston).}
We evaluate each selected frontier checkpoint under both self-play (paired with the same-seed partner) and cross-play (paired with different-seed partners) using $\mathrm{n}=5$ independently trained seeds.
We report mean $\pm$ standard deviation across pairings.
Cross-play success and realism closely match self-play at the same checkpoint, indicating that NOMAD does not rely on seed-specific conventions.
}
\centering
\label{tab:xp_sin_to_bos}
\resizebox{\linewidth}{!}{%
\begin{tabular}{@{}ccccc@{}}
\toprule
\multicolumn{1}{l}{\textbf{Checkpoint}} & \multicolumn{1}{l}{\textbf{Realism Self-Play}} & \multicolumn{1}{l}{\textbf{Realism Cross-Play}} & \multicolumn{1}{l}{\textbf{Success Rate Self-Play}} & \multicolumn{1}{l}{\textbf{Success Rate Cross-Play}} \\ \midrule
4860 & 0.740 $\pm$ 0.003 & 0.738 $\pm$ 0.003 & 0.751 $\pm$ 0.077 & 0.749 $\pm$ 0.046 \\
4910 & 0.740 $\pm$ 0.003 & 0.739 $\pm$ 0.003 & 0.724 $\pm$ 0.036 & 0.721 $\pm$ 0.023 \\
4920 & 0.738 $\pm$ 0.004 & 0.737 $\pm$ 0.004 & 0.756 $\pm$ 0.042 & 0.755 $\pm$ 0.029 \\
4940 & 0.738 $\pm$ 0.004 & 0.737 $\pm$ 0.002 & 0.776 $\pm$ 0.012 & 0.771 $\pm$ 0.012 \\
6180 & 0.737 $\pm$ 0.003 & 0.735 $\pm$ 0.002 & 0.783 $\pm$ 0.089 & 0.783 $\pm$ 0.055 \\
6240 & 0.736 $\pm$ 0.005 & 0.735 $\pm$ 0.004 & 0.828 $\pm$ 0.056 & 0.825 $\pm$ 0.035 \\
6250 & 0.736 $\pm$ 0.005 & 0.736 $\pm$ 0.004 & 0.832 $\pm$ 0.050 & 0.832 $\pm$ 0.031 \\
6290 & 0.735 $\pm$ 0.007 & 0.733 $\pm$ 0.004 & 0.846 $\pm$ 0.053 & 0.845 $\pm$ 0.033 \\
6310 & 0.736 $\pm$ 0.008 & 0.736 $\pm$ 0.005 & 0.824 $\pm$ 0.063 & 0.824 $\pm$ 0.040 \\
6370 & 0.736 $\pm$ 0.005 & 0.735 $\pm$ 0.004 & 0.836 $\pm$ 0.072 & 0.836 $\pm$ 0.048 \\
6800 & 0.733 $\pm$ 0.006 & 0.733 $\pm$ 0.004 & 0.854 $\pm$ 0.043 & 0.852 $\pm$ 0.032 \\
6840 & 0.732 $\pm$ 0.006 & 0.731 $\pm$ 0.004 & 0.862 $\pm$ 0.048 & 0.860 $\pm$ 0.031 \\
6880 & 0.732 $\pm$ 0.005 & 0.731 $\pm$ 0.003 & 0.870 $\pm$ 0.039 & 0.870 $\pm$ 0.025 \\
7040 & 0.729 $\pm$ 0.006 & 0.727 $\pm$ 0.004 & 0.885 $\pm$ 0.039 & 0.882 $\pm$ 0.023 \\
7520 & 0.731 $\pm$ 0.007 & 0.730 $\pm$ 0.004 & 0.880 $\pm$ 0.040 & 0.878 $\pm$ 0.024 \\
7530 & 0.731 $\pm$ 0.006 & 0.730 $\pm$ 0.004 & 0.873 $\pm$ 0.038 & 0.871 $\pm$ 0.025 \\
7540 & 0.730 $\pm$ 0.004 & 0.729 $\pm$ 0.003 & 0.881 $\pm$ 0.042 & 0.879 $\pm$ 0.027 \\
\bottomrule
\end{tabular}%
}
\end{table}

\begin{table}[h]
\caption{
\textbf{Self-play vs cross-play summary at Pareto-frontier checkpoints (Singapore$\rightarrow$\allowbreak Pittsburgh).}
We evaluate each selected frontier checkpoint under both self-play (paired with the same-seed partner) and cross-play (paired with different-seed partners) using $\mathrm{n}=5$ independently trained seeds.
We report mean $\pm$ standard deviation across pairings.
Cross-play success and realism closely match self-play at the same checkpoint, indicating that NOMAD does not rely on seed-specific conventions.
}
\centering
\label{tab:xp_sin_to_pitts}
\resizebox{\linewidth}{!}{%
\begin{tabular}{@{}ccccc@{}}
\toprule
\multicolumn{1}{l}{\textbf{Checkpoint}} & \multicolumn{1}{l}{\textbf{Realism Self-Play}} & \multicolumn{1}{l}{\textbf{Realism Cross-Play}} & \multicolumn{1}{l}{\textbf{Success Rate Self-Play}} & \multicolumn{1}{l}{\textbf{Success Rate Cross-Play}} \\ \midrule
6030 & 0.726 $\pm$ 0.007 & 0.726 $\pm$ 0.005 & 0.809 $\pm$ 0.029 & 0.809 $\pm$ 0.018 \\
6040 & 0.730 $\pm$ 0.005 & 0.728 $\pm$ 0.005 & 0.796 $\pm$ 0.024 & 0.793 $\pm$ 0.015 \\
6080 & 0.727 $\pm$ 0.006 & 0.727 $\pm$ 0.005 & 0.806 $\pm$ 0.029 & 0.804 $\pm$ 0.016 \\
6340 & 0.728 $\pm$ 0.006 & 0.726 $\pm$ 0.004 & 0.798 $\pm$ 0.041 & 0.794 $\pm$ 0.024 \\
6550 & 0.727 $\pm$ 0.005 & 0.725 $\pm$ 0.003 & 0.809 $\pm$ 0.025 & 0.812 $\pm$ 0.017 \\
7010 & 0.725 $\pm$ 0.004 & 0.724 $\pm$ 0.004 & 0.843 $\pm$ 0.031 & 0.838 $\pm$ 0.018 \\
7020 & 0.721 $\pm$ 0.003 & 0.721 $\pm$ 0.003 & 0.855 $\pm$ 0.016 & 0.857 $\pm$ 0.011 \\
7050 & 0.719 $\pm$ 0.006 & 0.720 $\pm$ 0.004 & 0.862 $\pm$ 0.030 & 0.858 $\pm$ 0.020 \\
7070 & 0.722 $\pm$ 0.005 & 0.722 $\pm$ 0.004 & 0.849 $\pm$ 0.031 & 0.849 $\pm$ 0.018 \\
7370 & 0.723 $\pm$ 0.004 & 0.723 $\pm$ 0.004 & 0.847 $\pm$ 0.027 & 0.845 $\pm$ 0.015 \\
7470 & 0.726 $\pm$ 0.005 & 0.726 $\pm$ 0.003 & 0.832 $\pm$ 0.014 & 0.836 $\pm$ 0.009 \\
7520 & 0.720 $\pm$ 0.005 & 0.719 $\pm$ 0.003 & 0.856 $\pm$ 0.017 & 0.852 $\pm$ 0.013 \\
\bottomrule
\end{tabular}%
}
\end{table}

%% file: appendices/SameSideTransfer.tex
\section{Transfer to Cities with Same-Side Traffic}\label{sec:samesidetransfer}

The main experiments evaluate NOMAD under transfer between cities with different driving sides, where the domain shift is particularly pronounced. To assess whether the observed gains are driven only by the driving-side mismatch, we also evaluate transfer between cities with the same driving side. Specifically, we consider Boston-to-Pittsburgh transfer, where both cities follow right-hand traffic.

\begin{figure*}[htpb]
  \centering
  \includegraphics[width=1.0\linewidth]{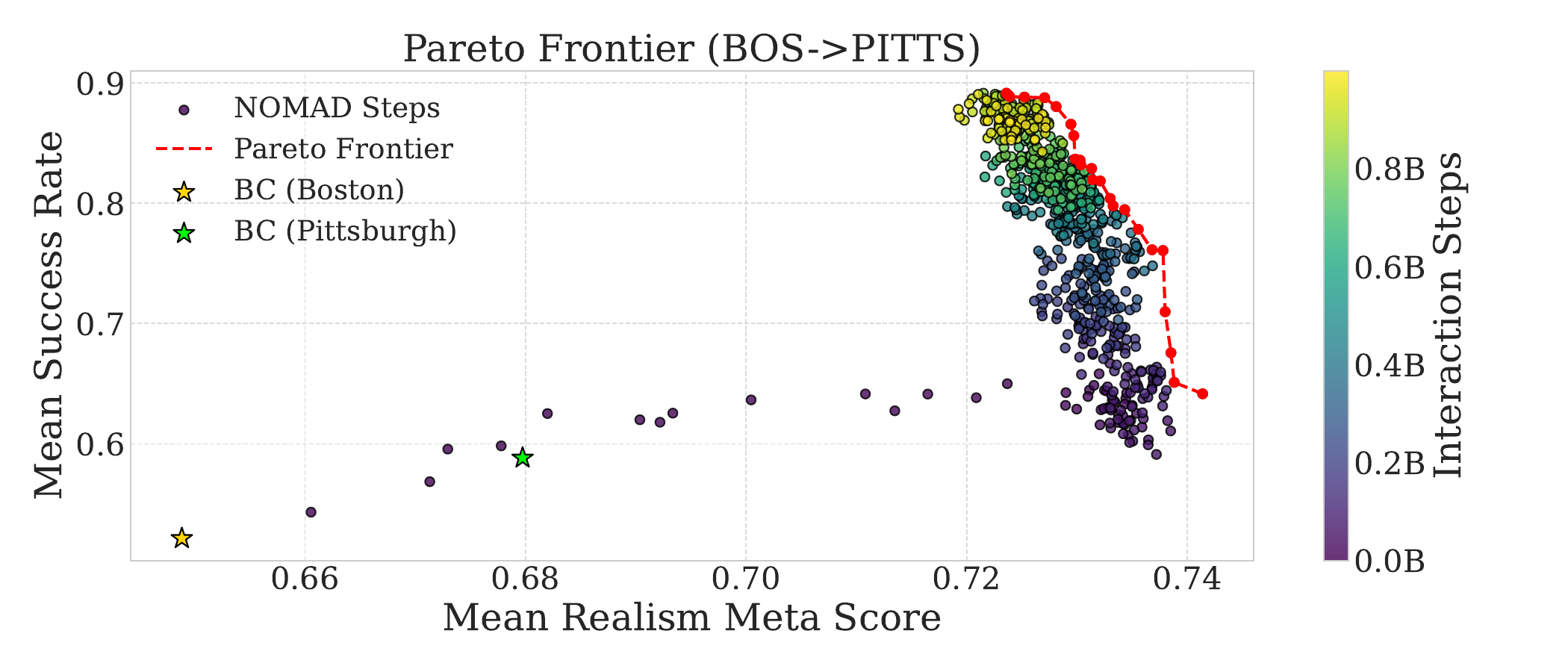}
  \caption{\textbf{Pareto frontier of success rate versus realism meta score across training checkpoints for Boston-to-Pittsburgh transfer.}
  Each point corresponds to a policy checkpoint, colored by the number of interaction steps.
  The yellow star denotes the zero-shot transfer policy $\pi^0$ from Boston, while the lime star shows behavior cloning using Pittsburgh trajectories.
  The red dashed line and points indicate the empirical Pareto frontier on this test set.
  Frontier checkpoints consistently dominate the baselines and exhibit only a mild trade-off between task success and realism.
  }
  \label{fig:bos_to_pitts}
\end{figure*}

The results shown in \cref{fig:bos_to_pitts} show that NOMAD still substantially improves over the zero-shot baseline, with gains comparable to those observed in the opposite-side transfer settings in \cref{fig:frontier}.
This suggests that NOMAD’s benefit is not limited to opposite-side transfer. Even when the driving side is shared, cities can differ substantially in road geometry, intersection topology, lane configurations, traffic-density patterns, and interaction structure. These map- and interaction-dependent differences cannot be resolved by a simple geometric transformation but can be addressed through NOMAD’s target-city self-play adaptation.

%% file: appendices/DrivingSide.tex
\section{Effect of Map Mirroring for Opposite-Side Driving}\label{sec:drivingside}

A natural question in cross-city transfer is whether a policy trained in a right-hand driving city can be adapted to a left-hand driving city by simply mirroring the target-city map.
Such a geometric transformation would, in principle, align the driving-side convention between cities such as Boston and Singapore, without requiring additional training.

To evaluate this hypothesis, we mirror the Singapore map and evaluate zero-shot transfer policies on the mirrored map.
Examples of scenarios after mirroring are provided in \cref{fig:cities}, and quantitative results are summarized in \cref{tab:mirror}.
Here, BOS and SIN denote scenarios from Boston and Singapore, respectively.
$\pi^0_{\text{BOS}}$ and $\pi^0_{\text{SIN}}$ denote BC policies trained in Boston (right-hand traffic) and Singapore (left-hand traffic), respectively.

The results show that map mirroring yields only marginal improvements in zero-shot success rate.
While mirroring partially addresses the left-hand versus right-hand driving mismatch, the resulting gains are small relative to those achieved by self-play fine-tuning.
This indicates that cross-city performance degradation cannot be attributed solely to driving-side conventions.
Importantly, this experiment isolates the effect of geometric alignment while holding the policy fixed.
The limited benefit of mirroring suggests that cross-city distribution shift arises from deeper factors, including differences in road geometry, intersection topology, lane-width distributions, and traffic density patterns.
These interaction-level and structural differences cannot be resolved through geometric transformations alone, motivating the need for adaptive interaction-aware learning as enabled by NOMAD.


\begin{table}[htbp]
\caption{\textbf{Effect of map mirroring on zero-shot cross-city transfer.}
BOS and SIN denote Boston and Singapore scenarios.
$\pi^0_{\text{BOS}}$ and $\pi^0_{\text{SIN}}$ denote behavior cloning policies trained in Boston and Singapore.
Mirroring the city map yields only modest performance improvements, indicating that cross-city distribution shift extends beyond driving-side conventions and cannot be resolved by geometric transformations alone.}
\label{tab:mirror}
\centering
\begin{tabular}{@{}cccc@{}}
\toprule
\textbf{Policy} & \textbf{Scenarios}   & \textbf{Realism Meta Score} & \textbf{Success Rate} \\ \hline
\multirow{3}{*}{$\pi^0_{\text{BOS}}$} 
& Original BOS           &   0.7022                     &  55.77\%                     \\
& Original SIN           &   0.6795                     &  42.25\%                     \\
& Mirrored SIN &   0.6756                    &  43.96\%                     \\
\hline
\multirow{3}{*}{$\pi^0_{\text{SIN}}$} 
&Original SIN           &     0.7011                    &   55.49\%                    \\
&Original BOS           &     0.6585                    &   36.67\%                    \\
& Mirrored BOS    &   0.6278                     & 38.20\%                      \\
\bottomrule
\end{tabular}
\end{table}


%% file: appendices/HoldoutEval.tex
\section{Additional Hold-Out Evaluation of Selected Frontier Checkpoints} \label{sec:holdouteval}

To assess potential selection bias, particularly overestimation when constructing the empirical Pareto frontier from a finite evaluation set, we additionally evaluate the checkpoints selected on the frontier using an extra held-out set of 800 scenarios (new test set) that is disjoint from both training and the original test set.
Ideally, both evaluation sets are i.i.d. samples from the same underlying scenario distribution and that metrics are noise-free, the re-evaluated frontier checkpoints would be expected to remain close to, and typically slightly below, the frontier estimated on the original test set.

\begin{figure*}[htpb]
  \centering
  \includegraphics[width=1.0\linewidth]{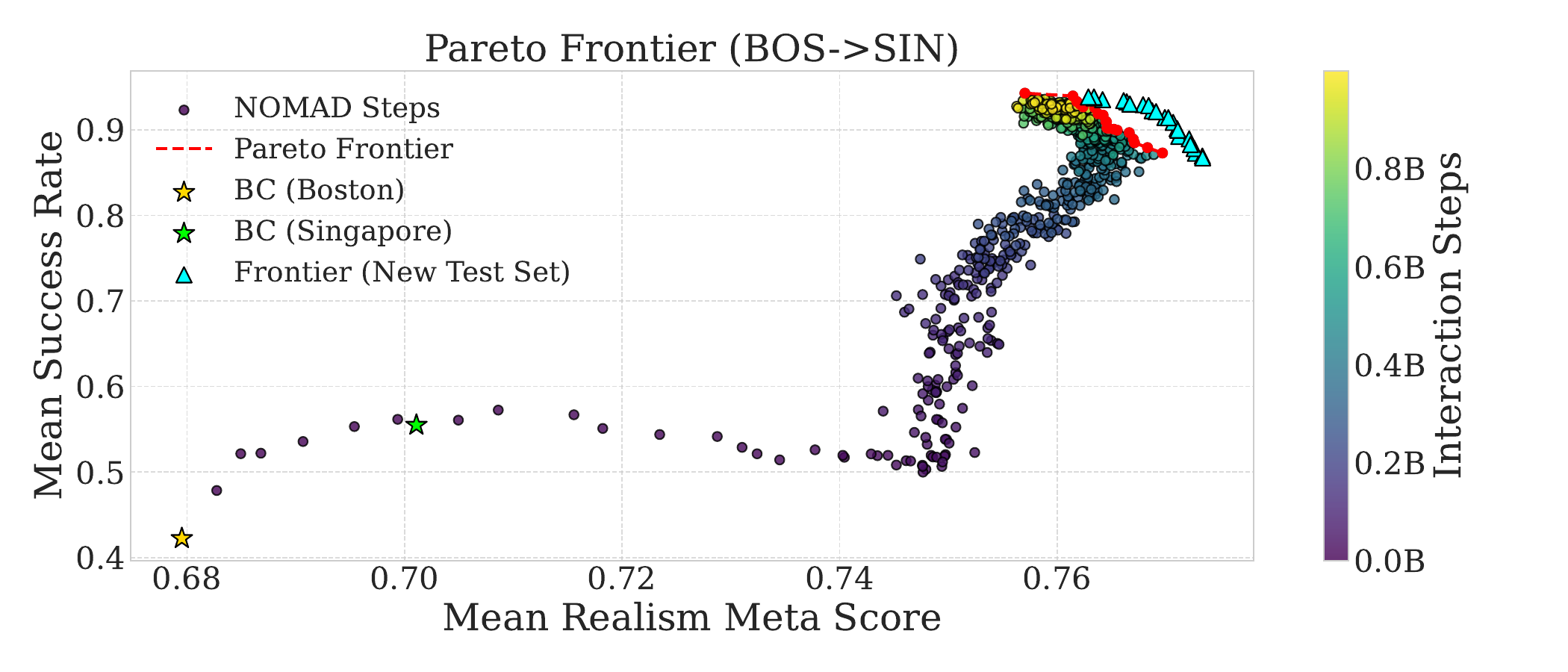}
  \caption{\textbf{Pareto frontier of success rate versus realism meta score across training checkpoints under Boston-to-Singapore transfer.}
  Each point corresponds to a policy checkpoint, colored by the number of interaction steps.
  The yellow star denotes the zero-shot transfer policy $\pi^0$ from Boston, while the lime star shows behavior cloning using Singapore trajectories.
  The red dashed line and points indicate the empirical Pareto frontier on this test set.
  Frontier checkpoints consistently dominate the baselines and exhibit only a mild trade-off between task success and realism.
  Blue triangles report the performance of selected frontier policies on a held-out Singapore test set, confirming that the Pareto dominance generalizes beyond the test data.
  }
  \label{fig:bos_to_sin_frontiernewtest}
\end{figure*}

\begin{figure*}[htpb]
  \centering
  \includegraphics[width=1.0\linewidth]{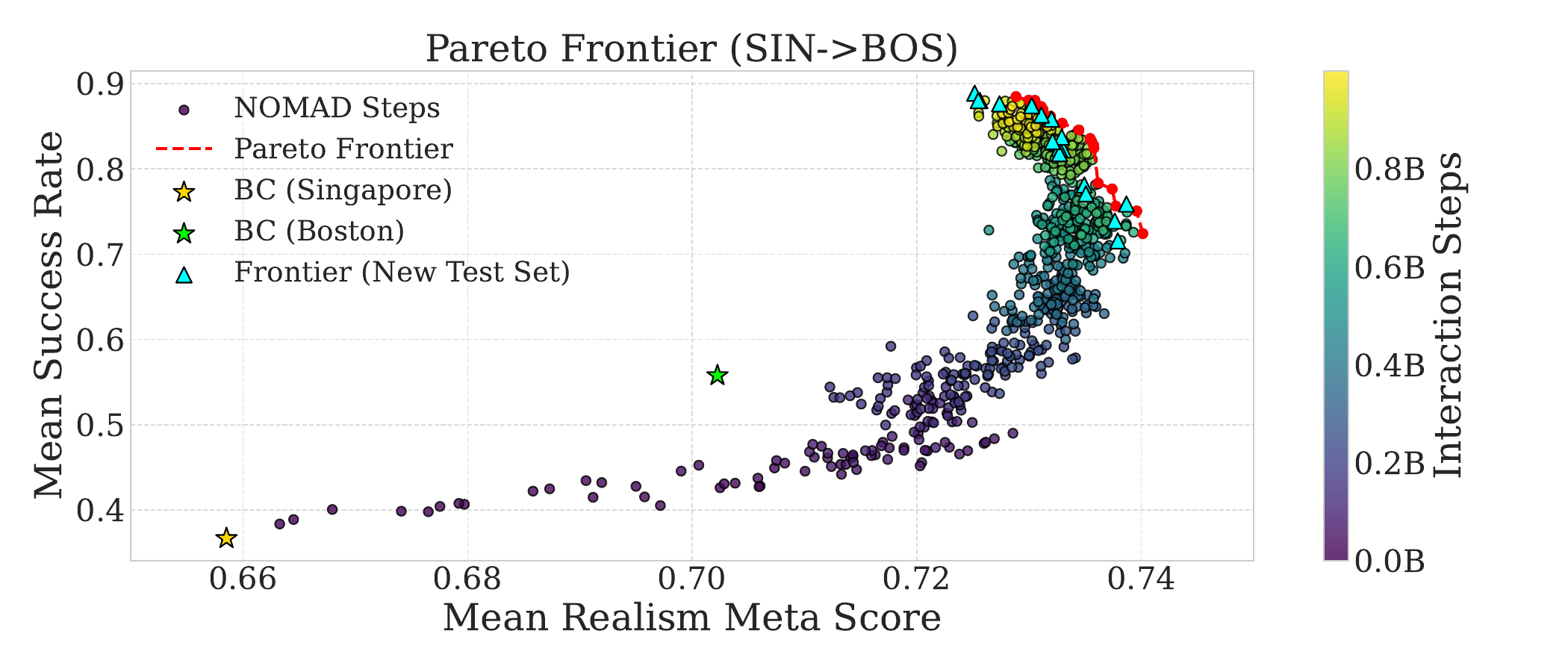}
  \caption{\textbf{Pareto frontier of success rate versus realism meta score across training checkpoints under Singapore-to-Boston transfer.}
  Each point corresponds to a policy checkpoint, colored by the number of interaction steps.
  The yellow star denotes the zero-shot transfer policy $\pi^0$ from Singapore, while the lime star shows behavior cloning using Boston trajectories.
  The red dashed line and points indicate the empirical Pareto frontier on this test set.
  Frontier checkpoints consistently dominate the baselines and exhibit only a mild trade-off between task success and realism.
  Blue triangles report the performance of selected frontier policies on a held-out Boston test set, confirming that the Pareto dominance generalizes beyond the test data.
  }
  \label{fig:sin_to_bos_frontiernewtest}
\end{figure*}

\begin{figure*}[htpb]
  \centering
  \includegraphics[width=1.0\linewidth]{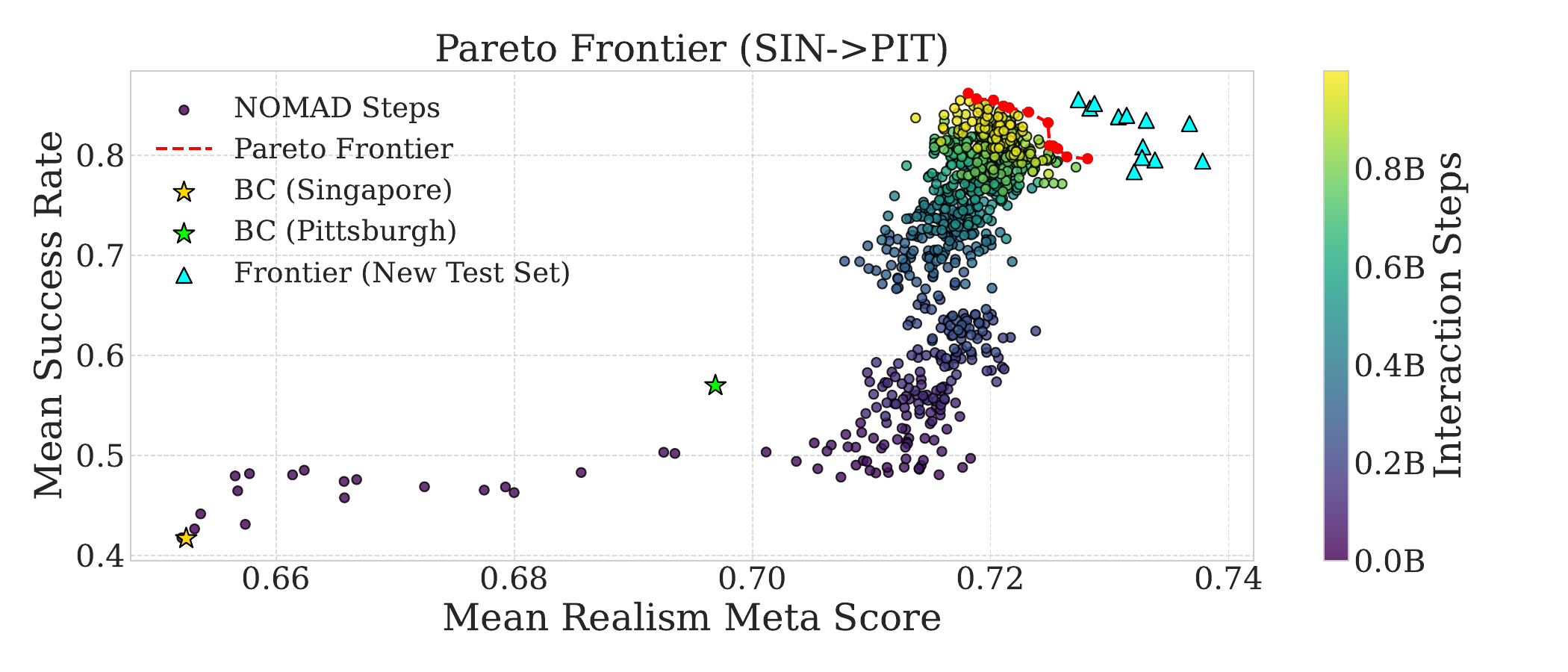}
  \caption{\textbf{Pareto frontier of success rate versus realism meta score across training checkpoints under Singapore-to-Pittsburgh transfer.}
  Each point corresponds to a policy checkpoint, colored by the number of interaction steps.
  The yellow star denotes the zero-shot transfer policy $\pi^0$ from Singapore, while the lime star shows behavior cloning using Pittsburgh trajectories.
  The red dashed line and points indicate the empirical Pareto frontier on this test set.
  Frontier checkpoints consistently dominate the baselines and exhibit only a mild trade-off between task success and realism.
  Blue triangles report the performance of selected frontier policies on a held-out Pittsburgh test set, confirming that the Pareto dominance generalizes beyond the test data.
  }
  \label{fig:sin_to_pitts_frontiernewtest}
\end{figure*}

\cref{fig:bos_to_sin_frontiernewtest}, \cref{fig:sin_to_bos_frontiernewtest}, and \cref{fig:sin_to_pitts_frontiernewtest} show the evaluation results with blue triangles in Boston--Singapore, Singapore--Boston, Singapore--Pittsburgh transfers, respectively.
For the success rate, the evaluation results on the new test set closely aligned with the results on the original test set, showing that there is little bias along our empirical Pareto frontier.
In contrast, along the realism axis, we observe that re-evaluated points can deviate slightly from our expectation and occasionally appear right to the original frontier, specifically for Boston--Singapore and Singapore--Pittsburgh transfers.

A possible reason is that WOSAC realism is computed relative to the set’s ground-truth trajectories; hence small differences in the composition of scenarios or expert driving style in the held-out set can lead to systematic shifts in the absolute realism meta score even when success rates remain similar.
To verify this argument, we checked the WOSAC realism meta score of human trajectories in different sets of scenarios from Singapore, which is the upper bound to their corresponding set, showing in \cref{tab:wosacscoreupperbound}.

\begin{table}[htbp]
\centering
\caption{The WOSAC realism meta scores of human trajectories in different sets of scenarios.}
\label{tab:wosacscoreupperbound}
\begin{tabularx}{0.5\linewidth}{@{}>{\raggedright\arraybackslash}X r@{}}
\toprule
\textbf{Dataset} & \textbf{Value} \\ \midrule
Test set@Singapore & 0.8581 \\
New test set@Singapore & 0.8602 \\
Test set@Boston & 0.8311 \\
New test set@Boston & 0.8284 \\
Test set@Pittsburgh & 0.8339 \\
New test set@Pittsburgh & 0.8392 \\\bottomrule
\end{tabularx}
\end{table}

The results demonstrate the quantitative difference in the realism meta score between different datasets even within one city, which explain the outward shift of re-evaluated points.
Considering this difference, we present that is not severe overestimation of our frontier checkpoint selection.

%% file: appendices/FailureCases.tex
\section{Failure Case Study} \label{sec:failurecasestudy}

In this section, we elaborate on identified failure cases and the insights into why they happen and how they can be mitigated. 
It is worth noting that these failure cases do not happen frequently, as can be inferred from the high success rate and realism score, but we acknowledge that highlighting them is important for future extension of the work.

\begin{figure*}[htpb]
  \centering
  \includegraphics[width=0.3\linewidth]{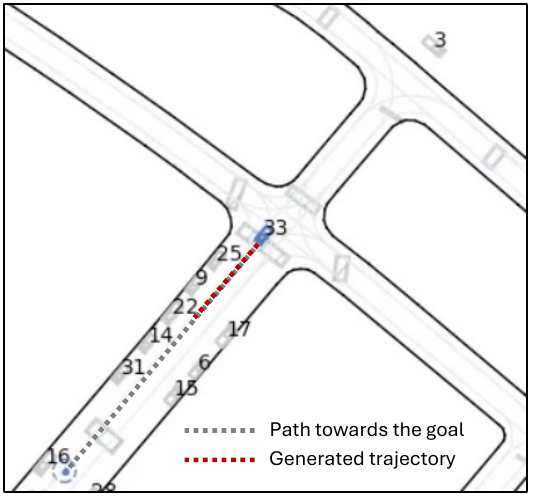}
  \caption{ 
    \textbf{Overly conservative behavior when facing curbed parking vehicles.}
  When facing many static parking vehicles, the vehicle 33 shows overly conservative behavior and thus fails to reach the goal in time.
  The gray dashed line shows the path the vehicle should follow and the red one shows the generated trajectory. 
  }
  \label{fig:failurecaseparkingvehicles}
\end{figure*}

\begin{figure*}[htpb]
  \centering
  \includegraphics[width=1.0\linewidth]{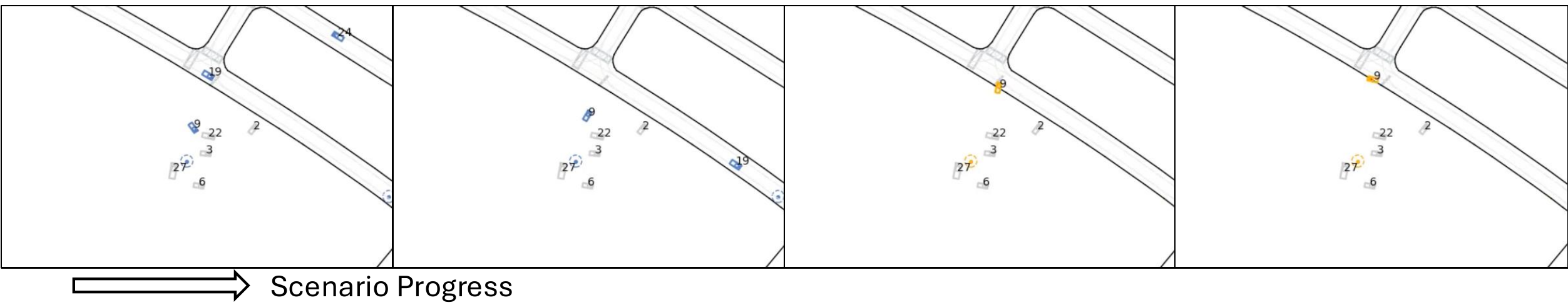}
  \caption{ 
    \textbf{Being lost if initialized off-road.}
    The vehicle 9 is initialized outside the road.
    At the beginning, the vehicle tries to go on road.
    However, as the goal is outside of the road as well, it finally turn left and drive outside of the road again.
  }
  \label{fig:failurecaseparking}
\end{figure*}

\begin{itemize}
    \item Overly conservative behavior when facing curbed parking vehicles. 
    As shown in \cref{fig:failurecaseparkingvehicles}, the vehicle is too conservative to reach the goal in time when it faces many static parking vehicles on the road. The maps available for these cities do not annotate
    curb parking zones; without this information, our scenario generator
    cannot position static vehicles at realistic curbside locations, leaving
    such configurations under-represented during training.
    \item Lost when initialized Off-road.
    In some scenarios, vehicles are initialized at positions that fall outside the mapped road network (for instance, in \cref{fig:failurecaseparking}, vehicle 9 is initialized in parking lots or access areas that are not represented in the available map data).
    Since NOMAD relies on map structure to reason about valid driving regions, the absence of road boundaries and lane geometry in these areas causes the ego policy to exhibit erratic or undefined behavior.
\end{itemize}

%% file: appendices/Qualitative.tex
\section{Qualitative results}\label{sec:qualitativeresults}

We provide extensive qualitative results in the form of representative frames of rendered videos to complement the quantitative evaluations in the main paper.
The visualizations focus on bidirectional city transfer between Boston and Singapore, comparing zero-shot transferred policies with policies adapted by NOMAD (we chose one checkpoint on the frontier for each city) under identical initial conditions.
These examples highlight characteristic failure modes induced by cross-city distribution shifts and illustrate how map-based self-play enables robust adaptation to target-city geometry and interaction patterns.

\begin{figure}[htpb]
  \centering
  \includegraphics[width=1.0\linewidth]{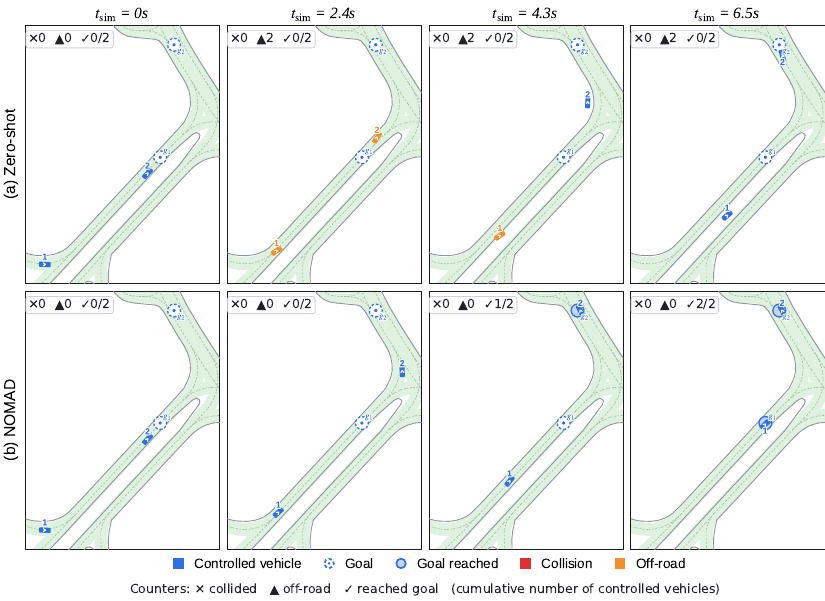}
  \caption{
  \textbf{Narrow-lane failure under Boston-to-Singapore transfer.}
  Representative frames from a Boston-to-Singapore transfer scenario illustrating geometric mismatch in narrow roads.
  \rev(a) Zero-shot transferred policy.
  (b) NOMAD-adapted policy.
  The zero-shot policy, trained in Boston, fails to account for the narrower lane geometry commonly found in Singapore: by $t_{\mathrm{sim}}=2.4$\,s vehicles 1 and 2 have both drifted outside the drivable region, and vehicle 1 never reaches its goal.
  In contrast, NOMAD adapts to the target-city road geometry through map-based self-play and maintains stable lane-centering throughout the maneuver, so both vehicles reach their goals.
  Numbers identify the controlled vehicles and $g_i$ marks the goal of vehicle $i$; a controlled vehicle is drawn in \textcolor[HTML]{F28E2B}{orange} while it is off-road.
    }
  \label{fig:qual_narrowlane}
\end{figure}

\begin{figure}[htpb]
  \centering
  \includegraphics[width=1.0\linewidth]{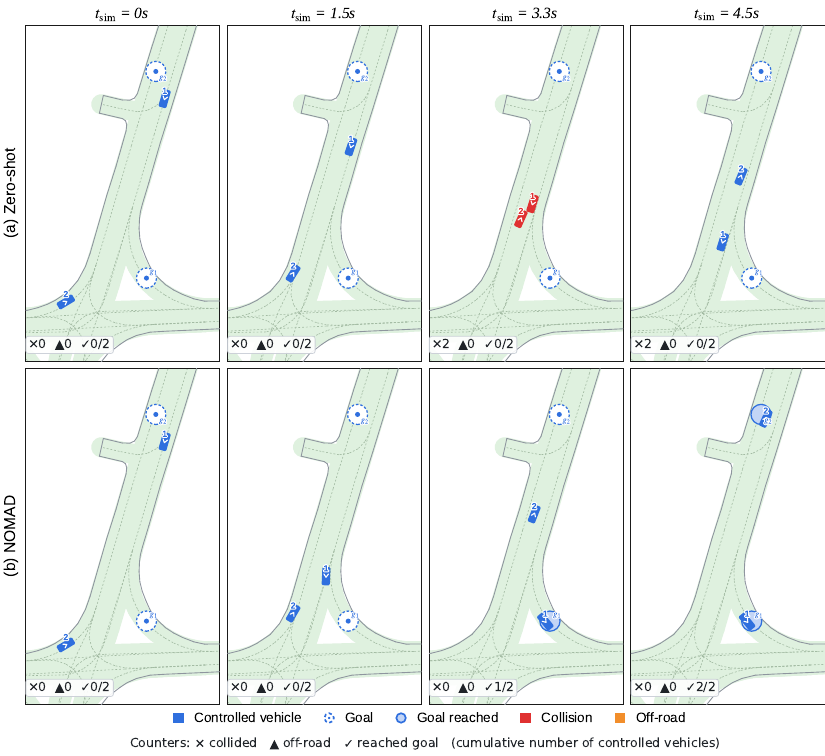}
  \caption{
  \textbf{Driving-side mismatch under Boston-to-Singapore transfer.}
  \rev(a) Zero-shot transferred policy.
  (b) NOMAD-adapted policy.
  The zero-shot policy exhibits a systematic bias toward right-hand driving learned in the source city: vehicles 1 (moving down) and 2 (moving up) both drift to their right, toward the opposing lane, and collide head-on at $t_{\mathrm{sim}}=3.3$\,s in the left-hand traffic setting of Singapore, and neither vehicle reaches its goal.
  In contrast, NOMAD adapts its lane-keeping behavior through map-based self-play and consistently drives on the correct side of the road, so both vehicles reach their goals.
  Numbers identify the controlled vehicles and $g_i$ marks the goal of vehicle $i$; a controlled vehicle is drawn in \textcolor[HTML]{E03131}{red} while it is in collision and in \textcolor[HTML]{F28E2B}{orange} while it is off-road.
    }
  \label{fig:qual_righthand}
\end{figure}

\begin{figure}[htpb]
  \centering
  \includegraphics[width=1.0\linewidth]{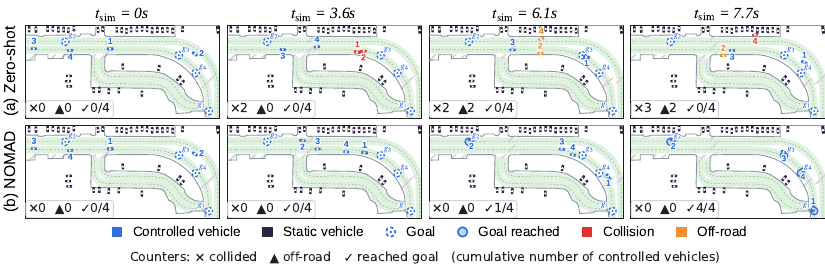}
  \caption{
  \textbf{Driving-side mismatch under Singapore-to-Boston transfer} \rev(same scenario as \cref{fig:qual_main}).
  (a) Zero-shot transferred policy.
  (b) NOMAD-adapted policy.
  The zero-shot policy retains a left-hand driving bias learned in the source city: vehicle 2 drifts into the lower lane, which belongs to oncoming traffic, and collides head-on with vehicle 1 at $t_{\mathrm{sim}}=3.6$\,s, and vehicle 4 drifts to its left until it leaves the road, resulting in lane violations, collisions, and off-road behavior in the right-hand traffic setting of Boston.
  In contrast, NOMAD adapts its lane-keeping behavior via map-based self-play and consistently drives on the correct side of the road, so all four vehicles reach their goals.
  Numbers identify the controlled vehicles and $g_i$ marks the goal of vehicle $i$; a controlled vehicle is drawn in \textcolor[HTML]{E03131}{red} while it is in collision and in \textcolor[HTML]{F28E2B}{orange} while it is off-road.
    }
  \label{fig:qual_lefthand_boston}
\end{figure}

\begin{figure}[htpb]
  \centering
  \includegraphics[width=1.0\linewidth]{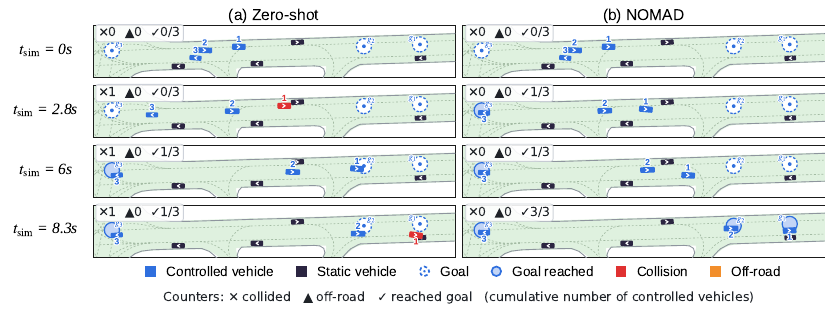}
  \caption{
  \textbf{Cascading collision failure from a Boston-to-Singapore transfer scenario involving dense static obstacles.}
  \rev(a) Zero-shot transferred policy.
  (b) NOMAD-adapted policy.
  The scene is rotated so that the vehicles of interest travel from left to right.
  Under the zero-shot policy, vehicle 1 fails to execute an early rightward avoidance maneuver and collides with the first static vehicle at $t_{\mathrm{sim}}=2.8$\,s. Following the collision, it initiates a right-turn corrective action, whereas a leftward maneuver is required to avoid the second obstacle, leading to a second collision at $t_{\mathrm{sim}}=8.0$\,s.
  In contrast, NOMAD selects appropriate avoidance actions at both decision points, and all three controlled vehicles reach their goals, illustrating improved closed-loop decision making under target-city geometric and obstacle-layout shifts.
  Numbers identify the controlled vehicles and $g_i$ marks the goal of vehicle $i$; a controlled vehicle is drawn in \textcolor[HTML]{E03131}{red} while it is in collision and in \textcolor[HTML]{F28E2B}{orange} while it is off-road.
    }
  \vspace{-0.5\baselineskip}
  \label{fig:qual_closedloop}
\end{figure}

\begin{figure}[htpb]
  \centering
  \includegraphics[width=1.0\linewidth]{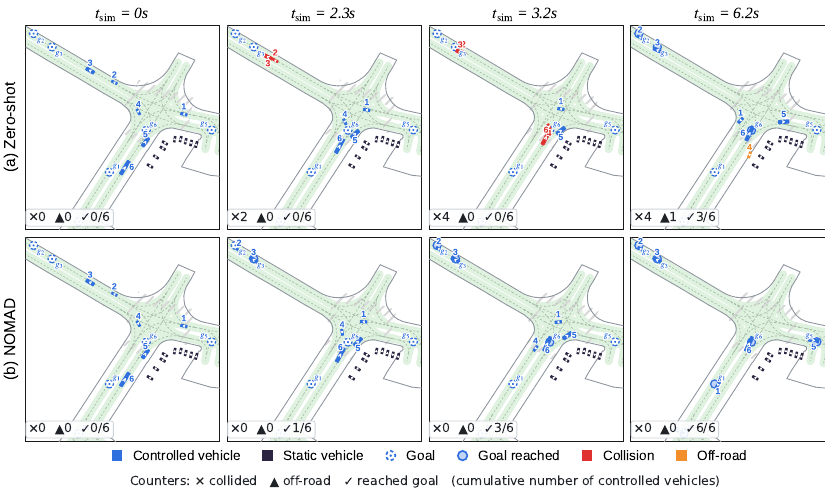}
  \caption{
  \textbf{Interaction mismatch frames from a Singapore-to-Boston transfer scenario involving multiple interacting vehicles.}
  \rev(a) Zero-shot transferred policy.
  (b) NOMAD-adapted policy.
  The zero-shot policy exhibits miscalibrated interaction behavior: on the upper-left approach, vehicle 2 travels considerably faster than vehicle 3 ahead of it (about 10 vs.\ 6\,m/s) and rear-ends it at $t_{\mathrm{sim}}=2.3$\,s, while at the intersection vehicle 6 enters without yielding to vehicle 4, causing a second collision at $t_{\mathrm{sim}}=3.2$\,s.
  In contrast, NOMAD learns more balanced speed regulation and yielding behavior through map-based self-play: all six controlled vehicles reach their goals without collision or off-road events.
  Numbers identify the controlled vehicles and $g_i$ marks the goal of vehicle $i$; a controlled vehicle is drawn in \textcolor[HTML]{E03131}{red} while it is in collision and in \textcolor[HTML]{F28E2B}{orange} while it is off-road.
    }
  \label{fig:qual_ma_boston}
\end{figure}

Across both transfer directions, zero-shot policies exhibit several recurring failure patterns. 
First, geometric mismatches (\cref{fig:qual_narrowlane}), such as narrower lane widths in Singapore, lead to unstable lane-keeping and off-road behavior when transferring from Boston.  
Second, systematic driving-side biases appear in both directions (\cref{fig:qual_righthand} and \cref{fig:qual_lefthand_boston}): policies trained in right-hand traffic tend to violate lane conventions in Singapore, while policies trained in left-hand traffic exhibit analogous failures in Boston, resulting in collisions and off-road events. 
Third, sequential decision errors under closed-loop execution arise in cluttered environments with static obstacles (\cref{fig:qual_closedloop}), where zero-shot policies make inappropriate avoidance choices at successive decision points, causing cascading collisions.
Finally, interaction-level mismatches are observed in multi-vehicle scenarios (\cref{fig:qual_ma_boston}), including miscalibrated speed regulation and incorrect yielding behavior at intersections, leading to unsafe interactions among moving agents.

In contrast, NOMAD consistently mitigates these failure modes across both transfer directions. By leveraging map-based self-play in a target-city simulator, NOMAD adapts its lane-keeping behavior to local road geometry, corrects driving-side biases, selects more appropriate avoidance and corrective actions in cluttered environments, and exhibits improved speed control and yielding behavior in multi-agent interactions. 
Together, these qualitative results reinforce the quantitative findings, demonstrating the efficacy of map-based self-play adaptation to the target city.

\clearpage 

%% file: checklist.tex
\section*{NeurIPS Paper Checklist}

\begin{enumerate}

\item {\bf Claims}
    \item[] Question: Do the main claims made in the abstract and introduction accurately reflect the paper's contributions and scope?
    \item[] Answer: \answerYes{} 
    \item[] Justification: The abstract claims that self-play multi-agent reinforcement learning can adapt driving policies to new cities where human demonstrations are unavailable, which is clearly demonstrated in \cref{sec:results}.

\item {\bf Limitations}
    \item[] Question: Does the paper discuss the limitations of the work performed by the authors?
    \item[] Answer: \answerYes{} 
    \item[] Justification: We clearly elaborate the limitations of this work in a dedicated `Limitations' Section~\ref{para:limitations} and we also show some failure cases in \cref{sec:failurecasestudy}.

\item {\bf Theory assumptions and proofs}
    \item[] Question: For each theoretical result, does the paper provide the full set of assumptions and a complete (and correct) proof?
    \item[] Answer: \answerNA{} 
    \item[] Justification: This is an empirical paper without any theoretical results.

    \item {\bf Experimental result reproducibility}
    \item[] Question: Does the paper fully disclose all the information needed to reproduce the main experimental results of the paper to the extent that it affects the main claims and/or conclusions of the paper (regardless of whether the code and data are provided or not)?
    \item[] Answer: \answerYes{} 
    \item[] Justification: We provide detailed algorithm procedure and hyperparameters used in our experiments.
    In addition, we provide code in the supplementary material.

\item {\bf Open access to data and code}
    \item[] Question: Does the paper provide open access to the data and code, with sufficient instructions to faithfully reproduce the main experimental results, as described in supplemental material?
    \item[] Answer: \answerYes{} 
    \item[] Justification: The project page \url{https://zerlinwang.github.io/NOMAD/} links to our public code, with instructions for environment setup, data download, training, and evaluation.

\item {\bf Experimental setting/details}
    \item[] Question: Does the paper specify all the training and test details (e.g., data splits, hyperparameters, how they were chosen, type of optimizer) necessary to understand the results?
    \item[] Answer: \answerYes{} 
    \item[] Justification: We provide comprehensive experimental details in Experimental Setup and Appendices.

\item {\bf Experiment statistical significance}
    \item[] Question: Does the paper report error bars suitably and correctly defined or other appropriate information about the statistical significance of the experiments?
    \item[] Answer: \answerYes{} 
    \item[] Justification: We reported results of multiple seeds in our experiments.

\item {\bf Experiments compute resources}
    \item[] Question: For each experiment, does the paper provide sufficient information on the computer resources (type of compute workers, memory, time of execution) needed to reproduce the experiments?
    \item[] Answer: \answerYes{} 
    \item[] Justification: Compute resources are detailed in the Experimental Setup section.
    
\item {\bf Code of ethics}
    \item[] Question: Does the research conducted in the paper conform, in every respect, with the NeurIPS Code of Ethics \url{https://neurips.cc/public/EthicsGuidelines}?
    \item[] Answer: \answerYes{} 
    \item[] Justification: No known violations of the Code of Ethics.

\item {\bf Broader impacts}
    \item[] Question: Does the paper discuss both potential positive societal impacts and negative societal impacts of the work performed?
    \item[] Answer: \answerYes{} 
    \item[] Justification: An Impact Statement section is included at the end of the main text.
    
\item {\bf Safeguards}
    \item[] Question: Does the paper describe safeguards that have been put in place for responsible release of data or models that have a high risk for misuse (e.g., pre-trained language models, image generators, or scraped datasets)?
    \item[] Answer: \answerNA{} 
    \item[] Justification: No high-risk application impact.

\item {\bf Licenses for existing assets}
    \item[] Question: Are the creators or original owners of assets (e.g., code, data, models), used in the paper, properly credited and are the license and terms of use explicitly mentioned and properly respected?
    \item[] Answer: \answerYes{} 
    \item[] Justification: All the assets used in the paper are properly cited.

\item {\bf New assets}
    \item[] Question: Are new assets introduced in the paper well documented and is the documentation provided alongside the assets?
    \item[] Answer: \answerNA{} 
    \item[] Justification: No new asset introduced in the paper.

\item {\bf Crowdsourcing and research with human subjects}
    \item[] Question: For crowdsourcing experiments and research with human subjects, does the paper include the full text of instructions given to participants and screenshots, if applicable, as well as details about compensation (if any)? 
    \item[] Answer: \answerNA{} 
    \item[] Justification: No crowdsourcing experiments.

\item {\bf Institutional review board (IRB) approvals or equivalent for research with human subjects}
    \item[] Question: Does the paper describe potential risks incurred by study participants, whether such risks were disclosed to the subjects, and whether Institutional Review Board (IRB) approvals (or an equivalent approval/review based on the requirements of your country or institution) were obtained?
    \item[] Answer: \answerNA{} 
    \item[] Justification: No human subject experiments.

\item {\bf Declaration of LLM usage}
    \item[] Question: Does the paper describe the usage of LLMs if it is an important, original, or non-standard component of the core methods in this research? Note that if the LLM is used only for writing, editing, or formatting purposes and does \emph{not} impact the core methodology, scientific rigor, or originality of the research, declaration is not required.
    \item[] Answer: \answerNA{} 
    \item[] Justification: LLMs are only used for purpose of paper polishing and formatting.

\end{enumerate}